\pdfoutput=1

\PassOptionsToPackage{numbers,compress,sort&compress}{natbib}
\documentclass{article}

\newif\ifanon
\anonfalse                    

\usepackage[preprint]{neurips_2026}

\makeatletter
\def\input@path{{./}}
\makeatother

\usepackage[utf8]{inputenc} 
\usepackage[T1]{fontenc} 
\usepackage{float} 
\usepackage{hyperref} 
\usepackage{url} 
\usepackage{booktabs} 
\usepackage{longtable} 
\usepackage{array} 
\usepackage{amsfonts} 
\usepackage{amsmath} 
\usepackage{nicefrac} 
\usepackage{microtype} 
\usepackage{xcolor} 
\usepackage{graphicx}
\usepackage{subcaption}
\graphicspath{{../../}{./}}
\usepackage{multirow}
\usepackage{enumerate}
\usepackage{fontawesome5}
\usepackage[textcolor=black, backgroundcolor=orange!20]{todonotes}

\usepackage{tikz}
\usepackage{pgfplots}
\pgfplotsset{compat=1.18}
\usepackage{amsmath,amssymb,amsfonts,bm}
\usepackage{xcolor}

\definecolor{tiger}{HTML}{FF5A00}
\definecolor{fire}{HTML}{FF4D00}
\definecolor{flicker}{HTML}{DC4B07}
\definecolor{black}{HTML}{878F92}
\definecolor{gunmetal}{HTML}{122128}
\definecolor{panel}{HTML}{202E35}

\definecolor{dark_navy}{HTML}{122128}
\definecolor{forgis_orange}{HTML}{FF5A00}
\definecolor{forgis_fire}{HTML}{FF4D00}
\definecolor{forgis_flicker}{HTML}{DC4B07}
\definecolor{blue}{HTML}{3A7BD5}
\definecolor{green}{HTML}{2ECC71}
\definecolor{panelbg}{HTML}{F0F4F8}
\definecolor{warm_amber}{HTML}{FFF3CD}

\definecolor{L1bg}{HTML}{FCF0E0}
\definecolor{L1border}{HTML}{FB9E6C}
\definecolor{L2bg}{HTML}{FCF0E0}
\definecolor{L2border}{HTML}{FF7F3A}
\definecolor{L3bg}{HTML}{FCF0E0}
\definecolor{L3border}{HTML}{FF5A00}
\definecolor{L4bg}{HTML}{FCF0E0}
\definecolor{L4border}{HTML}{81360F}

\definecolor{darknavy}{HTML}{122128}
\definecolor{forgisorange}{HTML}{FF5A00}
\definecolor{figgreen}{HTML}{2ECC71}
\definecolor{figsteel}{HTML}{878F92}
\definecolor{figcontext}{HTML}{D4A017} 
\definecolor{L1tint}{HTML}{4A6FA5} 
\definecolor{L2tint}{HTML}{C9A227} 
\definecolor{L3tint}{HTML}{FF5A00} 
\definecolor{L4tint}{HTML}{6F1A0A} 

\usetikzlibrary{arrows.meta, positioning, calc, fit, shapes.geometric,
  decorations.pathreplacing, decorations.markings, backgrounds, patterns, shadows}

\providecommand{\paperbibstyle}{unsrt}

\newif\ifneuripschecklist
\neuripschecklisttrue

\newif\ifworkshopcompact
\workshopcompactfalse

\newif\iffloatsinappendix
\floatsinappendixfalse

\newif\ificlrstatements
\iclrstatementsfalse

\newcommand{\artifactlinks}{%
  \ifanon
    {\faGithub}\ Code: \href{https://anonymous.4open.science/r/FactoryBench/README.md}{\nolinkurl{anonymous.4open.science/r/FactoryBench}} \\[0.15em]
    \raisebox{-0.25ex}{\includegraphics[height=1em]{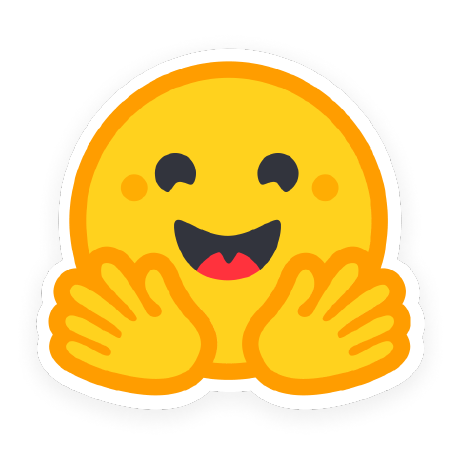}}\ Dataset: \href{https://huggingface.co/datasets/FactoryBench/FactoryBench}{\nolinkurl{huggingface.co/datasets/FactoryBench/FactoryBench}}%
  \else
    {\faGithub}\ Code: \href{https://github.com/Forgis-Labs/Factorybench}{\nolinkurl{github.com/Forgis-Labs/Factorybench}} \\[0.15em]
    \raisebox{-0.25ex}{\includegraphics[height=1em]{figures/huggingface_logo.png}}\ Dataset: \href{https://huggingface.co/datasets/FactoryBench/FactoryBench}{\nolinkurl{huggingface.co/datasets/FactoryBench/FactoryBench}}%
  \fi
}

\newcommand{\floatBenchmarkComparison}{%
\begin{table}[htbp]
    \centering
    \caption{Comparison of FactoryBench with existing benchmarks. ``Counterfactual'' = questions reasoning over alternative histories; ``Free-Form'' = open-ended answers judged by LLMs; ``Novel Dense TS'' = grounded in a purpose-collected dense multivariate corpus rather than public or synthetic sources alone.}
    \label{tab:benchmark_comparison}
    \small
    \resizebox{\textwidth}{!}{%
        \begin{tabular}{@{}lcccccccc@{}}
            \toprule
            Benchmark                    & \rotatebox{60}{Machine/Robot} & \rotatebox{60}{Multivariate TS} & Number of QAs & \rotatebox{60}{Counterfactual} & \rotatebox{60}{Ranking} & \rotatebox{60}{Numerical} & \rotatebox{60}{Free-Form} & \rotatebox{60}{Novel Dense TS} \\
            \midrule
            TimeSeriesExam               & \texttimes                    & \texttimes                      & $\sim$700     & \texttimes                     & \checkmark              & \texttimes                & \texttimes                & \texttimes                     \\
            ChatTS                       & \texttimes                    & \checkmark                      & $\sim$134k    & \texttimes                     & \checkmark              & \checkmark                & \texttimes                & \texttimes                     \\
            EngineMT-QA                  & \checkmark                    & \checkmark                      & $\sim$110k    & \texttimes                     & \checkmark              & \checkmark                & \checkmark                & \texttimes                     \\
            TSAQA                        & Partial                       & \texttimes                      & $\sim$210k    & \texttimes                     & \checkmark              & \texttimes                & \texttimes                & \texttimes                     \\
            Time-MQA                     & \texttimes                    & \checkmark                      & $\sim$200k    & \texttimes                     & \checkmark              & \checkmark                & \checkmark                & \texttimes                     \\
            MTBench                      & \texttimes                    & \checkmark                      & $\sim$3k      & \texttimes                     & \checkmark              & \checkmark                & \checkmark                & \texttimes                     \\
            QuAnTS                       & \texttimes                    & \checkmark                      & $\sim$150k    & \texttimes                     & \checkmark              & \checkmark                & \checkmark                & \texttimes                     \\
            CLEVR~\cite{johnson2017clevr} & \texttimes                    & \texttimes                      & $\sim$1M      & \texttimes                     & \texttimes              & \checkmark                & \texttimes                & \texttimes                     \\
            CLEVRER~\cite{yi2020clevrer} & \texttimes                    & \texttimes                      & $\sim$305k    & \checkmark                     & \texttimes              & \checkmark                & \checkmark                & \texttimes                     \\
            \midrule
            \textbf{FactoryBench (Ours)} & \checkmark                    & \checkmark                      & $\sim$71k     & \checkmark                     & \checkmark              & \checkmark                & \checkmark                & \checkmark                     \\
            \bottomrule
        \end{tabular}
    }
\end{table}
}

\newcommand{\floatLevelsExamples}{%
\begin{table}[htbp]
    \centering
    \caption{FactoryBench levels, what they test, example questions, and expected answer formats.}
    \label{tab:levels_examples}
    \footnotesize
    \setlength{\tabcolsep}{4pt}
    \renewcommand{\arraystretch}{1.15}
    \begin{tabular}{@{}c l >{\raggedright\arraybackslash}p{4.6cm} >{\raggedright\arraybackslash}p{3.0cm} >{\raggedright\arraybackslash}p{1.7cm}@{}}
        \toprule
        Level      & Type           & Example question                                                                                                                                                 & What it tests                                                                   & Answer format \\
        \midrule
        \textbf{1} & State          & ``We want to isolate the lifting phase in the robot's time series. Assuming a fixed window length of 15 timesteps, at which timestamp should the window begin?'' & Can the model understand what the machine is doing and how it is behaving?      & Tensor        \\
        \textbf{2} & Intervention   & ``A collision with a foam cube occurs at $T{=}850$\,ms. Rank signal segments (A--D) in the order you would expect them to appear after the event.''              & Can the model reason about how the machine will behave in the face of an event? & Ranking       \\
        \textbf{3} & Counterfactual & ``Had a payload misconfiguration occurred at $T{=}200$\,ms, what would the target torque on joint~2 at $T{+}50$\,ms have been in this counterfactual case?''     & Can the model reason accurately about theoretical scenarios?                    & Scalar        \\
        \textbf{4} & Decision       & ``Given the sensor stream below, does the machine show signs of anomalous behavior? If yes, identify the root cause and the steps to fix it.''                   & Can the model make informed decisions about the machine?                        & Free-form     \\
        \bottomrule
    \end{tabular}
\end{table}
}

\newcommand{\floatFactoryWaveCollage}{%
\begin{figure}[!h]
    \centering
    \includegraphics[width=0.95\linewidth]{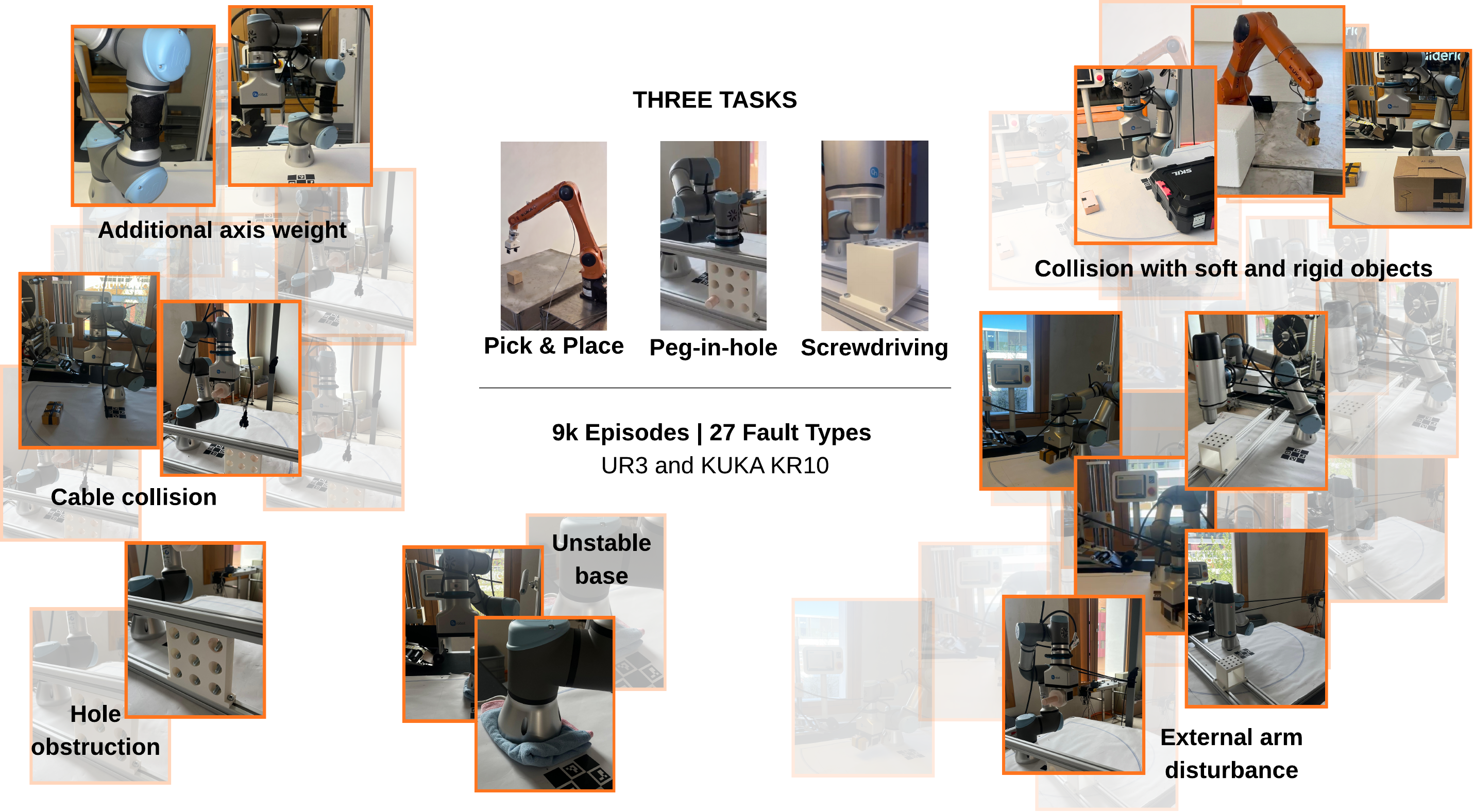}
    \caption{FactoryWave tasks and fault catalogue. Photographs of the robotic platforms (UR3, KUKA KR10) executing the three tasks (pick-and-place, peg-in-hole, screwdriving) and a representative subset of the 27 systematically injected fault conditions.}
    \label{fig:factorywave_collage}
\end{figure}
}

\newcommand{\floatDatasets}{%
\begin{table}[htbp]
    \centering
    \caption{Datasets incorporated into FactoryBench. FactoryWave is released with FactoryNet~\cite{othman2026factorynetlargescaledatasetindustrial}; all three are open-source datasets adapted to our schema. Tasks: PnP = pick-and-place, Scr = screwing, PiH = peg-in-hole.}
    \label{tab:datasets}
    \small
    \setlength{\tabcolsep}{4pt}
    \begin{tabular}{@{}l l c c c c@{}}
        \toprule
        Dataset                                & Robot          & Episodes & Frequency   & Tasks         & Anomalies \\
        \midrule
        AURSAD~\cite{leporowski2022aursad}     & UR3e           & 4094     & 100\,Hz     & Scr           & 4         \\
        voraus-AD~\cite{brockmann2023vorausad} & Yu-Cobot       & 2122     & 100/500\,Hz & PnP           & 12        \\
        FactoryWave~\cite{othman2026factorynetlargescaledatasetindustrial} & UR3, KUKA KR10 & 8983     & 125/83\,Hz  & PnP, Scr, PiH & 27        \\
        \bottomrule
    \end{tabular}
\end{table}
}

\providecommand{\herowidth}{\linewidth}

\providecommand{\herogapabove}{-0.4em}
\providecommand{\herogapbelow}{-1.5em}

\newif\ifheroteaser
\heroteasertrue

\newif\ifabstractpagebreak
\abstractpagebreakfalse

\providecommand{\mainresultsafrac}{0.88}
\providecommand{\mainresultsbfrac}{1.0}

\graphicspath{{./}{figures/}}

\renewcommand{\herowidth}{0.90\linewidth}

\neuripschecklistfalse

\iclrstatementsfalse

\title{\raisebox{-0.5ex}{\includegraphics[height=1.2em]{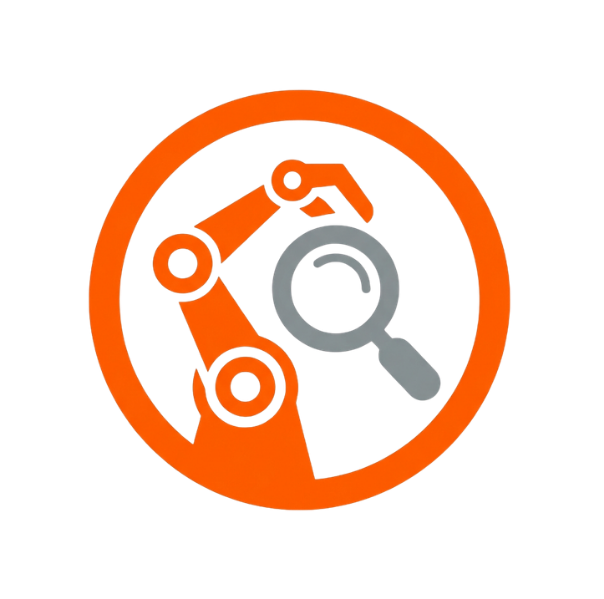}}FactoryBench: Evaluating Industrial\\Machine Understanding}

\ifanon
\author{\parbox{\dimexpr\textwidth-2\tabcolsep}{\centering\footnotesize
  Anonymous Authors\\[0.6em]
  \normalfont\footnotesize
  Paper under double-blind review \\[0.5em]
  \small
  \artifactlinks
}}
\else
\author{\parbox{\dimexpr\textwidth-2\tabcolsep}{\centering\footnotesize
  Yanis Merzouki$^{1,2,*}$ \quad
  Coral Izquierdo$^{2,3}$ \quad
  Matei Ignuta-Ciuncanu$^{4,5}$ \quad
  Marcos G\'{o}mez-Bracamonte$^{1,6}$ \\
  Riccardo Maggioni$^{2}$ \quad
  Alessandro Lombardi$^{2}$ \quad
  Camilla Mazzoleni$^{2}$ \quad
  Federico Martelli$^{2,1}$ \quad
  Bal\'{a}zs G\"{u}nther$^{1}$ \\
  Jonas Petersen$^{1,2,\dagger}$ \quad
  Philipp Petersen$^{7,\dagger}$ \\[0.6em]
  \normalfont\footnotesize
  $^{1}$ETH Zurich \enspace $^{2}$Forgis \enspace $^{3}$UC3M \enspace $^{4}$Imperial College London \enspace $^{5}$University of Berkeley \\
  $^{6}$KTH Royal Institute of Technology \enspace $^{7}$University of Vienna \\[0.2em]
  $^{*}$Correspondence: \texttt{ymerzouki@ethz.ch} \qquad $^{\dagger}$Equal senior contribution \\[0.5em]
  \small
  \artifactlinks
}}
\fi

\makeatletter
\let\fb@links@once\artifactlinks
\renewcommand{\artifactlinks}{%
  {\faGithub}\ Code: \href{https://github.com/Forgis-Labs/FactoryBench}%
    {\nolinkurl{github.com/Forgis-Labs/FactoryBench}} \\[0.15em]
  \raisebox{-0.25ex}{\includegraphics[height=1em]{figures/huggingface_logo.png}}\
    Dataset: \href{https://huggingface.co/datasets/Forgis/FactoryBench}%
    {\nolinkurl{huggingface.co/datasets/Forgis/FactoryBench}}%
  \gdef\artifactlinks{}%
}
\makeatother

\begin{document}
%
%

\maketitle

\begin{center}\vspace{-2.4em}\small
    \artifactlinks
\end{center}
\vspace{\herogapabove}
\ifheroteaser
\input{figures/figure_pipeline}
\vspace{\herogapbelow}
\fi
\begin{abstract}

    We introduce \textbf{FactoryBench}, a benchmark for evaluating time-series models and LLMs on machine understanding over industrial robotic telemetry. Q\&A pairs are organized along four causal levels (state, intervention, counterfactual, decision) instantiating Pearl's ladder of causation, and span five answer formats: four structured formats are scored deterministically and free-form answers are scored by an LLM-as-judge voting protocol. We propose a scalable Q\&A generation framework built around structured question templates and construct FactoryBench as a large-scale benchmark of $69{,}691$ Q\&A items grounded in roughly 15k normalized episodes: FactoryWave, the dense multitask corpus recorded on a UR3 cobot and a KUKA KR10 industrial arm and released with FactoryNet~\cite{othman2026factorynetlargescaledatasetindustrial}, together with AURSAD~\cite{leporowski2022aursad} and voraus-AD~\cite{brockmann2023vorausad}. Zero-shot evaluation of six frontier LLMs shows that no model exceeds 50\% (chance-corrected) on the structured levels and that the panel ordering reshuffles on the decision-making level, with the L1--L3 leader falling to fourth of six on L4; both findings reveal a wide gap between current models and operational machine understanding.
\end{abstract}

\ifabstractpagebreak\clearpage\fi

\section{Introduction}

Modern robotic manufacturing systems emit dense multivariate telemetry, encoding joint states, torques, forces, velocities, contact events, task phases, and fault indicators. Extracting actionable knowledge from these signals is central to monitoring, diagnostics, anomaly detection, and decision support.

Traditional time-series models excel at narrow tasks such as prediction, classification, and anomaly detection \cite{zhou2021informer,su2019omni,audibert2020usad}, but they are typically specialized, hard to interpret, and generalize poorly outside their training task \cite{chalapathy2019survey,lavin2015numenta,zeng2023transformers}, limiting their utility for automated engineering decisions in a factory setting. Large language models, in contrast, possess strong general reasoning and can produce structured technical explanations \cite{wei2022chain,yao2023tree}. They forecast univariate series competently \cite{gruver2023llmtime}, more so when reprogrammed into language \cite{jin2024timellm}; whether that transfers to dense multivariate industrial telemetry is what we test. Specialized transformer architectures meanwhile continue to advance time-series forecasting and representation learning \cite{vaswani2017attention,zhou2021informer,wu2021autoformer,zhou2022fedformer,nie2022time,ansari2024chronos,das2024decoder,petersen2026hepaselfsupervisedhorizonconditionedevent}.

Evaluating whether language-based systems truly reason about machine behavior, rather than pattern-matching on textual cues, remains challenging. By \emph{machine-behavior understanding} we mean the ability to (i)~interpret the current operational state from raw multivariate signals, (ii)~predict the effect of an intervention, (iii)~reason counterfactually about alternative histories, and (iv)~recommend remedial actions grounded in physical and engineering constraints. How current LLMs perform across these four competences on dense industrial telemetry has not yet been systematically measured. General benchmarks such as MMLU, BIG-bench, and MMMU \cite{hendrycks2021mmlu,srivastava2023bigbench,yue2024mmmu} target textual or multimodal understanding and cannot probe machine behavior; closer-in-spirit time-series Q\&A benchmarks (TimeSeriesExam~\cite{cai2024timeseriesexamtimeseriesunderstanding}, ChatTS~\cite{Xie_2025}, EngineMT-QA~\cite{wang2025itformerbridgingtimeseries}, TSAQA~\cite{jing2026tsaqatimeseriesanalysis}, Time-MQA~\cite{kong2025timemqatimeseriesmultitask}, MTBench~\cite{chen2026mtbenchmultimodaltimeseries}, QuAnTS~\cite{divo2025quants}; Table~\ref{tab:benchmark_comparison}) restrict themselves to synthetic or univariate data, omit counterfactual reasoning, or do not involve a robotic system under closed-loop control.

To address this gap we propose FactoryBench, a benchmark for machine-behavior reasoning in time-series models and LLMs. FactoryBench is grounded in \textbf{FactoryWave}~\cite{othman2026factorynetlargescaledatasetindustrial}, a dense multivariate dataset collected from collaborative and industrial one-arm platforms (UR3 at 125\,Hz, KUKA KR10 at 83\,Hz) with synchronized setpoint, context, feedback, and effort signals. Industrial-robot data of this kind is rare in public benchmarks: such systems live in restricted production environments behind proprietary controllers with limited telemetry and plant-level confidentiality. To populate the benchmark, we introduce a scalable question-generation framework of 21 structured templates spanning state, intervention, counterfactual, and decision-making reasoning, applicable to general machine data flows; instantiated over FactoryWave (and AURSAD / voraus-AD), they yield the full FactoryBench Q\&A corpus, whose level, answer-format, and context-length distribution we summarise in Appendix~\ref{app:qa_distribution}. For cheaper evaluation we additionally release \textbf{FactoryBench-Lite}, a $3{,}000$-item subset that is balanced both across templates and within each template on the dimension that determines its answer, so that level aggregates are not weighted by how many items a template happens to contribute (Appendix~\ref{app:lite}).

\section{Related work}

Recent advances in LLM reasoning include prompting and deliberative strategies (e.g., Chain-of-Thought and Tree-of-Thought) that improve multi-step performance \cite{wei2022chain,yao2023tree}, alongside tool-augmented paradigms such as Toolformer and ReAct that enable grounded computation through external modules \cite{schick2023toolformer,yao2023react,qin2023tool}. These ideas motivate industrial agent design where symbolic reasoning must be combined with numerical workflows.

Time-series modeling has advanced through transformer-based architectures (Informer, Autoformer, FEDformer, PatchTST) \cite{zhou2021informer,wu2021autoformer,zhou2022fedformer,nie2022time}, strong alternative baselines such as TFT and N-BEATS \cite{lim2021tft,oreshkin2020nbeats}, and critical comparisons against simpler linear models \cite{zeng2023transformers}. Foundation-model directions (e.g., Chronos) further explore transfer and zero/few-shot adaptation \cite{ansari2024chronos,das2024decoder}, with large-scale industrial time-series corpora such as FactoryNet \cite{othman2026factorynetlargescaledatasetindustrial}, on which FactoryBench builds. In parallel, anomaly and reliability monitoring literature includes OmniAnomaly, USAD, and Deep SVDD \cite{su2019omni,audibert2020usad,ruff2018deep}, with surveys and benchmarks highlighting persistent gaps in interpretability and root-cause actionability \cite{chalapathy2019survey,lavin2015numenta}.

Causal frameworks provide formal tools for interventions and counterfactuals \cite{pearl2009causality,peters2017elements,rubin1974causal}, while temporal methods such as Granger causality and modern nonlinear causal discovery support directional reasoning in time series \cite{granger1969causal,runge2019causal}. Knowledge-graph representations further structure domain knowledge for transformer-based reasoning \cite{petersen2026structuralinductivebiashelps}. Existing broad benchmarks (MMLU, BIG-bench, MMMU) \cite{hendrycks2021mmlu,srivastava2023bigbench,yue2024mmmu} and classical time-series resources \cite{laptev2015generic,dau2019ucr,wen2022transformers} do not directly evaluate machine-centered reasoning over industrial telemetry. FactoryBench targets this gap by jointly testing temporal interpretation, causal reasoning, and engineering decision support. Table~\ref{tab:benchmark_comparison} positions FactoryBench against closely related Q\&A and time-series benchmarks along eight axes. The column labels refer to concepts introduced later in the paper: Counterfactual denotes questions that reason about alternative histories (Level~3, Section~\ref{sec:levels}); Free-Form denotes open-ended natural-language answers, scored by an LLM-as-judge voting protocol (Appendix~\ref{app:answer-formats}); and Novel Dense TS indicates whether the benchmark contributes a new, high-frequency multivariate dataset rather than relying exclusively on public sources.

\iffloatsinappendix\else\floatBenchmarkComparison\fi

\section{FactoryBench framework}


\subsection{Four levels of machine understanding}
\label{sec:levels}

\iffloatsinappendix\else\floatLevelsExamples\fi

Our four-tier hierarchy is explicitly built on top of Pearl's ladder of causation (association, intervention, counterfactual reasoning), which has become the organizing principle for causal inference and is increasingly adopted as an evaluation axis for machine-learning systems \cite{pearl2009causality,peters2017elements,jin2023cladder,pawlowski2020deep_scm,peyrard2020ladder}. We extend the causal hierarchy with a fourth decision-making level that reflects how industrial robotic platforms are actually operated: after interpreting state and causal relationships, operators must select and execute corrective procedures, typically prescribed by vendor manuals (e.g., Universal Robots error-code handbooks). This final level aligns with the diagnose-then-act loop that is standard in fault-tolerant control. Grounding the hierarchy in these established principles ensures that each level probes a qualitatively distinct reasoning skill and that failure modes are interpretable in terms of the underlying causal or decision-theoretic primitive.

To systematically evaluate machine-behavior reasoning, we organize question-answering tasks according to this four-tier hierarchy, each probing distinct reasoning capabilities:

\textbf{Level 1: State.} This level assesses the agent's ability to interpret the current state of the machine under normal conditions, including identification of operational modes, prediction and recognition of sensor patterns. Questions at this level require accurate extraction and interpretation of time series features.

\textbf{Level 2: Intervention.} At this level, the agent must reason about the consequences of interventions or events occurring at the present timestep that might perturb the distribution of machine states. Tasks include predicting the immediate impact of control actions, diagnosing faults as they arise, and understanding causal relationships in real time for both normal and anomalous episodes.

\textbf{Level 3: Counterfactual.} This level evaluates the agent's ability to reason about hypothetical scenarios, such as the effect of an event or intervention at a previous timestep. Questions require the agent to simulate alternative histories and assess how outcomes would differ under counterfactual conditions, while still considering the history they know.

\textbf{Level 4: Decision Making.} The highest level evaluates the agent's ability to act on its understanding of the system. Given a recorded episode and a task description, the agent must produce a sequence of actions or recommendations to answer that prompt. In FactoryBench, this targets troubleshooting and optimization, and combines the skills exercised at the previous three levels: reading the system state, reasoning about causes, and weighing alternative interventions. It reflects what we expect from an LLM acting as an engineering assistant on the factory floor.

By structuring Q\&A tasks along these four levels, FactoryBench enables rigorous and granular assessment of machine-behavior reasoning, from basic state recognition to advanced decision support. Figure~\ref{fig:pearl_ladder} (Appendix~\ref{app:templates}) instantiates Pearl's three causal levels (L1--L3) on robotic time-series data and adds the L4 decision-making layer that FactoryBench introduces on top. Each question is emitted in one of five answer formats (single-select MCQ, multi-select MCQ, ranking, tensor, free-form); the four structured formats are scored deterministically and free-form answers are scored by an LLM-as-judge voting protocol, with per-format details in Appendix~\ref{app:answer-formats} and an analysis of the three judges' inter-rater reliability in Appendix~\ref{app:judge_reliability}.

\subsection{Time series data and causal schema (SCE)}
\label{sec:sce}

So that the dataset structure itself encodes causality, all time-series signals are organized into three causal groups: \textbf{Setpoint} (the controller's commanded target), \textbf{Context} (physical and configured conditions, split into static episode metadata and dynamic time-series channels), and \textbf{Effort+Feedback} (the machine's measured response). Under healthy operation the response is a known function of setpoint and context; faults manifest as structured residuals from that baseline, giving a single operational fault definition that applies across machines and tasks. Full per-group channel mappings, the formal residual definition, and per-robot calibration details are deferred to Appendix~\ref{app:sce}.

The data conforms to this schema across two source types:

\begin{itemize}
    \item \textbf{FactoryWave:} A custom dataset generated from one-arm robotic platforms executing industrial tasks (e.g., pick-and-place) under varied conditions and systematically injected anomalies.
    \item \textbf{Open Source Datasets:} Adapted datasets such as AURSAD~\cite{leporowski2022aursad} and voraus-AD~\cite{brockmann2023vorausad}, offering diverse industrial scenarios and preprocessed to conform to the unified episode structure.
\end{itemize}

The datasets integrated into FactoryBench are summarized in Table~\ref{tab:datasets}; all provide the full setpoint--context--effort triplet at 83\,Hz or higher and have been (re)labelled to conform to our unified episode schema.

\iffloatsinappendix\else\floatDatasets\fi

\section{Reliable Q\&A generation}

\subsection{Scalable generation via extensive labeling}

Scalable ground-truth generation is the central challenge of any Q\&A benchmark grounded in raw data. FactoryBench addresses this by coupling a structured labeling ontology with a context-free grammar (CFG)-style template system, designed by robotics experts and PhD researchers. Each template contains variable slots filled at generation time from one of two sources: label information read directly from the episode data (signal names, timestamps, anomaly labels, root causes), or a predefined sampling distribution attached to that slot (prediction horizons, numerical thresholds, curated vocabularies for discrete options). The discrete vocabularies are seeded from the label sets of AURSAD~\cite{leporowski2022aursad} and voraus-AD~\cite{brockmann2023vorausad}, extended to cover FactoryWave-specific cases, and released in full so every instantiation is inspectable and reproducible. The context time series used to fill the variables is sampled uniformly by data source (open source vs FactoryWave), then dataset, experiment, length within controlled bounds, and finally placement, yielding a combinatorial expansion from a small set of carefully designed templates into a large, diverse question pool. Three parts of this pipeline are specified in full in the appendix: which episodes are eligible to be drawn from at all (Appendix~\ref{app:episode_filtering}), how the window is placed relative to an injected fault so that the evidence needed to answer is actually inside it (Appendix~\ref{app:relevance}), and which channels and window lengths make up the context a model receives (Appendix~\ref{app:context_construction}).

Each of the 21 templates is manually authored to probe one specific machine-understanding level while remaining general enough to admit a wide range of instantiations across episode segments, signal names, event descriptions, timestamps, and predicted values. Answer options for single- and multi-select questions are drawn from a shared pool of pre-defined verifiable statements (for example, the question \emph{``What anomaly is present in this robotic sensor time series?''} draws its options from the full catalogue of documented anomalies), each paired with a rule that can be evaluated deterministically against the time series given the densely labelled data, so ground truth is computed from the underlying labels rather than imputed; for single-select MCQ items, the option-sampling step is designed so that exactly one of the four sampled options evaluates to true on the chosen episode. This is also our primary defense against the natural concern that template-generated benchmarks can be solved by learning template surface structure rather than the underlying reasoning: every template admits a combinatorially large instantiation space over signal names, timestamps, prediction horizons, payload conditions, and injected-fault types, so two questions sharing the same template rarely overlap in more than a fraction of their variable slots, and the correct answer is never recoverable from the question text alone since it is always determined by the paired time series, which varies across instantiations. We test that claim rather than assert it: Appendix~\ref{app:solvability} reports the solvability audit, including a noise-substitution check that replaces the context with uninformative signal to detect templates answerable from wording alone. Representative instantiated questions with their gold answers, covering every level and every answer format, are given in Appendix~\ref{app:qa_examples} alongside the full template inventory.

\subsection{Density of FactoryWave}
\label{sec:density}

\iffloatsinappendix\else\floatFactoryWaveCollage\fi

FactoryWave is collected from two physical robots (UR3 and the KUKA KR10; Figure~\ref{fig:factorywave_collage}) executing up to three industrial tasks each: pick-and-place, screwing, and peg-in-hole. Each complete task cycle constitutes one episode, recorded at 125 and 83\,Hz across over 100 sensor channels covering joint setpoints, feedback, torques, speeds, estimated contact forces, TCP pose, gripper state, and task-phase labels. To compensate for the limited tool-side telemetry exposed by the KUKA KSS controller, the collection setup adds a 9-DoF inertial measurement unit on the KUKA gripper, streaming its acceleration, angular-rate, and orientation channels in sync with the controller signals. The full per-robot, per-task, per-condition breakdown of the 8{,}983 episodes is given in Table~\ref{tab:factorywave_distribution} of Appendix~\ref{app:data_recording}.

Episodes are organized into three experiment types. \textbf{Normal} episodes capture nominal operation across three payload levels (light, medium, heavy for pick-and-place). \textbf{Fault} episodes inject one of 27 fault types (spanning gripper failures, misconfigurations, collisions, and peg-in-hole-specific anomalies), each recorded with fault-specific metadata and, where applicable, a precise injection timestep. \textbf{Counterfactual} episodes provide ground truth for Level~3 causal reasoning by approximating the do-operation on physical hardware, via the protocol described next.

Counterfactual generation approximates the last step of Pearl's ladder on physical hardware, which cannot reproduce a bit-identical pre-injection state. For each baseline the task is re-executed 3--5 times with every controllable condition held fixed (object pose, payload, controller configuration, exact scripted trajectory) and the target fault injected at a fixed timestep $t$. The run retained is the one whose pre-injection segment minimizes the signature kernel MMD against the baseline, and it serves as an approximate ground truth for hypothetical divergence of baseline.

\subsection{Knowledge graph and protocol extraction}
\label{sec:knowledge-graph}

Reliable ground truth for Level 4 troubleshooting questions requires more than episode labels: it demands machine-specific recovery protocols grounded in manufacturer documentation. For each robot in FactoryBench we parse the manufacturer's official runtime error documentation into a structured mapping from error codes to error names, descriptions, and recommended recovery steps, capturing what the controller reports when a fault occurs and what an operator should do to resolve it. We then map every anomaly type studied in FactoryBench to the most likely corresponding manufacturer error code, with PhD-level robotics experts reasoning about the physical cause of each anomaly and identifying which runtime error it would most plausibly trigger on the target robot. Where a physical fault injection (such as gripper misactivation, peg misalignment, or external collision) has no well-defined controller error, the same experts author the recovery protocol directly using the same structure for consistency.

The complete mapping (error-derived and expert-authored protocols alike) is ingested into a knowledge graph alongside robot specifications, task definitions, and signal schema. At question generation time, the pipeline queries this graph to automatically assemble the relevant protocol context and inject it as ground truth into Level 4 troubleshooting Q\&A pairs, without requiring per-question human review.

\section{Evaluation and results}

\subsection{Zero-shot evaluation}
\label{sec:zero-shot}

We evaluate a cross-vendor panel of frontier LLMs, each held at a pinned provider snapshot: Claude Sonnet 4.6 (\texttt{claude-sonnet-4-6}) \cite{anthropic2025sonnet46} and GPT-5.1 (\texttt{gpt-5.1-2025-11-13}) \cite{openai2025gpt51} on the closed-weight side, and DeepSeek V3.2 \cite{deepseek2025v32}, Mistral Large 3 \cite{mistral2025large3}, and Qwen3-235B (\texttt{qwen3-235b-a22b-instruct-2507}) \cite{qwen2025qwen3} on the open-weight side. We also include Qwen3-4B \cite{qwen2025qwen3} as a lightweight open-weight reference. For calibration we report a non-LLM baseline dispatching per item by answer format: linear regression on the target signal for scalar and tensor templates, and uniform random over the option set for single- and multi-select MCQ and ranking templates. Free-form Level~4 templates are excluded (no traditional method maps cleanly to a remediation protocol). Because a linear fit is a weak stand-in for specialised time-series modelling, we also run two time-series foundation models of comparable size but disjoint pretraining, Chronos-Bolt~\cite{ansari2024chronos} and TimesFM-2.5~\cite{das2024decoder}, on the numeric templates they can address; their results, with parsing, scoring and chance correction held fixed across both, are in Appendix~\ref{app:chronos-baseline}. Figure~\ref{fig:results} reports chance-corrected accuracy for the full panel. The panel is evaluated on FactoryBench-Lite (Appendix~\ref{app:lite}), the balanced $3{,}000$-item subset, so each level aggregate weights its templates equally rather than inheriting the episode pool's template mass; every model answers identical items.

\begin{figure}[htbp]
    \centering
    \begin{subfigure}[t]{\linewidth}
        \centering
        \includegraphics[width=\mainresultsafrac\linewidth]{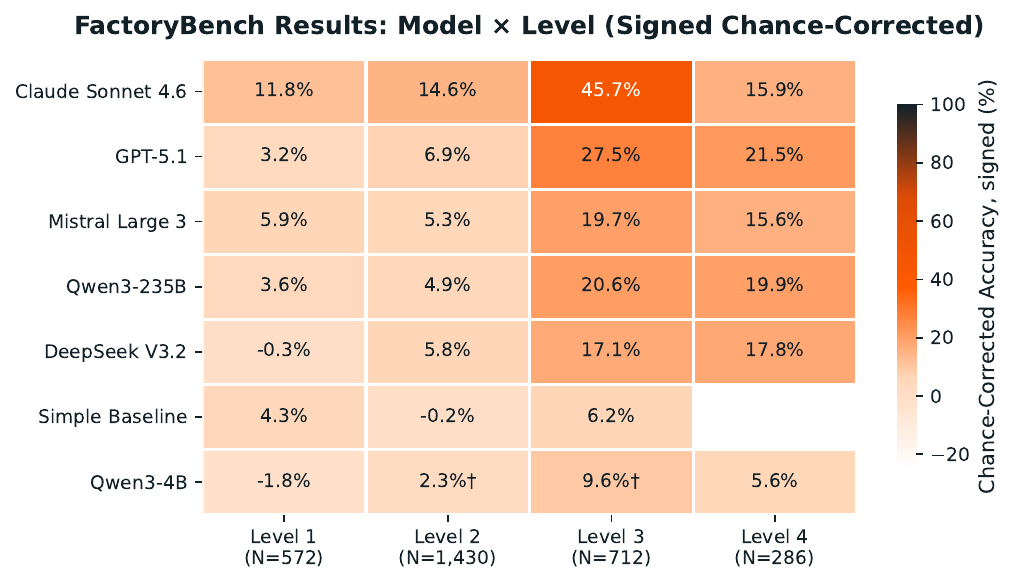}
        \caption{Chance-corrected accuracy (\%) on FactoryBench's four reasoning levels. $\dagger$~Rendered under the earlier correction; regenerated for the camera-ready.}
        \label{fig:results-heatmap}
    \end{subfigure}\\[0.6em]
    \begin{subfigure}[t]{\linewidth}
        \centering
        \makebox[\linewidth][c]{\includegraphics[width=\mainresultsbfrac\linewidth]{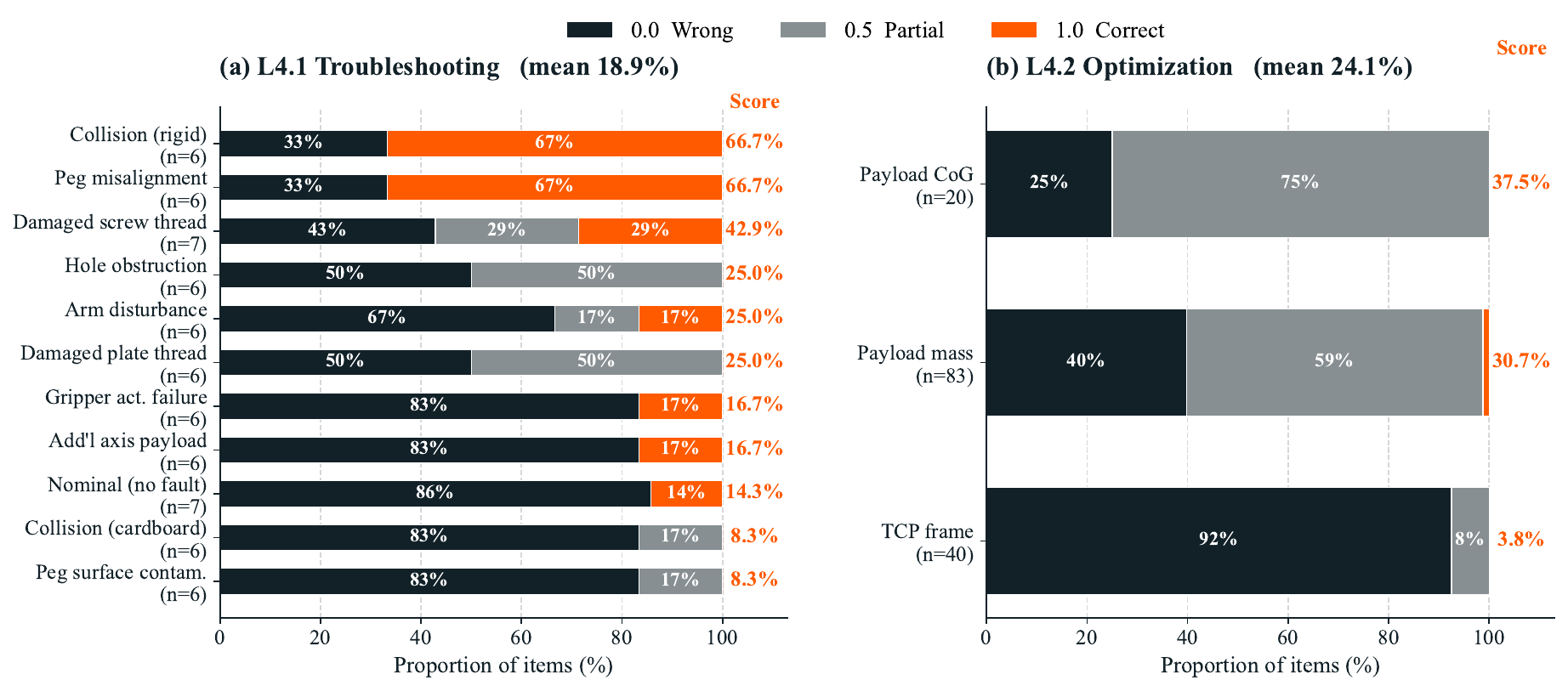}}
        \caption{GPT-5.1 L4 score distribution by ground-truth root cause (left, troubleshooting) and misconfigured parameter (right, optimization). Each bar shows the proportion of items scored 0/0.5/1; the right-edge value is the per-category mean accuracy.}
        \label{fig:results-gpt51-l4}
    \end{subfigure}
    \caption{Main results. \textbf{(a)} Cross-model panel performance across all four reasoning levels. \textbf{(b)} Per-fault and per-parameter breakdown of the GPT-5.1 L4 mass.}
    \label{fig:results}
\end{figure}

\subsection{Signal comprehension across Levels 1--3}

\paragraph{Most models fail to separate from the simple baseline on Level~1, including the largest in the panel.}
Under the signed chance correction the non-LLM baseline lands at $+4.3\%$ on L1 (a value we treat as the empirical zero). Four of six LLMs fail to clear it: Qwen3-4B ($-1.8\%$) and DeepSeek V3.2 ($-0.3\%$) score \emph{below chance} outright, and GPT-5.1 ($+3.2\%$) and Qwen3-235B ($+3.6\%$) sit below the baseline despite being the largest models in the panel by parameter-count and inference-cost proxies. Only Mistral Large 3 ($+5.9\%$) and Claude Sonnet 4.6 ($+11.8\%$) clear it, and Mistral only barely. Raw scale does not reliably confer the ability to extract precise state from dense industrial signals; L1 measures something general-purpose pretraining does not optimize for.

\paragraph{Models that struggle on Level~1 still outperform the baseline on Levels~2 and~3.}
DeepSeek V3.2 ($+5.8\%$, $+17.1\%$) and GPT-5.1 ($+6.9\%$, $+27.5\%$) both lift clear of the L2/L3 baseline values ($-0.2\%$, $+6.2\%$) despite trailing it on L1. That being said, the model rank order is nearly identical across L1--L3: Claude Sonnet 4.6 stays in the lead at L2 and L3 ($+14.6\%$, $+45.7\%$), GPT-5.1 is second on both, and Qwen3-235B ($+4.9\%$, $+20.6\%$) is the strongest open-weight model, with Mistral Large 3 just behind ($+5.3\%$, $+19.7\%$). This shows a strong correlation between the score of Levels~1--3, and points towards the fact that a stronger state comprehension translates into better intervention and counterfactual reasoning: L2 and L3 build on L1 ability rather than substituting for it. The L3 margins should be read with the caveat of Appendix~\ref{app:l3-tensor-caveat}: the trajectory-prediction template L3.5 awards the non-LLM baseline $+37.2\%$, which is not possible if its assumed chance level were correct, so part of every model's L3 figure reflects that template's scoring margin rather than counterfactual competence.

\paragraph{A human expert reaches near-ceiling on Levels~1--2, but not on Level~3.}
To establish what score is achievable on these items, a robotics expert answered a $60$-item sample under open-tool, no-LLM conditions, graded by the benchmark's own scorer. On raw scores over identical items the expert reaches $0.962$ at L1 and $0.980$ at L2, against $0.712$ and $0.662$ for the best model and $0.35$--$0.52$ for the rest. The L1--L2 gap is therefore not an artifact of ambiguous items: the information is in the released signals, and reading it out is what the panel fails at. Level~3 behaves differently: the expert reaches only $0.625$ and does not lead the panel, which we attribute to counterfactual prediction being poorly served by existing tooling rather than to defective items, so the L3 headroom should be read as smaller than at L1--L2. Two further artifacts are different evidence, not pooled with that run. A verified $102$-item expert solution set spanning all four levels ($25$/$42$/$23$/$12$) is released with a derivation and runnable code per item; it is audited ground truth, not a scored human attempt, so it yields no human accuracy and certifies no item beyond those $102$. Blind re-solving by an independent solver covers $20$ Level-1--2 items, reproducing $19$ stored answers (Appendix~\ref{app:human_baseline}).

\subsection{Engineering decision-making at Level 4}
\label{subsec:l4}

\paragraph{Level~4 exposes an industrial readiness gap.}
On Level~4 the leader reaches 21.5\% under the free-form rubric and the other five score between 5.6\% and 19.9\% (recall that L4 uses an uncorrected 0/0.5/1 rubric and its magnitudes are not directly comparable to the chance-corrected L1--L3 scores; the point of interest here is the rank-order within L4). Claude Sonnet 4.6, which leads L1--L3 and holds the panel's largest margin on L3, falls to fourth of six on L4 at 15.9\%. The model receives the time series and a machine description but must produce the root-cause diagnosis and multi-step recovery procedure from its own knowledge; the poor L4 performance indicates that even models with strong L1--L3 signal comprehension either lack the machine-specific technical knowledge to diagnose faults and prescribe remediation, or cannot ground that knowledge in the observed signals. Closing this gap is the central challenge FactoryBench is designed to measure. We probe this representation-versus-read-out question directly in Appendix~\ref{app:probing}: linear probes on a frozen open-weight panel member recover fault \emph{presence} from the model's activations well above chance ($0.83$) even where the model's own answer stays below chance (a \emph{read-out} failure), whereas fault \emph{type} is only weakly decodable and no better than an untrained-network baseline, pointing to a representational gap rather than merely a decoding one. Part of the L4 collapse is thus miscalibrated read-out, and part is missing representation.

\paragraph{GPT-5.1 leads on the most practically relevant level despite underperforming on the others.}
The L4 ordering reshuffles relative to L1--L3: GPT-5.1, second of six on the structured levels and far behind Claude Sonnet 4.6 there, leads L4 at 21.5\%, and Qwen3-235B ($19.9\%$) and DeepSeek V3.2 ($17.8\%$) also overtake Claude Sonnet 4.6 ($15.9\%$) despite trailing it by $25$ to $29$ points on L3. L4 requires naming the root cause and producing a multi-step remediation procedure grounded in manufacturer documentation; we conjecture that GPT-5.1's reversal reflects disproportionate exposure to structured industrial documentation in pretraining, letting it match a protocol to a fault description even when its signal comprehension is weak. The pattern reveals a dissociation between \emph{signal comprehension} and \emph{protocol retrieval}: optimizing for one does not deliver the other.

\paragraph{Within Level~4, GPT-5.1 shows asymmetric failure modes across troubleshooting and optimization.}
Breaking GPT-5.1's L4 performance down by question type (Figure~\ref{fig:results-gpt51-l4}) exposes a structural difference in \emph{how} the model fails. On troubleshooting ($n{=}143$, mean 0.189) the score distribution is sharply bimodal: 76.2\% zero, 14.0\% perfect, only 9.8\% partial. Because Lite draws root causes evenly, the perfects cannot be attributed to an over-represented class, and they separate cleanly by fault physics (Figure~\ref{fig:results-gpt51-l4}, left). GPT-5.1 scores highest on faults whose signature is strong and unmistakable: peg-insertion misalignment and rigid-object collision (both $0.67$), foam-object collision ($0.50$) and damaged screw thread ($0.43$), each of which writes a TCP force spike, a speed-scaling collapse or a large torque drift into the signal. It scores exactly zero on every fault that changes assembly \emph{state} without a mechanical transient: missing screw, extra assembly component, loosening phase, unstable mounting platform, gripper release during motion and self-collision link interference all sit at $0.00$. Nominal episodes are not the easy case they might appear: GPT-5.1 correctly returns ``no anomaly'' on only $14\%$ of them, so its errors run in both directions. Even when the model senses that something is off, it cannot recover the canonical root cause unless the fault announces itself as a force or velocity event.

On optimization ($n{=}143$, mean 0.241) GPT-5.1 earns a perfect score exactly once (1 perfect, 67 partial, 75 zero; Figure~\ref{fig:results-gpt51-l4}, right), and it is the only model in the panel to score a single perfect on this template at all: the other five return $0$ perfects out of $143$ each. Ground-truth optimization answers are single-parameter corrections (e.g.\ ``payload mass is set to 0.0\,kg but the actual payload weighs 1.5\,kg; update the installation settings''). GPT-5.1 instead emits lists of 8--12 generic motion-tuning suggestions (reduce acceleration, retune PID gains, add waypoints, lower speed scaling, enable force-mode compliance) and mentions payload configuration only as one undifferentiated item without specifying direction or magnitude, earning 0.5 at best (according to our grading policy). The failure is sharply parameter-dependent (Figure~\ref{fig:results-gpt51-l4}, right): GPT-5.1 reaches $0.31$ on payload mass and $0.38$ on payload centre of gravity, but only $0.04$ on TCP frame misconfiguration, where a wrong tool-centre offset perturbs the pose-conditioned response without changing any single channel's magnitude in a way the model reads as diagnostic. Together these patterns show that even though GPT-5.1 leads the panel on L4, it remains weak at decision-making in an industrial context: a template on which no model in the panel can reliably name the single misconfigured parameter is not one any of them has solved.

\subsection{Modular and extensible by design}

FactoryBench is built to grow. The Q\&A generator decouples reasoning templates from the underlying robot, task, and signal schema: every template is parameterized over generic primitives (signal names, phase indices, anomaly labels, fault categories) rather than hard-coded to a machine, so adding a new platform only requires populating the labeling ontology and per-robot signal mapping. All existing templates then instantiate automatically, and the format scorers, knowledge graph, and judge prompts carry over. Planned extensions include further robots and machine families (industrial arms, mobile manipulators, CNC and additive-manufacturing platforms), tasks (assembly variants, polishing, inspection), gripper and end-effector types, and templates targeting under-represented reasoning patterns. Each addition slots into the existing schema without breaking prior versions; the versioned release track pins every reported number to a fixed dataset state while the benchmark continues to grow. Licensing, the public/held-out split, and the exact release version underlying every number in this paper are documented in Appendix~\ref{app:license}.

\subsection{Limitations}

FactoryBench has three substantive limitations. First, Level~3 \emph{counterfactual} ground truth on physical hardware is fundamentally approximate and resource-intensive to generate: two real runs cannot share an identical pre-injection state, so our signature-kernel-MMD-selected CF pair is the closest empirical proxy for the do-operation rather than the do-operation itself, and L3 scores therefore conflate model reasoning with proxy quality (simulation closes the gap only at the cost of physical realism, a trade-off we quantify in Appendix~\ref{app:simulation_setup} by measuring the sim-to-real gap on matched episodes). Second, as the first benchmark of its sort our Q\&A pairs use a simplified fault model: industrial faults in production are typically compound, gradual, and detected through aggregate KPIs, while ours are atomic, fast, and drawn from a closed catalogue of 27 injected mechanisms, making FactoryBench a measurement of in-distribution fault \emph{recognition} rather than the fault \emph{detection} problem operators actually face. Third, Level~4 ground truth is drawn from a finite catalogue of manufacturer error codes and expert-authored recovery procedures, so a model is effectively evaluated on \emph{closed-book retrieval} of protocols it may or may not have seen during pretraining rather than on open-ended engineering reasoning; the low L4 scores we report therefore conflate ``does the model know this specific vendor manual?'' with ``can it reason from signals to a corrective action?'', and the two failure modes can only be separated by controlling for the model's prior exposure to the underlying documentation. As a first step toward separating representational absence from read-out failure, Appendix~\ref{app:probing} uses linear probing to show that fault presence is linearly encoded in an open-weight model's activations even where its answers collapse to chance, locating that part of the behavioural gap at the read-out rather than the representation; fault type shows neither a learned representation nor a usable read-out.

\section{Conclusion}

We introduced FactoryBench, a four-level benchmark for machine understanding over industrial robotic telemetry, grounded in FactoryWave and Pearl's causal hierarchy. Across our zero-shot panel of six frontier LLMs, signal comprehension and protocol-grounded decision-making do not co-develop: the ordering on Levels~1--3 reshuffles on Level~4, with GPT-5.1 taking the L4 lead despite trailing on the structured levels while the L1--L3 leader drops to fourth of six. Six models cannot prove the two never co-develop, but none of ours does. Structured-level scores stay well below 50\% chance-corrected and no model nears saturation. To test whether composition rather than scale closes the gap, we build a tool-augmented agent delegating quantitative subtasks to four tools (a forecaster, a numerical sandbox, signal statistics, and manual retrieval) \cite{schick2023toolformer,yao2023react,qin2023tool}. It does not close the gap: against zero-shot GPT-5.1 on matched items the agent is statistically level ($-2.9$ pp, CI $[-7.4, +1.6]$), and that average hides two opposing effects. The numerical sandbox is worth roughly $+20$ pp on the Level~1 templates that use it, while delegated forecasting costs $-10.5$ pp on Level~3, concentrated where the agent adopts the forecaster's point estimate verbatim (Appendix~\ref{app:agent-baseline}). The gap therefore survives tool access under the one delegation policy we test, which never asks whether the tool beats the caller; a confidence-gated policy might do better, and we have not built one. The gap is wide, not incidental: until models can read industrial signals, reason causally about them, and translate that reasoning into operator-grade decisions, they have no business on the factory floor. FactoryBench draws that line and gives the community the instrument to measure every step toward crossing it.

\newpage
\ificlrstatements\input{_iclr_statements}\fi

\bibliographystyle{\paperbibstyle}
\bibliography{references}

\begin{thebibliography}{10}

\bibitem{leporowski2022aursad}
B{\l}a{\.z}ej Leporowski, Daniella Tola, Casper Hansen, and Alexandros
  Iosifidis.
\newblock {AURSAD}: Universal robot screwdriving anomaly detection dataset.
\newblock {\em arXiv preprint arXiv:2102.01409}, 2021.

\bibitem{brockmann2023vorausad}
Jan~Thie{\ss} Brockmann, Marco Rudolph, Bodo Rosenhahn, and Bastian Wandt.
\newblock The {voraus-AD} dataset for anomaly detection in robot applications.
\newblock {\em IEEE Transactions on Robotics}, 40:438--451, 2024.
\newblock arXiv:2311.04765.

\bibitem{othman2026factorynetlargescaledatasetindustrial}
Karim Othman, Jonas Petersen, Matei Ignuta-Ciuncanu, Camilla Mazzoleni,
  Federico Martelli, Alessandro Lombardi, Riccardo Maggioni, and Philipp
  Petersen.
\newblock Factorynet: A large-scale dataset toward industrial time-series
  foundation models, 2026.

\bibitem{zhou2021informer}
Haoyi Zhou, Shanghang Zhang, Jieqi Peng, Shuai Zhang, Jianxin Li, Hui Xiong,
  and Wancai Zhang.
\newblock Informer: Beyond efficient transformer for long sequence time-series
  forecasting.
\newblock In {\em Proceedings of the AAAI conference on artificial
  intelligence}, volume~35, pages 11106--11115, 2021.

\bibitem{su2019omni}
Ya~Su, Youjian Zhao, Chenhao Niu, Rong Liu, Wei Sun, and Dan Pei.
\newblock Robust anomaly detection for multivariate time series through
  stochastic recurrent neural network.
\newblock In {\em Proceedings of the ACM SIGKDD International Conference on
  Knowledge Discovery and Data Mining}, 2019.

\bibitem{audibert2020usad}
Julien Audibert, Pietro Michiardi, Frédéric Guyard, Sébastien Marti, and
  Maria~A. Zuluaga.
\newblock {USAD}: Unsupervised anomaly detection on multivariate time series.
\newblock In {\em Proceedings of the ACM SIGKDD International Conference on
  Knowledge Discovery and Data Mining}, 2020.

\bibitem{chalapathy2019survey}
Raghavendra Chalapathy and Sanjay Chawla.
\newblock Deep learning for anomaly detection: A survey.
\newblock {\em arXiv preprint arXiv:1901.03407}, 2019.

\bibitem{lavin2015numenta}
Alexander Lavin and Subutai Ahmad.
\newblock Evaluating real-time anomaly detection algorithms---the {Numenta}
  anomaly benchmark.
\newblock In {\em Proceedings of the IEEE International Conference on Machine
  Learning and Applications (ICMLA)}, pages 38--44, 2015.

\bibitem{zeng2023transformers}
Ailing Zeng, Muxi Chen, Lei Zhang, and Qiang Xu.
\newblock Are transformers effective for time series forecasting?
\newblock In {\em Proceedings of the AAAI Conference on Artificial
  Intelligence}, 2023.

\bibitem{wei2022chain}
Jason Wei, Xuezhi Wang, Dale Schuurmans, Maarten Bosma, Brian Ichter, Fei Xia,
  Ed~Chi, Quoc Le, and Denny Zhou.
\newblock Chain-of-thought prompting elicits reasoning in large language
  models.
\newblock In {\em Advances in Neural Information Processing Systems (NeurIPS)},
  2022.

\bibitem{yao2023tree}
Shunyu Yao, Dian Yu, Jeffrey Zhao, Izhak Shafran, Thomas~L. Griffiths, Yuan
  Cao, and Karthik Narasimhan.
\newblock Tree of thoughts: Deliberate problem solving with large language
  models.
\newblock In {\em Advances in Neural Information Processing Systems (NeurIPS)},
  2023.

\bibitem{gruver2023llmtime}
Nate Gruver, Marc Finzi, Shikai Qiu, and Andrew~Gordon Wilson.
\newblock Large language models are zero-shot time series forecasters.
\newblock In {\em Advances in Neural Information Processing Systems (NeurIPS)},
  2023.

\bibitem{jin2024timellm}
Ming Jin, Shiyu Wang, Lintao Ma, Zhixuan Chu, James~Y. Zhang, Xiaoming Shi,
  Pin-Yu Chen, Yuxuan Liang, Yuan-Fang Li, Shirui Pan, and Qingsong Wen.
\newblock {Time-LLM}: Time series forecasting by reprogramming large language
  models.
\newblock In {\em Proceedings of the International Conference on Learning
  Representations (ICLR)}, 2024.

\bibitem{vaswani2017attention}
Ashish Vaswani, Noam Shazeer, Niki Parmar, Jakob Uszkoreit, Llion Jones,
  Aidan~N. Gomez, Łukasz Kaiser, and Illia Polosukhin.
\newblock Attention is all you need.
\newblock In {\em Advances in Neural Information Processing Systems (NeurIPS)},
  2017.

\bibitem{wu2021autoformer}
Haixu Wu, Jiehui Xu, Jianmin Wang, and Mingsheng Long.
\newblock {Autoformer}: Decomposition transformers with auto-correlation for
  long-term series forecasting.
\newblock In {\em Advances in Neural Information Processing Systems (NeurIPS)},
  2021.

\bibitem{zhou2022fedformer}
Tian Zhou, Ziqing Ma, Qingsong Wen, Xue Wang, Liang Sun, and Rong Jin.
\newblock Fedformer: Frequency enhanced decomposed transformer for long-term
  series forecasting.
\newblock In {\em International conference on machine learning}, pages
  27268--27286. PMLR, 2022.

\bibitem{nie2022time}
Yuqi Nie, Nam~H Nguyen, Phanwadee Sinthong, and Jayant Kalagnanam.
\newblock A time series is worth 64 words: Long-term forecasting with
  transformers.
\newblock {\em arXiv preprint arXiv:2211.14730}, 2022.

\bibitem{ansari2024chronos}
Abdul~Fatir Ansari, Lorenzo Stella, Caner Turkmen, Xiyuan Zhang, Pedro Mercado,
  Huibin Shen, Oleksandr Shchur, Syama~Sundar Rangapuram, Sebastian
  Pineda~Arango, Shubham Kapoor, Jasper Zschiegner, Danielle~C. Maddix, Hao
  Wang, Michael~W. Mahoney, Kari Torkkola, Andrew~Gordon Wilson, Michael
  Bohlke-Schneider, and Yuyang Wang.
\newblock {Chronos}: Learning the language of time series.
\newblock {\em Transactions on Machine Learning Research}, 2024.
\newblock arXiv:2403.07815.

\bibitem{das2024decoder}
Abhimanyu Das, Weihao Kong, Rajat Sen, and Yichen Zhou.
\newblock A decoder-only foundation model for time-series forecasting.
\newblock In {\em Proceedings of the International Conference on Machine
  Learning (ICML)}, 2024.
\newblock arXiv:2310.10688.

\bibitem{petersen2026hepaselfsupervisedhorizonconditionedevent}
Jonas Petersen, Gian-Alessandro Lombardi, Riccardo Maggioni, Camilla Mazzoleni,
  Federico Martelli, and Philipp Petersen.
\newblock Hepa: A self-supervised horizon-conditioned event predictive
  architecture for time series, 2026.

\bibitem{hendrycks2021mmlu}
Dan Hendrycks, Collin Burns, Steven Basart, Andy Zou, Mantas Mazeika, Dawn
  Song, and Jacob Steinhardt.
\newblock Measuring massive multitask language understanding.
\newblock In {\em Proceedings of the International Conference on Learning
  Representations (ICLR)}, 2021.

\bibitem{srivastava2023bigbench}
Aarohi Srivastava et~al.
\newblock Beyond the imitation game: Quantifying and extrapolating the
  capabilities of language models.
\newblock {\em Transactions on Machine Learning Research}, 2023.

\bibitem{yue2024mmmu}
Xiang Yue, Yuansheng Ni, Kai Zhang, Tianyu Zheng, Ruoqi Liu, Ge~Zhang, Samuel
  Stevens, Dongfu Jiang, Weiming Ren, Yuxuan Sun, et~al.
\newblock {MMMU}: A massive multi-discipline multimodal understanding and
  reasoning benchmark for expert {AGI}.
\newblock In {\em Proceedings of the IEEE/CVF Conference on Computer Vision and
  Pattern Recognition (CVPR)}, 2024.

\bibitem{cai2024timeseriesexamtimeseriesunderstanding}
Yifu Cai, Arjun Choudhry, Mononito Goswami, and Artur Dubrawski.
\newblock {TimeSeriesExam}: A time series understanding exam, 2024.

\bibitem{Xie_2025}
Zhe Xie, Zeyan Li, Xiao He, Longlong Xu, Xidao Wen, Tieying Zhang, Jianjun
  Chen, Rui Shi, and Dan Pei.
\newblock {ChatTS}: Aligning time series with {LLMs} via synthetic data for
  enhanced understanding and reasoning.
\newblock {\em Proceedings of the VLDB Endowment}, 18(8):2385--2398, 2025.

\bibitem{wang2025itformerbridgingtimeseries}
Yilin Wang, Peixuan Lei, Jie Song, Yuzhe Hao, Tao Chen, Yuxuan Zhang, Lei Jia,
  Yuanxiang Li, and Zhongyu Wei.
\newblock {ITFormer}: Bridging time series and natural language for multi-modal
  {QA} with large-scale multitask dataset, 2025.

\bibitem{jing2026tsaqatimeseriesanalysis}
Baoyu Jing, Sanhorn Chen, Lecheng Zheng, Boyu Liu, Zihao Li, Jiaru Zou, Tianxin
  Wei, Zhining Liu, Zhichen Zeng, Ruizhong Qiu, Xiao Lin, Yuchen Yan, Dongqi
  Fu, Jingchao Ni, Jingrui He, and Hanghang Tong.
\newblock {TSAQA}: Time series analysis question and answering benchmark, 2026.

\bibitem{kong2025timemqatimeseriesmultitask}
Yaxuan Kong, Yiyuan Yang, Yoontae Hwang, Wenjie Du, Stefan Zohren, Zhangyang
  Wang, Ming Jin, and Qingsong Wen.
\newblock {Time-MQA}: Time series multi-task question answering with context
  enhancement, 2025.

\bibitem{chen2026mtbenchmultimodaltimeseries}
Jialin Chen, Aosong Feng, Ziyu Zhao, Juan Garza, Gaukhar Nurbek, Cheng Qin, Ali
  Maatouk, Leandros Tassiulas, Yifeng Gao, and Rex Ying.
\newblock {MTBench}: A multimodal time series benchmark for temporal reasoning
  and question answering, 2026.

\bibitem{divo2025quants}
Felix Divo, Maurice Kraus, Anh~Q. Nguyen, Hao Xue, Imran Razzak, Flora~D.
  Salim, Kristian Kersting, and Devendra~Singh Dhami.
\newblock {QuAnTS}: Question answering on time series.
\newblock {\em arXiv preprint arXiv:2511.05124}, 2025.

\bibitem{schick2023toolformer}
Timo Schick, Jane Dwivedi-Yu, Roberto Dessì, Roberta Raileanu, Maria Lomeli,
  Eric Hambro, Luke Zettlemoyer, Nicola Cancedda, and Thomas Scialom.
\newblock Toolformer: Language models can teach themselves to use tools.
\newblock In {\em Advances in Neural Information Processing Systems (NeurIPS)},
  2023.

\bibitem{yao2023react}
Shunyu Yao, Jeffrey Zhao, Dian Yu, Nan Du, Izhak Shafran, Karthik Narasimhan,
  and Yuan Cao.
\newblock {ReAct}: Synergizing reasoning and acting in language models.
\newblock In {\em Proceedings of the International Conference on Learning
  Representations (ICLR)}, 2023.

\bibitem{qin2023tool}
Yujia Qin, Shengding Hu, Yankai Lin, Weize Chen, Ning Ding, Ganqu Cui, Zheni
  Zeng, Xuanhe Zhou, Yufei Huang, Chaojun Xiao, Chi Han, et~al.
\newblock Tool learning with foundation models.
\newblock {\em ACM Computing Surveys}, 2024.
\newblock arXiv:2304.08354.

\bibitem{lim2021tft}
Bryan Lim, Sercan~Ö. Arık, Nicolas Loeff, and Tomas Pfister.
\newblock Temporal fusion transformers for interpretable multi-horizon time
  series forecasting.
\newblock {\em International Journal of Forecasting}, 37(4):1748--1764, 2021.

\bibitem{oreshkin2020nbeats}
Boris~N. Oreshkin, Dmitri Carpov, Nicolas Chapados, and Yoshua Bengio.
\newblock {N-BEATS}: Neural basis expansion analysis for interpretable time
  series forecasting.
\newblock In {\em Proceedings of the International Conference on Learning
  Representations (ICLR)}, 2020.

\bibitem{ruff2018deep}
Lukas Ruff, Robert~A. Vandermeulen, Nico Görnitz, Lucas Deecke, Shoaib~Ahmed
  Siddiqui, Alexander Binder, Emmanuel Müller, and Marius Kloft.
\newblock Deep one-class classification.
\newblock In {\em Proceedings of the International Conference on Machine
  Learning (ICML)}, 2018.

\bibitem{pearl2009causality}
Judea Pearl.
\newblock {\em Causality: Models, Reasoning, and Inference}.
\newblock Cambridge University Press, 2nd edition, 2009.

\bibitem{peters2017elements}
Jonas Peters, Dominik Janzing, and Bernhard Sch{\"o}lkopf.
\newblock {\em Elements of Causal Inference}.
\newblock {MIT} Press, 2017.

\bibitem{rubin1974causal}
Donald~B. Rubin.
\newblock Estimating causal effects of treatments in randomized and
  nonrandomized studies.
\newblock {\em Journal of Educational Psychology}, 66(5):688--701, 1974.

\bibitem{granger1969causal}
Clive W.~J. Granger.
\newblock Investigating causal relations by econometric models and
  cross-spectral methods.
\newblock {\em Econometrica}, 37(3):424--438, 1969.

\bibitem{runge2019causal}
Jakob Runge, Peer Nowack, Marlene Kretschmer, Seth Flaxman, and Dino
  Sejdinovic.
\newblock Detecting and quantifying causal associations in large nonlinear time
  series datasets.
\newblock {\em Science Advances}, 5(11):eaau4996, 2019.

\bibitem{petersen2026structuralinductivebiashelps}
Jonas Petersen, Camilla Mazzoleni, Gian-Alessandro Lombardi, Federico Martelli,
  and Riccardo Maggioni.
\newblock What structural inductive bias helps transformers reason over
  knowledge graphs? a study with tabula rasa, 2026.

\bibitem{laptev2015generic}
Nikolay Laptev, Saeed Amizadeh, and Ian Flint.
\newblock Generic and scalable framework for automated time-series anomaly
  detection.
\newblock In {\em Proceedings of the ACM SIGKDD International Conference on
  Knowledge Discovery and Data Mining}, 2015.

\bibitem{dau2019ucr}
Hoang~Anh Dau, Anthony Bagnall, Kaveh Kamgar, Chin-Chia~Michael Yeh, Yan Zhu,
  Shaghayegh Gharghabi, Chotirat~Ann Ratanamahatana, and Eamonn Keogh.
\newblock The {UCR} time series archive.
\newblock {\em IEEE/CAA Journal of Automatica Sinica}, 6(6):1293--1305, 2019.

\bibitem{wen2022transformers}
Qingsong Wen, Tian Zhou, Chaoli Zhang, Weiqi Chen, Ziqing Ma, Junchi Yan, and
  Liang Sun.
\newblock Transformers in time series: A survey.
\newblock In {\em Proceedings of the Thirty-Second International Joint
  Conference on Artificial Intelligence (IJCAI)}, pages 6778--6786, 2023.
\newblock arXiv:2202.07125.

\bibitem{johnson2017clevr}
Justin Johnson, Bharath Hariharan, Laurens van~der Maaten, Li~Fei-Fei,
  C.~Lawrence Zitnick, and Ross Girshick.
\newblock {CLEVR}: A diagnostic dataset for compositional language and
  elementary visual reasoning.
\newblock In {\em Proceedings of the IEEE Conference on Computer Vision and
  Pattern Recognition (CVPR)}, 2017.
\newblock arXiv:1612.06890.

\bibitem{yi2020clevrer}
Kexin Yi, Chuang Gan, Yunzhu Li, Pushmeet Kohli, Jiajun Wu, Antonio Torralba,
  and Joshua~B. Tenenbaum.
\newblock {CLEVRER}: Collision events for video representation and reasoning.
\newblock In {\em Proceedings of the International Conference on Learning
  Representations (ICLR)}, 2020.
\newblock arXiv:1910.01442.

\bibitem{jin2023cladder}
Zhijing Jin, Yuen Chen, Felix Leeb, Luigi Gresele, Ojasv Kamal, Zhiheng Lyu,
  Kevin Blin, Fernando Gonzalez~Adauto, Max Kleiman-Weiner, Mrinmaya Sachan,
  and Bernhard Sch\"olkopf.
\newblock {CLadder}: Assessing causal reasoning in language models.
\newblock In {\em Advances in Neural Information Processing Systems (NeurIPS)},
  2023.

\bibitem{pawlowski2020deep_scm}
Nick Pawlowski, Daniel~C. Castro, and Ben Glocker.
\newblock Deep structural causal models for tractable counterfactual inference.
\newblock In {\em Advances in Neural Information Processing Systems (NeurIPS)},
  2020.

\bibitem{peyrard2020ladder}
Maxime Peyrard and Robert West.
\newblock A ladder of causal distances.
\newblock {\em arXiv preprint arXiv:2005.02480}, 2020.

\bibitem{anthropic2025sonnet46}
{Anthropic}.
\newblock {Claude Sonnet 4.6} system card, 2026.
\newblock Anthropic system card, February 2026.
  \url{https://www.anthropic.com/claude-sonnet-4-6-system-card}.

\bibitem{openai2025gpt51}
{OpenAI}.
\newblock {GPT-5.1} system card, 2025.
\newblock OpenAI system card addendum.
  \url{https://openai.com/index/gpt-5-system-card-addendum-gpt-5-1/}.

\bibitem{deepseek2025v32}
{DeepSeek-AI}.
\newblock {DeepSeek-V3.2}: Pushing the frontier of open large language models,
  2025.
\newblock arXiv:2512.02556. \url{https://arxiv.org/abs/2512.02556}.

\bibitem{mistral2025large3}
{Mistral AI}.
\newblock Introducing {Mistral 3}, 2025.
\newblock Mistral AI announcement. \url{https://mistral.ai/news/mistral-3}.

\bibitem{qwen2025qwen3}
{Qwen Team}.
\newblock {Qwen3} technical report, 2025.
\newblock arXiv:2505.09388.

\bibitem{alain2016probes}
Guillaume Alain and Yoshua Bengio.
\newblock Understanding intermediate layers using linear classifier probes.
\newblock {\em arXiv preprint arXiv:1610.01644}, 2016.

\bibitem{tas2025words}
{\"O}mer~{\c{S}}ahin Ta{\c{s}} and Royden Wagner.
\newblock Words in motion: Extracting interpretable control vectors for motion
  transformers.
\newblock In {\em International Conference on Learning Representations (ICLR)},
  2025.
\newblock arXiv:2406.11624.

\bibitem{hewitt2019control}
John Hewitt and Percy Liang.
\newblock Designing and interpreting probes with control tasks.
\newblock In {\em Proceedings of the 2019 Conference on Empirical Methods in
  Natural Language Processing (EMNLP)}, 2019.

\end{thebibliography}

%
%

\newpage
\appendix


\section{Answer formats and scoring}
\label{app:answer-formats}

FactoryBench emits questions in five answer formats. The four structured formats (single-select MCQ, multi-select MCQ, ranking, tensor) are scored deterministically by per-format rules; free-form answers, used only at Level~4, are scored by an LLM-as-judge voting protocol whose inter-judge reliability we analyse in Appendix~\ref{app:judge_reliability}. Each format is designed to remain hard to answer in the absence of correct reasoning over the time series. Table~\ref{tab:answer_formats} summarises the raw scoring rules; the rest of this section gives the implementation details together with the chance-correction step that places every format on a comparable scale before aggregation.

\begin{table}[htbp]
    \centering
    \caption{Answer formats used in FactoryBench and their raw scoring rules. The chance-correction step described in this section is applied on top of these rules before per-level aggregation. Free-form answers (Level~4 only) are scored by an LLM-as-judge voting protocol with three judges (GPT-5.1, Claude Sonnet 4.6, DeepSeek V3.2) aggregated by the median of the three votes.}
    \label{tab:answer_formats}
    \small
    \setlength{\tabcolsep}{4pt}
    \begin{tabular}{@{}l p{4.6cm} p{6.2cm}@{}}
        \toprule
        \textbf{Format}   & \textbf{Description}                                                                                                                 & \textbf{Raw scoring (before chance correction)}                                                                                                                                                                                                                                                                          \\
        \midrule
        Single-select MCQ & Three or four mutually exclusive options (A--C or A--D); the model selects one.                                                      & Exact match: $1$ if the parsed letter equals the gold letter, $0$ otherwise. Per-item chance level $E=1/k$ where $k\in\{3,4\}$ is the option count for that question.                                                                                                                                                    \\
        \addlinespace
        Multi-select MCQ  & Four independent statements (A--D) about the outcome of an event or intervention; the model emits a T/F string of length four.       & Positional fraction: $1/4$ per slot whose T/F matches the gold; raw score in $\{0, 0.25, 0.5, 0.75, 1\}$. Chance level $E=1/2$.                                                                                                                                                                                          \\
        \addlinespace
        Ranking           & Four time-series segments or outcomes (A--D) ordered by a specified criterion. Answer is a permutation string (e.g.\ \texttt{DABC}). & Positional fraction: $1/4$ per slot whose letter matches the gold permutation; raw score in $\{0, 0.25, 0.5, 0.75, 1\}$. Chance level $E=1/4$.                                                                                                                                                                           \\
        \addlinespace
        Tensor            & One or more scalar values (e.g.\ expected sensor reading after an intervention). Answer is a list of floats.                         & Per-scalar three-level piecewise score: $1$ if $|\hat{v}-v^{\star}|\!\leq\! m$, $0.5$ if $m\!<\!|\hat{v}-v^{\star}|\!\leq\!2m$, else $0$. Per-channel margin calibrated to $m_j = R_j/12$ so the chance level matches MCQ ($E = 1/4$).                                                                                   \\
        \addlinespace
        Free-form         & Open-ended natural-language response, used only for the two Level~4 question types: troubleshooting and optimization.                & Score $\in\{0, 0.5, 1\}$ by LLM-as-judge against the reference answer: $1.0$ if semantically equivalent to the reference, $0.5$ if partially correct or imprecise but on-topic, $0.0$ if incorrect, irrelevant, or refusing. The verbatim judge prompt is given in Appendix~\ref{app:answer-formats}. \\
        \bottomrule
    \end{tabular}
\end{table}

\paragraph{Chance correction.} Without correction, formats with different chance baselines are not directly comparable: a model scoring $0.5$ on a multi-select item (chance $E=1/2$) has demonstrated nothing, while a model scoring $0.5$ on a single-select item (chance $E=1/4$) has demonstrated above-chance performance. We therefore transform every raw item score $s$ into a chance-corrected score
\begin{equation*}
    \tilde{s} \;=\; \frac{s - E}{1 - E},
\end{equation*}
where $E$ is the format's chance level. By linearity of expectation, uniform random guessing produces $\mathbb{E}[\tilde{s}] = 0$ for every format by construction; $\tilde{s} = 1$ corresponds to a perfect answer. The transform is signed: items scoring below chance receive $\tilde{s} < 0$, so a model that is worse than uniform random is visibly penalised in the aggregate mean rather than hidden by a floor. The per-format floor is $\tilde{s}_{\min} = -E/(1-E)$, which ranges from $-1$ for multi-select ($E=1/2$) to $-1/3$ for tensor / ranking / four-way MCQ ($E=1/4$); aggregate means below zero should therefore be interpreted with the level's format mix in mind. The non-LLM baseline used in Section~\ref{sec:zero-shot} doubles as the empirical audit: after chance correction it sits close to $0$ on Levels~1 and~2 ($+4.3\%$ and $-0.2\%$) and at $+6.2\%$ on Level~3, where the residual traces to template L3.5's over-generous margin rather than to reasoning (discussed below). We check this before treating any model score as evidence of reasoning.

\paragraph{Single-select MCQ.} Raw score $s\in\{0,1\}$. The chance level is read off the question payload as $E=1/k$. The corrected score is $\tilde{s} = (ks - 1)/(k-1)$, mapping a correct answer to $\tilde{s} = 1$ and a wrong answer to $\tilde{s} = -1/(k-1)$ (i.e. $-1/2$ for $k=3$, $-1/3$ for $k=4$).

\paragraph{Multi-select MCQ.} Raw score $s \in \{0, 0.25, 0.5, 0.75, 1\}$ with chance level $E = 1/2$ (each T/F slot matches the gold with probability $1/2$ under uniform random guessing). The corrected score is $\tilde{s} = 2s - 1$, mapping $\{0, 0.25, 0.5, 0.75, 1\}$ to $\{-1, -0.5, 0, 0.5, 1\}$.

\paragraph{Ranking.} Raw score is the positional-match fraction over a permutation of length four. The expected number of fixed points of a uniformly random permutation is exactly $1$ regardless of length, so the chance level is $E = 1/n = 1/4$. The corrected score is $\tilde{s} = (4s - 1)/3$, mapping $\{0, 0.25, 0.5, 0.75, 1\}$ to $\{-1/3, 0, 1/3, 2/3, 1\}$.

\paragraph{Tensor.} Each scalar $j$ in a vector answer is scored on a three-level scale: $1$ if $|\hat{v}_j - v_j^{\star}| \leq m_j$, $0.5$ if within $2 m_j$, $0$ otherwise. The raw item score is the average over the $n$ scalars, $s = \tfrac{1}{n}\sum_j s_j$. The discrete form mirrors MCQ/ranking and still credits near-misses inside $2 m_j$. The per-channel margin is calibrated to $m_j = R_j / 12$, where $R_j$ is the channel's empirical $[p_2, p_{98}]$ range across the unified episodes (precomputed once); this sets the piecewise scorer's chance expectation to $E = 1/4$, matching single-select MCQ. The corrected item score follows the MCQ form, $\tilde{s} = (4s - 1)/3$. Scalars with a degenerate channel range ($R_j \approx 0$) are dropped from the average.

\paragraph{Free-form.} Free-form answers do not admit a meaningful chance baseline: the answer space is unbounded natural language, so there is no uniform random distribution to subtract. We therefore leave free-form scores uncorrected. The lack of correction does not introduce cross-format bias because free-form is used \emph{only} at Level~4, and Level~4 contains no other answer format, so free-form scores are never aggregated alongside structured-format scores at the same level. As a consequence, Level~4 scores and Levels~1--3 scores are reported on different underlying scales (an uncorrected 0/0.5/1 rubric versus a chance-corrected structured metric) and are \emph{not directly comparable in magnitude}; when we contrast them in Section~\ref{sec:zero-shot} we refer to rank-order and gap-to-baseline effects rather than absolute-difference claims.

Every free-form answer is scored by the LLM judge against the gold reference answer (assembled from the FactoryBench knowledge graph) using the prompt below, reproduced \emph{verbatim} from the scoring code and applied identically to both Level~4 question types (troubleshooting and optimization). The placeholders \texttt{\{question\}}, \texttt{\{reference\}}, and \texttt{\{prediction\}} are filled per item:

{\footnotesize\begin{verbatim}
You are an expert evaluator for a robotics sensor-data Q&A benchmark.
Score a model's free-form answer against a reference answer.

Scoring:
  1.0 (Correct):  semantically equivalent to the reference: it gives the correct
                  root cause / misconfigured parameter AND a plausible corrective
                  step (exact wording of the fix is NOT required).
  0.5 (Neutral):  on-topic AND it identifies the correct root cause / misconfigured
                  parameter, even if imprecise or without the corrective step.
  0.0 (Wrong):    incorrect, irrelevant, names the WRONG root cause, is a generic
                  list that MISSES the specific parameter, claims a fault when the
                  reference says the machine is normal (or vice versa), or refuses
                  to answer.

Being merely "on-topic" is NOT enough for 0.5: the answer must identify the actual
root cause / parameter from the reference. A generic list of tuning suggestions
that does not single out the reference's specific cause/parameter is 0.0.

Respond with ONLY a JSON object on a single line, no markdown:
{"score": <0.0|0.5|1.0>, "reason": "<one short sentence>"}

Question:
{question}

Reference answer:
{reference}

Model answer:
{prediction}
\end{verbatim}}

\noindent A held-out audit confirms the panel applies this rubric consistently: an independent grader (outside the three-judge panel) given this exact prompt matches the judge-median on $95\%$ of a representative $100$-item sample (quadratic-weighted $\kappa=0.79$).

\paragraph{Aggregation.} Per-question chance-corrected scores $\tilde{s}$ are averaged uniformly within a template and level.

\subsection{A caveat on the tensor chance level at L3.5}
\label{app:l3-tensor-caveat}

The chance level for the tensor and numerical formats is not measured but
derived: the three-level piecewise scorer with per-channel margin
$m_j = R_j/12$ is constructed to give $E = 3m/R = 1/4$ under uniform guessing.
That derivation assumes the guess is uniform over the channel's range. It is a
poor description of a forecasting template, where the quantity being predicted
is strongly autocorrelated and a constant-continuation guess is far better than
uniform.

The non-LLM baseline makes the gap visible. On L3.5 (trajectory prediction) it
scores $+37.2\%$ chance-corrected, which is not a possible outcome for a method
with no access to the answer if $E = 1/4$ were the operative chance level for
that template; the same baseline scores $-17.8\%$ on L3.4 and $-0.2\%$ on the L2
aggregate. The natural reading is that $m_j = R_j/12$ is too generous for L3.5
specifically, so a prediction that simply continues the recent trajectory lands
inside the margin more often than the calibration assumes.

Two consequences. First, every model's L3 aggregate is inflated by whatever
share L3.5 contributes, and cross-level comparisons involving L3 should be read
with that in mind; the ordering \emph{within} L3 is unaffected, since all models
face the same margin. Second, L3.4 is not implicated: the baseline is well below
chance there while models reach $+19.7\%$ to $+63.4\%$, so that template's
separation is real. We report the numbers as measured rather than recalibrating
$m_j$ post hoc, since changing the margin would silently redefine the metric and
break comparability with previously published FactoryBench results; the baseline
row is included in Figure~\ref{fig:results} precisely so a reader can see the
size of the effect.

\subsection{Inter-judge reliability of the Level-4 panel}
\label{app:judge_reliability}

Because free-form Level-4 answers are scored by three judges (GPT-5.1, Claude Sonnet 4.6, DeepSeek V3.2) aggregated by median (Table~\ref{tab:answer_formats}), the L4 scores inherit whatever (dis)agreement those judges carry. We therefore quantify their reliability. We pool every Level-4 answer for which all three judges returned a valid verdict across the six-model evaluee panel ($n{=}6{,}617$ answers, three verdicts each in $\{0,0.5,1\}$); inter-rater agreement is a property of the judges, so pooling across evaluees is the appropriate unit of analysis.

\begin{figure}[htbp]
    \centering
    \includegraphics[width=\linewidth]{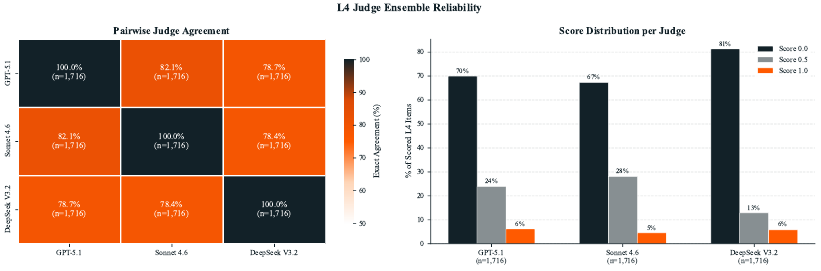}
    \caption{\textbf{The three-judge Level-4 panel is reliable.} (a)~Pairwise exact agreement ($91.3\%$, $93.4\%$, $89.7\%$; mean $91.5\%$) and quadratic-weighted $\kappa$ for each judge pair, with the $3$-way Fleiss $\kappa{=}0.60$ and Krippendorff $\alpha{=}0.78$ as reference. (b)~Consensus across the three judges: all three agree on $87.5\%$ of items, $11.9\%$ are $2$-of-$3$ majorities, and only $0.6\%$ are three-way splits. Pooled over $6{,}617$ Level-4 answers with three valid verdicts.}
    \label{fig:judge_agreement}
\end{figure}

\paragraph{The panel is reliable.} Agreement is substantial-to-good: Fleiss' $\kappa=0.60$ and interval Krippendorff's $\alpha=0.78$ across the three raters, with $87.5\%$ of items unanimous, $11.9\%$ carrying a $2$-of-$3$ majority, and only $0.6\%$ three-way splits (Figure~\ref{fig:judge_agreement}b). Pairwise, exact agreement is $89.7$--$93.4\%$ (mean $91.5\%$) and the quadratic-weighted $\kappa$ is $0.74$--$0.81$ (Figure~\ref{fig:judge_agreement}a), with GPT-5.1 and DeepSeek agreeing most and Claude and DeepSeek least. The pairwise confusion matrices are strongly diagonal and disagreements are almost never polar ($0\leftrightarrow1$): they are overwhelmingly between adjacent grades ($0\leftrightarrow0.5$ or $0.5\leftrightarrow1$), consistent with the ordinal rubric.

\paragraph{A small leniency spread justifies median aggregation.} The judges differ only mildly in strictness: mean verdict $0.101$ (Claude), $0.082$ (GPT-5.1), $0.066$ (DeepSeek), so Claude is the most lenient and DeepSeek the strictest. This directional bias is visible in the off-diagonals of the confusion matrices (e.g.\ Claude assigns $0.5$ where DeepSeek assigns $0$ on $8\%$ of items), and it is exactly the regime in which the \emph{median} is preferable to the mean: it is robust to a single lenient or strict judge and returns the majority verdict whenever one exists, which it does for $99.4\%$ of items. High agreement together with the robustness of median aggregation indicates that the reported Level-4 scores are not an artefact of any single judge.

\section{Released Q\&A distribution}
\label{app:qa_distribution}

\begin{figure}[htbp]
    \centering
    \includegraphics[width=\linewidth]{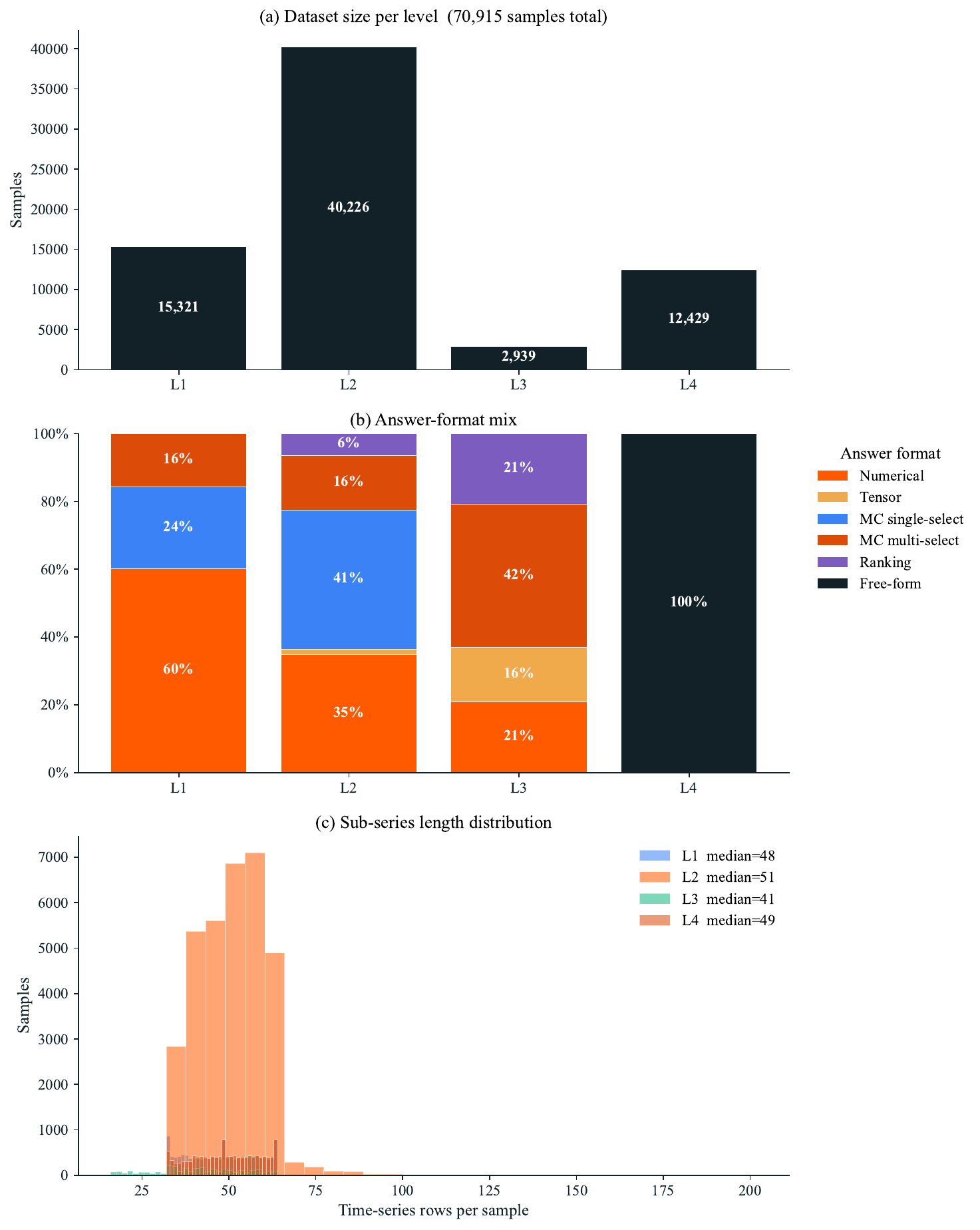}
    \caption{Distribution of the 69{,}691 released Q\&A items. \textbf{(a)} Level sizes. \textbf{(b)} Answer-format mix per level. \textbf{(c)} Sub-series length (number of time-series rows) per item.}
    \label{fig:qa_overview}
\end{figure}

Figure~\ref{fig:qa_overview} summarises how the released 69{,}691 Q\&A items are distributed across levels, answer formats, and time-series context lengths. Three patterns are worth flagging.

\paragraph{Level sizes are uneven by design.} L2 (40{,}226 items) is by far the largest level and L3 (2{,}939 items) is the smallest. The asymmetry follows directly from the underlying episode pool. Most FactoryWave episodes are anomalous (we deliberately oversample fault scenarios so that diagnostic templates have enough material), and L2 templates can be instantiated against any anomalous episode, which inflates L2's count. Counterfactual templates (L3) by contrast require paired (baseline, fault) runs collected with the signature-kernel-MMD protocol of Section~\ref{sec:density}, which is far more expensive to produce per episode; the released L3 corpus is therefore the slice of items for which a high-quality CF pair exists, and is intrinsically small.

\paragraph{Answer-format mix tracks the reasoning level.} L1 leans on numerical (69\,\%) and single-select MCQ (28\,\%) formats; L2 spreads across single-select MCQ (48\,\%), numerical (41\,\%), and multi-select (7\,\%), with small ranking and tensor slices (2\,\% each); L3 is the most format-diverse level, splitting across multi-select (42\,\%), numerical (21\,\%), ranking (20\,\%), and tensor (17\,\%), reflecting the counterfactual ranking and trajectory-prediction templates; L4 is entirely free-form because every L4 template asks for natural-language remediation or optimisation answers. Free-form appears only at L4: no L1--L3 template asks for one.

\paragraph{Sub-series lengths are tightly bounded.} The displayed time-series window is between 32 and 77 timesteps (>64 is possible for ranking since multiple time series) for almost every released item (median $\approx$50 across all four levels), keeping the prompt within the context budget of every evaluated model.

\subsection{FactoryBench-Lite}
\label{app:lite}

Evaluating a full panel on all 69{,}691 items is expensive, and the level
aggregates it produces are weighted by how many items each template happens to
contribute, which is a property of the episode pool rather than of the
benchmark's design (Appendix~\ref{app:qa_distribution}). We therefore also
release \textbf{FactoryBench-Lite}, a 3{,}000-item subset intended as the
default entry point for cheap evaluation and for reporting that is not
distorted by template mass.

Lite is not a random draw. It is balanced twice over: across templates, so no
single template dominates a headline number, and \emph{within} each template,
on whatever dimension actually determines that template's answer. The second
is the substantive one. A template can be balanced overall and still be
lopsided in the way that matters: the anomaly-identification templates can be
even across option letters while over-representing one fault class, and the
Level-4 optimisation template, which is free-form and so has no letters to
balance at all, can be sampled uniformly while still asking about payload mass
far more often than about TCP frame. Each template therefore
declares a stratum key -- the answer string for multi-select and ranking
templates, the correct class for anomaly, robot, and root-cause templates, the
misconfigured parameter for optimisation, and the source episode's (fault,
task) pair for numerical templates -- and selection round-robins over strata
rather than sampling uniformly.

All $21$ templates are represented, at $142$--$143$ items each ($572$ L1,
$1{,}430$ L2, $712$ L3, $286$ L4). On the six multi-select templates every one
of the four propositions sits at or near $50\,\%$ true, in the range
$[0.434,\,0.503]$ and within $0.007$ of parity on five of the six; the residual
is L1.3, where the proposition in question is genuinely scarce in the source
corpus. The $30$ anomaly classes of L2.7 appear $4$--$5$ times each, the $7$
optimisation variants of L4.2 appear $20$--$21$ times each, the $24$ Level-4
root-cause categories appear $5$--$6$ times each, and the robot-identification
templates are even to within one item per robot. Where a stratum is genuinely
scarce, round-robin takes what exists rather than over-sampling a thin class,
so the achieved distribution is reported rather than assumed.

Lite is distributed alongside the full set in the same Hugging Face repository,
with each item carrying its original identifier so that Lite results can be
joined back to full-set results item-by-item.

\subsection{Question-template inventory}
\label{app:templates}
Table~\ref{tab:template_inventory} lists the number of question templates currently producing released Q\&A items, broken down by answer format. Templates that are defined in the generator but do not yet emit any questions in the released set are excluded.

\begin{figure}[htbp]
    \centering
    \resizebox{\textwidth}{!}{%
        \begin{tikzpicture}[
                every node/.style={font=\sffamily},
            ]

            \def\panelw{8.4}
            \def\panelh{5.0}
            \def\panelhgap{0.35}
            \def\panelvgap{0.35}

            \coordinate (P1center) at (-\panelw/2-\panelhgap/2, 0);
            \coordinate (P2center) at ( \panelw/2+\panelhgap/2, 0);
            \coordinate (P3center) at (-\panelw/2-\panelhgap/2, -\panelh-\panelvgap);
            \coordinate (P4center) at ( \panelw/2+\panelhgap/2, -\panelh-\panelvgap);

            \newcommand{\drawminirobot}[4]{%
                \ifnum#3=1
                    \begin{scope}[transparency group, opacity=0.32]
                        \node[anchor=center, inner sep=0pt] at (#1, #2)
                        {\includegraphics[width=2.0cm]{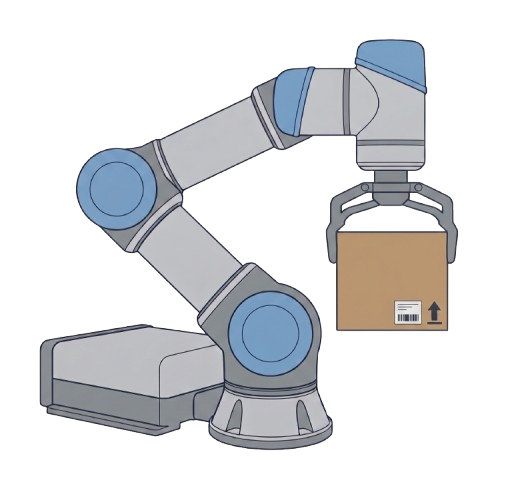}};
                    \end{scope}
                \else
                    \node[anchor=center, inner sep=0pt] at (#1, #2)
                    {\includegraphics[width=2.0cm]{figures/ur_robot_figure.png}};
                \fi
            }

            \pgfplotsset{
                pearlplot/.style={
                        width=4.5cm, height=2.8cm,
                        xmin=0, xmax=500, ymin=0, ymax=10,
                        xlabel={Time (ms)}, ylabel={Signal},
                        xlabel style={font=\sffamily\scriptsize, darknavy, yshift=2pt},
                        ylabel style={font=\sffamily\scriptsize, darknavy, yshift=-7pt},
                        tick label style={font=\sffamily\scriptsize, darknavy!85},
                        xtick={0,250,500}, ytick={0,5,10},
                        tick style={darknavy!60, line width=0.3pt},
                        axis lines=left,
                        axis line style={darknavy!70, line width=0.4pt},
                        ymajorgrids, xmajorgrids,
                        major grid style={darknavy!10, line width=0.25pt},
                        legend style={at={(1.0,1.6)}, anchor=north east, font=\sffamily\scriptsize,
                                draw=darknavy!20, line width=0.25pt,
                                fill=white, fill opacity=0.92, text opacity=1,
                                row sep=0pt, legend columns=1, inner sep=2pt},
                        clip=true,
                        axis background/.style={fill=white},
                    }
            }

            \begin{scope}[shift={(P1center)}]
                \fill[panelbg, rounded corners=4pt] (-\panelw/2, -\panelh/2) rectangle (\panelw/2, \panelh/2);
                \draw[L1tint!85!darknavy, line width=1.0pt, rounded corners=4pt] (-\panelw/2, -\panelh/2) rectangle (\panelw/2, \panelh/2);

                \filldraw[L1tint] (-\panelw/2+0.34, \panelh/2-0.30) circle (0.20);
                \node[font=\sffamily\bfseries\tiny, white] at (-\panelw/2+0.34, \panelh/2-0.30) {L1};
                \node[anchor=west, font=\sffamily\bfseries\small, L1tint!80!darknavy] at (-\panelw/2+0.62, \panelh/2-0.30) {State};
                \node[anchor=east, font=\sffamily\scriptsize, text=darknavy!85] at ( \panelw/2-0.30, \panelh/2-0.30) {$P(E \mid S, C)$};
                \draw[L1tint!25, line width=0.35pt] (-\panelw/2+0.20, \panelh/2-0.62) -- (\panelw/2-0.20, \panelh/2-0.62);

                \drawminirobot{-3.0}{-0.4}{0}{L1tint}

                \draw[figgreen, line width=1.4pt] (-1.85, 0.55) -- (-1.45, 0.55);
                \node[anchor=west, font=\sffamily\scriptsize, darknavy] at (-1.40, 0.55) {$S$};
                \draw[forgisorange, line width=1.4pt] (-1.85, 0.15) -- (-1.45, 0.15);
                \node[anchor=west, font=\sffamily\scriptsize, darknavy] at (-1.40, 0.15) {$E$};
                \draw[figcontext, line width=1.4pt] (-1.85, -0.25) -- (-1.45, -0.25);
                \node[anchor=west, font=\sffamily\scriptsize, darknavy] at (-1.40, -0.25) {$C$};

                \begin{scope}[shift={(0.9, -0.05)}]
                    \begin{axis}[pearlplot]
                        \addplot[figgreen, line width=1pt, smooth] coordinates { (0,4) (100,4.05) (200,4) (300,4.05) (400,4) (500,4) };
                        \addplot[forgisorange, line width=1pt, smooth] coordinates { (0,3.9) (100,4) (200,4.2) (250,5.5) (300,6.8) (350,7.2) (400,7) (500,6.2) };
                        \addplot[figcontext, line width=1pt, smooth] coordinates { (0,2.6) (100,2.6) (200,2.6) (300,2.6) (400,2.6) (500,2.6) };
                    \end{axis}
                \end{scope}

                \node[anchor=south, font=\sffamily\scriptsize\itshape, darknavy, text width=7.8cm, align=center]
                at (0, -\panelh/2+0.45) {``What is the current of joint 3?''};
                \node[anchor=south, font=\sffamily\tiny, figsteel, text width=7.8cm, align=center]
                at (0, -\panelh/2+0.12) {Reading and interpreting current sensor values from passive observation.};
            \end{scope}

            \begin{scope}[shift={(P2center)}]
                \fill[panelbg, rounded corners=4pt] (-\panelw/2, -\panelh/2) rectangle (\panelw/2, \panelh/2);
                \draw[L2tint!85!darknavy, line width=1.0pt, rounded corners=4pt] (-\panelw/2, -\panelh/2) rectangle (\panelw/2, \panelh/2);

                \filldraw[L2tint] (-\panelw/2+0.34, \panelh/2-0.30) circle (0.20);
                \node[font=\sffamily\bfseries\tiny, white] at (-\panelw/2+0.34, \panelh/2-0.30) {L2};
                \node[anchor=west, font=\sffamily\bfseries\small, L2tint!75!darknavy] at (-\panelw/2+0.62, \panelh/2-0.30) {Intervention};
                \node[anchor=east, font=\sffamily\scriptsize, text=darknavy!85] at ( \panelw/2-0.30, \panelh/2-0.30) {$P(E_{t>0} \mid \mathrm{do}(C))$};
                \draw[L2tint!30, line width=0.35pt] (-\panelw/2+0.20, \panelh/2-0.62) -- (\panelw/2-0.20, \panelh/2-0.62);

                \drawminirobot{-3.0}{-0.4}{0}{L2tint}

                \draw[L2tint, line width=0.6pt, fill=white, rounded corners=1pt]
                (-4.10, 1.00) rectangle (-2.0, 1.55);
                \node[font=\sffamily\scriptsize\bfseries, L2tint!80!darknavy, anchor=center]
                at (-3, 1.38) {do(C)};
                \node[font=\sffamily\tiny\itshape, L2tint!80!darknavy, anchor=center]
                at (-3, 1.13) {Force Perturbation};

                \draw[L2tint, line width=1.2pt, line cap=round,
                        -{Stealth[length=6pt, width=5pt]}]
                (-3, 1.00) .. controls (-3, 0.45) and (-2.55, 0.35) .. (-2.55, -0.15);

                \draw[figcontext, line width=1.4pt] (-1.85, 0.35) -- (-1.45, 0.35);
                \node[anchor=west, font=\sffamily\scriptsize, darknavy] at (-1.40, 0.35) {$C$ \,(forced)};
                \draw[forgisorange, line width=1.4pt] (-1.85, -0.05) -- (-1.45, -0.05);
                \node[anchor=west, font=\sffamily\scriptsize, darknavy] at (-1.40, -0.05) {$E$};

                \begin{scope}[shift={(0.9, -0.05)}]
                    \begin{axis}[pearlplot]
                        \fill[L2tint!10] (axis cs:160,0) rectangle (axis cs:500,10);

                        \fill[forgisorange!30, draw=none]
                        (axis cs:160, 3.9) -- (axis cs:200, 4.95) -- (axis cs:250, 6.10) --
                        (axis cs:300, 6.60) -- (axis cs:400, 6.65) -- (axis cs:500, 6.75) --
                        (axis cs:500, 4.65) -- (axis cs:400, 4.95) -- (axis cs:300, 5.40) --
                        (axis cs:250, 5.10) -- (axis cs:200, 4.40) -- cycle;

                        \node[anchor=center, font=\sffamily\tiny\bfseries, L2tint] at (axis cs:330, 8.5) {Future Prediction};

                        \draw[darknavy, line width=0.5pt, dashed] (axis cs:160,0) -- (axis cs:160,9.5);
                        \node[anchor=south west, font=\sffamily\tiny, darknavy] at (axis cs:50, 0.5) {Now};

                        \addplot[figcontext, line width=1pt] coordinates { (0,4) (160,4) (175,4.8) (300,4.8) (400,4.8) (500,4.8) };
                        \addplot[forgisorange, line width=1pt, smooth] coordinates { (0,3.9) (160,3.9) (250,5.6) (300,6) (400,5.8) (500,5.7) };
                    \end{axis}
                \end{scope}

                \node[anchor=south, font=\sffamily\scriptsize\itshape, darknavy, text width=7.8cm, align=center]
                at (0, -\panelh/2+0.45) {``The robot is exhibiting unexpected\_payload. What would most likely happen to the joint torques next?''};
                \node[anchor=south, font=\sffamily\tiny, figsteel, text width=7.8cm, align=center]
                at (0, -\panelh/2+0.12) {Predicting the future consequences of a physical perturbation happening now.};
            \end{scope}

            \begin{scope}[shift={(P3center)}]
                \fill[panelbg, rounded corners=4pt] (-\panelw/2, -\panelh/2) rectangle (\panelw/2, \panelh/2);
                \draw[L3tint!85!darknavy, line width=1.0pt, rounded corners=4pt] (-\panelw/2, -\panelh/2) rectangle (\panelw/2, \panelh/2);

                \filldraw[L3tint] (-\panelw/2+0.34, \panelh/2-0.30) circle (0.20);
                \node[font=\sffamily\bfseries\tiny, white] at (-\panelw/2+0.34, \panelh/2-0.30) {L3};
                \node[anchor=west, font=\sffamily\bfseries\small, L3tint!80!darknavy] at (-\panelw/2+0.62, \panelh/2-0.30) {Counterfactual};
                \node[anchor=east, font=\sffamily\scriptsize, text=darknavy!85] at ( \panelw/2-0.30, \panelh/2-0.30) {$P(E_{C=x} \mid S{=}s, E{=}e)$};
                \draw[L3tint!25, line width=0.35pt] (-\panelw/2+0.20, \panelh/2-0.62) -- (\panelw/2-0.20, \panelh/2-0.62);

                \drawminirobot{-3.1}{-0.30}{0}{L3tint}

                \fill[L3tint!80!darknavy, opacity=0.6] (-2.35, 0.25) circle (0.045);
                \fill[L3tint!80!darknavy, opacity=0.7] (-2.20, 0.50) circle (0.060);
                \fill[L3tint!80!darknavy, opacity=0.8] (-2.05, 0.78) circle (0.075);

                \fill[white] (-1.55, 1.05) ellipse (1.10 and 0.50);
                \draw[L3tint!80!darknavy, line width=0.55pt, dashed] (-1.55, 1.05) ellipse (1.10 and 0.50);

                \begin{scope}[transparency group, opacity=0.45]
                    \node[anchor=center, inner sep=0pt] at (-2.1, 1.05)
                    {\includegraphics[width=0.75cm]{figures/ur_robot_figure.png}};
                \end{scope}
                \node[font=\sffamily\scriptsize\bfseries, red!60!darknavy, anchor=west] at (-1.80, 1.15) {STALL!};
                \node[font=\sffamily\scriptsize\itshape, L3tint!80!darknavy, anchor=west] at (-1.80, 0.90) {(if forced)};

                \draw[figgreen, line width=1.4pt] (-1.85, 0.30) -- (-1.45, 0.30);
                \node[anchor=west, font=\sffamily\scriptsize, darknavy] at (-1.40, 0.30) {$S$};
                \draw[forgisorange, line width=1.4pt] (-1.85, -0.05) -- (-1.45, -0.05);
                \node[anchor=west, font=\sffamily\scriptsize, darknavy] at (-1.40, -0.05) {$E$ \,(factual)};
                \draw[forgisorange, line width=1.4pt, dashed] (-1.85, -0.40) -- (-1.45, -0.40);
                \node[anchor=west, font=\sffamily\scriptsize, darknavy] at (-1.40, -0.40) {$E$ \,(CF)};

                \begin{scope}[shift={(0.9, -0.05)}]
                    \begin{axis}[pearlplot]
                        \fill[forgisorange!30, draw=none]
                        (axis cs:20, 3.9) -- (axis cs:50, 4.7) -- (axis cs:100, 6.2) --
                        (axis cs:200, 9.0) -- (axis cs:300, 9.7) -- (axis cs:400, 9.7) --
                        (axis cs:500, 9.7) --
                        (axis cs:500, 9.1) -- (axis cs:400, 9.1) -- (axis cs:300, 8.9) --
                        (axis cs:200, 8.0) -- (axis cs:100, 5.4) -- (axis cs:50, 4.3) -- cycle;

                        \draw[darknavy, line width=0.5pt, dashed] (axis cs:20,0) -- (axis cs:20,9.5);
                        \node[anchor=south west, font=\sffamily\tiny\bfseries,
                            text=figcontext!60!darknavy] at (axis cs:23, 0.0) {Event\,$\mathrm{do}(C{=}x)$};
                        \draw[figcontext!60!darknavy, line width=0.35pt, dotted, line cap=round]
                        (axis cs:23, 5.85) -- (axis cs:21, 4.2);

                        \addplot[figgreen, line width=1pt] coordinates { (0,4) (500,4) };
                        \addplot[forgisorange, line width=1pt, smooth] coordinates { (0,3.9) (100,4.1) (200,4.1) (300,4.1) (400,4) (500,4) };
                        \addplot[forgisorange, line width=1pt, dashed, smooth] coordinates { (0,3.9) (50,4.5) (100,5.8) (200,8.5) (300,9.3) (400,9.4) (500,9.4) };
                        \filldraw[darknavy] (axis cs:20,3.9) circle (1.5pt);
                    \end{axis}
                \end{scope}

                \node[anchor=south, font=\sffamily\scriptsize\itshape, darknavy, text width=7.8cm, align=center]
                at (0, -\panelh/2+0.45) {``If that event had occurred at $t{=}20$\,ms, what would the trajectory have been?''};
                \node[anchor=south, font=\sffamily\tiny, figsteel, text width=7.8cm, align=center]
                at (0, -\panelh/2+0.12) {Reasoning about alternative histories conditioned on observed facts.};
            \end{scope}

            \begin{scope}[shift={(P4center)}]
                \fill[panelbg, rounded corners=4pt] (-\panelw/2, -\panelh/2) rectangle (\panelw/2, \panelh/2);
                \draw[L4tint!85!darknavy, line width=1.0pt, rounded corners=4pt] (-\panelw/2, -\panelh/2) rectangle (\panelw/2, \panelh/2);

                \filldraw[L4tint] (-\panelw/2+0.34, \panelh/2-0.30) circle (0.20);
                \node[font=\sffamily\bfseries\tiny, white] at (-\panelw/2+0.34, \panelh/2-0.30) {L4};
                \node[anchor=west, font=\sffamily\bfseries\small, L4tint!85!darknavy] at (-\panelw/2+0.62, \panelh/2-0.30) {Decision};
                \node[anchor=east, font=\sffamily\scriptsize, text=darknavy!85] at ( \panelw/2-0.30, \panelh/2-0.30) {$\pi^{*}=\arg\max_{\pi} U(\pi\mid E,S,C,K)$};
                \draw[L4tint!30, line width=0.35pt] (-\panelw/2+0.20, \panelh/2-0.62) -- (\panelw/2-0.20, \panelh/2-0.62);

                \drawminirobot{-3.0}{-0.4}{0}{L4tint}

                \node[draw=L4tint, fill=L4tint!15, rounded corners=1.5pt, line width=0.6pt,
                    font=\sffamily\scriptsize\bfseries, text=L4tint!80!darknavy,
                    minimum width=1.45cm, inner xsep=3pt, inner ysep=1pt,
                    anchor=west, name=stepA] at (-1.85, 0.50) {a) stop};
                \node[draw=L4tint, fill=L4tint!15, rounded corners=1.5pt, line width=0.6pt,
                    font=\sffamily\scriptsize\bfseries, text=L4tint!80!darknavy,
                    minimum width=1.45cm, inner xsep=3pt, inner ysep=1pt,
                    anchor=west, name=stepB] at (-1.85, 0.0) {b) inspect};
                \node[draw=L4tint, fill=L4tint!15, rounded corners=1.5pt, line width=0.6pt,
                    font=\sffamily\scriptsize\bfseries, text=L4tint!80!darknavy,
                    minimum width=1.45cm, inner xsep=3pt, inner ysep=1pt,
                    anchor=west, name=stepC] at (-1.85, -0.50) {c) resume};

                \draw[L4tint!80!darknavy, line width=0.7pt,
                        -{Stealth[length=3.5pt, width=3.5pt]}]
                (stepA.south) -- (stepB.north);
                \draw[L4tint!80!darknavy, line width=0.7pt,
                        -{Stealth[length=3.5pt, width=3.5pt]}]
                (stepB.south) -- (stepC.north);

                \node[draw=darknavy!40, fill=white, rounded corners=2pt, line width=0.5pt,
                    font=\sffamily\scriptsize, text=darknavy, align=left,
                    minimum width=3.8cm, minimum height=2.2cm, inner sep=6pt]
                (plan) at (1.85, -0.10) {%
                    \textbf{\textcolor{L4tint!80!darknavy}{Diagnosis:}} \emph{collision\_foam}\\[0.2em]
                    \textbf{\textcolor{L4tint!80!darknavy}{Plan:}}\\[0.05em]
                    \;a) stop \& verify\\
                    \;b) inspect workspace\\
                    \;c) resume at low speed
                };
                \node[anchor=south, font=\sffamily\scriptsize\itshape, darknavy, text width=7.8cm, align=center]
                at (0, -\panelh/2+0.45) {``Generate actionable recovery procedures given faults, manuals, and sensors.''};
                \node[anchor=south, font=\sffamily\tiny, figsteel, text width=7.8cm, align=center]
                at (0, -\panelh/2+0.12) {Synthesizing actions from observations and the knowledge graph $K$.};
            \end{scope}

        \end{tikzpicture}
    }
    \caption{The four levels of machine understanding defined by FactoryBench.}
    \label{fig:pearl_ladder}
\end{figure}

Figure~\ref{fig:pearl_ladder} instantiates Pearl's three causal levels (Association, Intervention, Counterfactual) on robotic time-series data and adds the Level~4 decision-making layer that FactoryBench introduces on top. Each panel pairs the formal causal primitive with a representative FactoryBench template.

\begin{table}[htbp]
    \centering
    \caption{Number of active question templates per level, by answer format. Counts reflect templates that produce released Q\&A items in the current dataset.}
    \label{tab:template_inventory}
    \small
    \begin{tabular}{@{}lcccccc@{}}
        \toprule
        Level                       & MC single-select & MC multi-select & Ranking & Tensor / Numerical & Free-form & Total       \\
        \midrule
        \textbf{1} (State)          & 1                & 1               & 0       & 2                  & 0         & 4           \\
        \textbf{2} (Intervention)   & 3                & 2               & 2       & 3                  & 0         & 10          \\
        \textbf{3} (Counterfactual) & 0                & 2               & 1       & 2                  & 0         & 5           \\
        \textbf{4} (Decision)       & 0                & 0               & 0       & 0                  & 2         & 2           \\
        \midrule
        \textbf{Total}              & 4                & 5               & 3       & 7                  & 2         & \textbf{21} \\
        \bottomrule
    \end{tabular}
\end{table}

\paragraph{Expert-authored severity ranking.}
Several Level-2 ranking templates (L2.1, L2.9) require ordering anomalies or signal segments by severity. Severity is not directly readable from the labels: it depends on physical impact, controller reaction, recovery cost, and downstream production consequences. The reference ordering used for these templates was authored and cross-checked by a panel of PhD-level robotics experts, who ranked each anomaly category along these axes and reviewed the resulting ranks for cross-anomaly consistency. The fixed expert ranking is shipped with the benchmark and used as the gold permutation by the deterministic positional-match scorer (Appendix~\ref{app:answer-formats}).

\subsection{Full template list}
\label{app:template_list}

Table~\ref{tab:template_list} lists every active template (the templates that emit released Q\&A items in the current dataset). Placeholders in braces (\{task\}, \{anomaly\}, \{phase\}, \{signal\}, \{n\}, \{event\}, \{window\_length\}, \{joint\_signal\}) are filled at generation time from the episode metadata or the relevance specification. The trailing answer-format reminders (\emph{``Answer only with $\ldots$''}) are truncated for readability, matching the convention used in Table~\ref{tab:qa_examples}.

\begin{small}
    \renewcommand{\arraystretch}{1.4}
    \begin{longtable}{@{}>{\centering\arraybackslash}p{0.05\textwidth}>{\raggedright\arraybackslash}p{0.22\textwidth}>{\raggedright\arraybackslash}p{0.66\textwidth}@{}}
        \caption{Full list of FactoryBench question templates by causal level. \emph{Single-choice MCQ} requires one letter; \emph{multi-select MCQ} requires a binary string over a fixed option set; \emph{ranking} requires a permutation over $k$ items; \emph{tensor (scalar)} is a single real number accepted within a tolerance band; \emph{tensor (6-vector)} is a JSON array of six real numbers, one per joint axis; \emph{free-form} responses are evaluated by an LLM-as-judge against a reference answer.}\label{tab:template_list} \\
        \toprule
        \textbf{ID} & \textbf{Answer format} & \textbf{Template (boilerplate truncated)}                                                                                                                                                                                                                                                                                                                                                                                                                                                          \\
        \midrule
        \endfirsthead
        \multicolumn{3}{@{}l}{\footnotesize\emph{Table~\ref{tab:template_list} continued from previous page}}                                                                                                                                                                                                                                                                                                                                                                                                                                     \\
        \toprule
        \textbf{ID} & \textbf{Answer format} & \textbf{Template (boilerplate truncated)}                                                                                                                                                                                                                                                                                                                                                                                                                                                          \\
        \midrule
        \endhead
        \midrule
        \multicolumn{3}{r@{}}{\footnotesize\emph{continued on next page}}                                                                                                                                                                                                                                                                                                                                                                                                                                                                         \\
        \endfoot
        \bottomrule
        \endlastfoot

        \multicolumn{3}{@{}l@{}}{\textbf{Level 1: State}}                                                                                                                                                                                                                                                                                                                                                                                                                                                                                         \\
        \midrule
        L1.1        & Tensor (scalar)        & The robot is performing a \{task\} task. We want to isolate the \{phase\} in the robot's time series. Assuming a fixed window length of \{window\_length\} timesteps, at which timestamp should the window begin?                                                                                                                                                                                                                                                                                  \\
        L1.3        & Multi-select MCQ       & What changed between the two instances of robotic time series data?                                                                                                                                                                                                                                                                                                                                                                                                                                \\
        L1.6        & Single-choice MCQ      & What robot does this sensor data originate from?                                                                                                                                                                                                                                                                                                                                                                                                                                                   \\
        L1.7        & Tensor (scalar)        & Given the sensor stream below, what is the expected value of \{signal\} at T+\{n\}ms?                                                                                                                                                                                                                                                                                                                                                                                                              \\
        \midrule

        \multicolumn{3}{@{}l@{}}{\textbf{Level 2: Intervention}}                                                                                                                                                                                                                                                                                                                                                                                                                                                                                  \\
        \midrule
        L2.1        & Ranking                & The sensor stream below is from a robot exhibiting \{anomaly\}. Rank the signal segments listed in the `options' field in the order you would expect them to appear as the anomaly manifests.                                                                                                                                                                                                                                                                                                      \\
        L2.2        & Multi-select MCQ       & The sensor stream below is from a robot exhibiting \{anomaly\}. What would most likely happen next?                                                                                                                                                                                                                                                                                                                                                                                                \\
        L2.3        & Multi-select MCQ       & The sensor stream below is from a robot exhibiting \{anomaly\}. Select all statements that apply in the options provided.                                                                                                                                                                                                                                                                                                                                                                          \\
        L2.4        & Tensor (scalar)        & The sensor stream below is from a robot exhibiting \{anomaly\}. What is the expected value of \{signal\} at T+\{n\}ms?                                                                                                                                                                                                                                                                                                                                                                             \\
        L2.5        & Tensor (6-vector)      & The sensor stream below is from a robot exhibiting \{anomaly\}. What are the expected values of \{joint\_signal\} at T+\{n\}ms?                                                                                                                                                                                                                                                                                                                                                                    \\
        L2.6        & Tensor (scalar)        & Knowing that the robot suffers from \{anomaly\} in the given context time series, we want to isolate the \{phase\} phase. Assuming a fixed window length of \{window\_length\} timesteps, at which timestamp should the window begin?                                                                                                                                                                                                                                                              \\
        L2.7        & Single-choice MCQ      & Given the sensor data from a robot performing the task, determine what anomaly is present?                                                                                                                                                                                                                                                                                                                                                                                                         \\
        L2.8        & Single-choice MCQ      & You are provided with two sensor streams originating from robots accomplishing tasks. What differences between the two given instances of robotic time series data (if any) do you notice?                                                                                                                                                                                                                                                                                                         \\
        L2.9        & Ranking                & Rank the following robot time series segments from most to least severe anomaly.                                                                                                                                                                                                                                                                                                                                                                                                                   \\
        L2.10       & Single-choice MCQ      & Knowing that the robot suffers from \{anomaly\} in the given context time series, what robot does this sensor data originate from?                                                                                                                                                                                                                                                                                                                                                                 \\
        \midrule

        \multicolumn{3}{@{}l@{}}{\textbf{Level 3: Counterfactual}}                                                                                                                                                                                                                                                                                                                                                                                                                                                                                \\
        \midrule
        L3.1        & Ranking                & Given the baseline sensor stream below and the counterfactual scenario where \{event\}, rank the signal segments listed in the `options' field in the order you would expect them to appear as the event manifests.                                                                                                                                                                                                                                                                                \\
        L3.2        & Multi-select MCQ       & Given the counterfactual scenario where \{event\}, what would most likely happen next?                                                                                                                                                                                                                                                                                                                                                                                                             \\
        L3.3        & Multi-select MCQ       & Given the counterfactual scenario where \{event\}, select all statements that would apply.                                                                                                                                                                                                                                                                                                                                                                                                         \\
        L3.4        & Tensor (scalar)        & In the counterfactual scenario where \{event\}, what would be the expected value of \{signal\} at T+\{n\}ms?                                                                                                                                                                                                                                                                                                                                                                                       \\
        L3.5        & Tensor (6-vector)      & In the counterfactual scenario where \{event\}, what would be the expected values of \{joint\_signal\} at T+\{n\}ms?                                                                                                                                                                                                                                                                                                                                                                               \\
        \midrule

        \multicolumn{3}{@{}l@{}}{\textbf{Level 4: Decision}}                                                                                                                                                                                                                                                                                                                                                                                                                                                                                      \\
        \midrule
        L4.1        & Free-form              & Given the sensor stream below, does the machine show signs of anomalous behavior? If yes, identify the most likely root cause and describe the steps you would take to fix it.                                                                                                                                                                                                                                                                                                                     \\
        L4.2        & Free-form              & An engineer wants to increase the effectiveness and accuracy of this machine. Based on the sensor stream below, what operational changes or parameter adjustments could help achieve this? Do not justify your answer, simply indicate steps to take (or specify if nothing can be done).                                                                                                                                                                                                          \\
    \end{longtable}
\end{small}

\paragraph{Prompt structure and knowledge-graph augmentation.}
Models under evaluation receive a single user message; no system role is set on the evaluated model. The message is assembled at prompt-build time from five ordered blocks:

\begin{verbatim}
<machine_sentence>
<task_phase_reference>
<acronym_mapping>
<time_series>
Question: <question_text>
Here are the options:
<options>
\end{verbatim}

\noindent The leading \texttt{<machine\_sentence>} is filled at prompt-build time from the FactoryBench knowledge graph, specifically the machine and gripper tables, keyed on the dataset that the question is grounded in. This guarantees that identical question templates are grounded in the correct hardware description regardless of which underlying dataset the episode comes from. For a UR3e-grounded question on FactoryWave, the resulting line expands to:

\begin{quote}\small
    The following sensor data comes from Universal Robots UR3e, a collaborative robot (cobot) from the E-series with 3 kg payload, 6 degrees of freedom, controlled via UR PolyScope (teach pendant), UrScript, typically used for light assembly tasks, automated workbench scenarios. It is equipped with a 2FG14 Finger Gripper OnRobot.
\end{quote}

\noindent The augmentation is conditional: identification templates such as L1.6 (``What robot does this sensor data originate from?'') and L2.10 suppress the machine sentence so the model cannot read the answer off the prompt header, and templates that probe gripping behavior without revealing tooling can selectively suppress only the gripper clause. The acronym-mapping block decodes the time-series header into long-form channel names and is included whenever the question carries one.

\noindent The \texttt{<task\_phase\_reference>} block defines the task's phases, and is attached to every prompt whose question or options refer to a phase, or whose context exposes the \texttt{task\_phase} channel. Without it the phase vocabulary is ambiguous in two distinct ways: the \texttt{task\_phase} channel appears in the series as a bare integer with no legend, and the question templates refer to phases in prose (``the grasp of the object phase'') that does not match the canonical phase names in the knowledge graph. The block resolves both by listing the phases of the episode's task in their fixed execution order, giving for each the canonical name, the wording the templates use, and a one-line description, with the integer label included only when the context actually exposes \texttt{task\_phase}. For a pick-and-place episode it expands to:

\begin{quote}\small
    Task: Pick and Place. Robot picks an object from a source location and places it at a target location (bin). The task runs through the following phases in this fixed order, labelled by the \texttt{task\_phase} signal: \texttt{0 = Above Pick} (referred to in questions as ``approach to the object''): EEF moves to a waypoint directly above the object. $\ldots$ \texttt{3 = Close} (referred to in questions as ``grasp of the object''): Gripper closes around the object. $\ldots$
\end{quote}

\noindent The block is deliberately timing-free: it states what a phase \emph{is} and where it sits in the task's fixed order, never which timesteps of the episode at hand belong to it. That distinction matters because several templates (L2.6 in particular) ask the model to locate a phase boundary inside the displayed window, and naming the timesteps would hand over the answer. Phase definitions are shared context; phase segmentation is the thing under test.

Level-4 free-form responses are scored by an LLM-as-judge under a strict three-point rubric ($\{0.0,\,0.5,\,1.0\}$) that is template-specific: troubleshooting items (template L4.1) and optimization items (template L4.2) use different prompts, since the relevant axes of correctness differ (root cause + remediation steps vs.\ problematic parameter + adjustment direction).

\paragraph{Troubleshooting prompt (L4.1).}
\begin{verbatim}
You are scoring a model's troubleshooting answer for a robotics
sensor-data benchmark.

Rubric (apply strictly, no other values allowed):
  1.0  - the answer correctly identifies the underlying root cause AND
         the proposed remediation steps are mostly right. The steps do
         NOT need to match the reference exactly: minor wording
         differences, additional sensible checks, or omitted minor
         steps are all fine, as long as the overall remediation is on
         the right track and would plausibly resolve the issue.
  0.5  - the answer correctly identifies the underlying root cause but
         the remediation is wrong, missing, or off-topic (e.g. proposes
         an unrelated fix, contradicts the reference, or refuses to
         give steps).
  0.0  - the root cause is wrong or absent (model refused, identified
         the wrong fault, or stated "no anomaly" when one was present).

The known root cause is provided to you separately. An answer counts
as identifying the root cause when it names the same physical
phenomenon - equivalent paraphrases are fine, but a different fault
category (e.g. saying "payload too heavy" when the root cause is
"TCP misconfiguration") is wrong.

Respond with ONLY a JSON object on a single line, no markdown:
{"score": <0.0|0.5|1.0>, "reason": "<one short sentence>"}

Question:        {question}
Reference:       {reference}
Known root cause:{root_cause}
Model answer:    {prediction}
\end{verbatim}

\paragraph{Optimization prompt (L4.2).}
\begin{verbatim}
You are scoring a model's optimization answer for a robotics sensor-
data benchmark.

Rubric (apply strictly, no other values allowed):
  1.0  - the answer identifies the problematic parameter AND proposes
         adjusting it in the correct direction (increase vs. decrease,
         matching the reference). Exact magnitudes do not matter; only
         direction matters.
  0.5  - the answer identifies the problematic parameter but proposes
         the wrong direction, no direction, or contradictory directions.
  0.0  - the problematic parameter is not identified, or the model
         recommends adjusting an unrelated parameter, or refuses to
         answer.

The reference answer states the correct parameter and its target
value, from which the correct adjustment direction can be inferred
relative to the configured value.

Respond with ONLY a JSON object on a single line, no markdown:
{"score": <0.0|0.5|1.0>, "reason": "<one short sentence>"}

Question:    {question}
Reference:   {reference}
Model answer:{prediction}
\end{verbatim}

\noindent The judge ensemble comprises three frontier LLMs from different vendors: \textbf{GPT-5.1}, \textbf{Claude Sonnet 4.6}, and \textbf{DeepSeek V3.2}. Each judge independently scores every Level-4 free-form response with the appropriate template-specific prompt above, and the per-item score reported in our results is the median of the three valid votes (snapped to the $\{0,\,0.5,\,1\}$ grid).

\section{Episode eligibility filtering}
\label{app:episode_filtering}

Before any sub-window is drawn (Appendix~\ref{app:relevance}), each template restricts the pool of source episodes to those that can support a well-defined ground-truth answer. The filter is template-aware and combines a level-specific condition predicate with a length/coverage check.

\paragraph{Level-specific condition predicate.} The level a template lives at fixes which episode condition the template can pull from: Level~1 templates probe nominal state and therefore consume only \emph{baseline} (normal) episodes, Level~2 templates probe the immediate effect of an intervention and therefore consume only \emph{anomalous} (fault) episodes, and Level~3 templates probe counterfactual reasoning and therefore consume only \emph{counterfactual} pairs (one baseline plus one fault run, paired by the protocol described in Appendix~\ref{app:data_recording}). Level~4 templates draw from baseline and fault episodes depending on the troubleshooting/optimization variant. Episodes whose condition does not match the template are removed from the pool before sampling.

\paragraph{Ground-truth feasibility.} Eligible episodes must additionally carry the labels and trajectory length the template needs to derive its answer. Concretely, we require that (i) the episode contains every metadata field the template instantiates (root-cause label, anomaly type, fault-injection timestep, target phase, etc.), (ii) the episode is long enough that a relevance-compliant sub-window of the requested length (Appendix~\ref{app:context_construction}) fits, and (iii) for templates whose answer is computed from a future segment of the trajectory (numerical and tensor forecasts at L1.7, L2.4, L2.5), the trajectory extends at least the requested forecast horizon beyond the displayed window. Episodes that fail any of these checks are dropped at filter time rather than at sampling time, so each template's sample distribution is over its true eligibility pool.

\paragraph{Effect on the released set.} The filter is applied per (template, dataset, condition) cell before release, so the public 70k Q\&A items released on Hugging Face are exactly the items for which a verifiable ground-truth answer exists.

\section{Fault-relevance window sampling}
\label{app:relevance}

For Q\&A pairs grounded in a fault episode, the time-series context shown to the model is a sub-window of the full episode rather than the entire trajectory. A naive uniformly sampled sub-window can miss the segment where the fault is actually visible: for a transient collision lasting tens of milliseconds, a uniformly sampled one-second window has a non-negligible probability of falling entirely outside the disturbed segment, producing an item whose ground-truth label is correct but whose evidence is invisible to any model. To prevent this, each fault type is paired with a \emph{relevance specification} that constrains how the sub-window is drawn so that the fault signature is, by construction, present in the displayed context.

\paragraph{Four temporal footprints.} Each fault is assigned, by domain annotation, to one of four classes that describe how its signature is distributed across the episode:

\begin{itemize}
    \item \emph{Global}: the signature is present throughout the entire episode, for instance a payload misconfiguration that biases every torque reading. Sub-windows are sampled uniformly under the requested length bounds.
    \item \emph{Event-anchored}: the signature is a transient localized at a specific timestep recorded in the episode (collisions, gripper misactivation). The sub-window is required to contain that timestep; its length and start are otherwise random.
    \item \emph{Phase-gated}: the signature is only diagnostic during specific task phases, for instance a peg-misalignment fault that is only visible during insertion. The sub-window is required to overlap one of the target phases by at least a minimum number of rows, where the relevant phases depend on the task.
    \item \emph{Cumulative}: the signature has a monotonically growing footprint that becomes visible only after a certain phase begins, for instance progressive torque drift. The sub-window is required to be at least a minimum length and, where applicable, to start no earlier than the onset phase.
\end{itemize}

\paragraph{Sampling and validation.} Given a fault episode and the requested length bounds, the sampler draws a window that satisfies the corresponding constraint and records the constraint that was applied alongside the question's provenance, so downstream analyses can condition on it. A post-hoc check verifies that the returned window meets the locality's requirement; if it cannot be met (e.g., a phase-gated fault whose target phase is missing from the recording), the item is either discarded or, for templates where the anomaly signature is helpful but not required for correctness, the sub-window falls back to uniform sampling and the item is flagged accordingly.

\paragraph{Effect on the dataset.} Relevance-aware sampling is applied to all fault-grounded items in Levels~1, 2, and~4. Level~3 counterfactual pairs use a separate fixed-injection-timestep protocol described in Section~\ref{sec:density} and do not rely on this mechanism. In pilot generation runs without relevance constraints, a non-trivial fraction of fault items had sub-windows that did not contain the fault signature, inflating the apparent difficulty of Levels~1 and~2 in a way that could not be distinguished from genuine model failure. The mechanism can be disabled to recover uniform sampling for ablation studies that require a no-relevance baseline.

\section{Time-series context: features and window length}
\label{app:context_construction}

Two design choices govern what each Q\&A item actually shows the model: \emph{which} channels of the multivariate trajectory are included, and \emph{how long} the displayed window is. Both choices are deliberately constrained.

\paragraph{Resampling to a uniform 10\,Hz.} The source datasets record at heterogeneous frequencies (83--125\,Hz for FactoryWave, 100\,Hz for AURSAD~\cite{leporowski2022aursad}, 100 or 500\,Hz for voraus-AD~\cite{brockmann2023vorausad}). Before any Q\&A item is built, every episode is anti-alias filtered and its \emph{context window} is downsampled to a common nominal 10\,Hz timeline. This applies to the rows the model sees in the prompt; the \texttt{timestamp\_ms} column is preserved from the nearest source-rate sample and therefore is not exactly on a 100\,ms grid (consecutive row offsets typically fall in $[95, 105]$\,ms depending on the native rate). Post-event windows used to compute forecast ground truth (Levels~2 and~3, templates 4 and 5) additionally retain the native sampling rate to preserve fast transients around fault onset; the answer horizon can therefore be sub-100\,ms even though the visible context rows are $\sim$100\,ms apart. This context-downsampling serves three purposes: it normalizes the time axis across robots so that templates can specify window lengths in absolute seconds without conditioning on the source platform, it keeps the displayed context within the prompt budget of every evaluated model, and it removes high-frequency content that would otherwise dominate the token sequence with information not needed for the reasoning levels we evaluate. Ten Hertz is admittedly low by industrial-sensor standards: at higher sampling rates one can resolve fast transients (millisecond-scale collisions, valve switches) that are smoothed away here. We accept this trade-off because the fault catalogue evaluated in this paper is deliberately simplified to atomic, well-isolated mechanisms, and we observe empirically that 10\,Hz is sufficient to expose every fault signature in our catalogue at a level that PhD-level annotators can verify on the displayed context. Future versions of the benchmark targeting compound or sub-second faults would need to revisit this choice.

\paragraph{Per-template feature pre-selection.} The raw signal schema is large: FactoryWave exposes more than 120 channels per timestep (joint positions, velocities, accelerations, currents, target torques, contact forces, TCP pose, gripper state, and so on). Including every channel verbatim would produce contexts that are both prohibitively long and dominated by signals irrelevant to the question being asked. We therefore equip every question template with a hand-curated subset of \emph{important features}: 30~channels per template for Levels~1--3, and roughly 20~for Level~4 (whose templates focus on the diagnostic and remediation signals that matter for troubleshooting). The selections are authored by PhD-level robotics experts on the team and target the channels most informative for the specific reasoning skill the template probes; for instance, an anomaly-detection template emphasizes torques and contact forces, while a phase-isolation template foregrounds joint positions and velocities. The full per-template feature lists are released alongside the benchmark so that every Q\&A item can be reproduced bit-for-bit from the underlying episode.

\paragraph{Window length.} Across all four levels, each Q\&A item shows a sub-window of between 32 and 64 timesteps drawn from the 10\,Hz-resampled trajectory, corresponding to roughly $3.2$--$6.4$\,s of robot motion. The lower bound is set so that even the shortest displayed window contains enough samples to expose the dynamics the templates probe (transient onsets, phase boundaries, force spikes); the upper bound keeps the prompt within the context limits of all evaluated models, including the smaller open-weight ones. Within these bounds, the exact length and start position are randomized at generation time, with the start position constrained by the relevance specification of Appendix~\ref{app:relevance} when the item is fault-grounded. These bounds apply per displayed window rather than per item, so templates that show more than one window carry a proportionally longer total context: comparison templates present two sub-windows side by side, and ranking templates add four candidate segments on top of the context window.

\paragraph{Why these choices matter for interpretability.} Fixing the resampling rate, the feature subset, and the window-length range across templates means that performance differences between models on a given template cannot be attributed to one model receiving a longer, faster, or richer context than another, only to its ability to interpret the same content. The randomization within bounds prevents models from exploiting positional or length cues to recognize templates by their surface form: two items instantiated from the same template are typically presented with different lengths, different start positions, and -- once relevance constraints and per-template feature subsets are honoured -- with the fault evidence falling at different relative positions in the displayed window.

\section{Validation of solvability}
\label{app:solvability}

This appendix collects the diagnostic checks we ran to convince ourselves that FactoryBench items are answerable from the released time-series context, rather than from question wording alone, and that the templates flagged as ``too easy'' by the noise-substitution check are nevertheless grounded in genuine signal structure.

\subsection{Noise-substitution check}
\label{app:noise_substitution}

To probe whether the model scores in Figure~\ref{fig:results} reflect actual reading of the time-series context or merely priors over question templates and option positions, we constructed a noise-substituted counterpart of the dataset and re-ran every model against it. This check was an early diagnostic study run on a previous, smaller model panel (Claude Haiku 4.5, DeepSeek V3.1, gpt-5.1, Mistral Large 3) and a 400-item Q\&A subset (100 items per level); it predates the main evaluation reported in the body of the paper, so the absolute scores below are not directly comparable to those in Figure~\ref{fig:results}. In particular, this early subset was generated with substantially longer time-series contexts than the released benchmark, which depresses scores across the board for every model in the panel; the relative differences between Orig and Sub remain interpretable, but absolute levels sit well below the comparable cells in the main results.

\paragraph{Construction.} For each of the 400 generated Q\&A pairs (100 per level), we duplicate the item and rewrite only the numerical time-series content. For each numeric \emph{float} feature we preserve the original first value and substitute every subsequent value with the cumulative sum of independent Gaussian increments:
\begin{equation*}
    v'_0 = v_0, \qquad v'_{t+1} = v'_t + \mathcal{N}(0, \sigma^2), \qquad \sigma = 0.5,
\end{equation*}
where $\sigma$ is shared across all features. The same substitution is applied to time-series chunks embedded in option strings (Level~1 severity-ranking, Level~2/3 chunk-ranking).

\paragraph{Expected behavior.} A model that scores by genuinely reading the time series should drop sharply on the noise-substituted variant, because the substitution destroys the signal-trajectory information that the gold answer hinges on. A model that scores by template-level priors (e.g.\ guessing the most plausible MC option given the question wording) or by option-position bias should be roughly invariant, since the question text and options are unchanged.

\paragraph{Result.} Table~\ref{tab:noise_substitution} reports the side-by-side mean signed chance-corrected scores. Three patterns emerge.

\begin{enumerate}
    \item On Level~1 every model is essentially flat or improves slightly under substitution ($\Delta \in [-0.8, +4.0]$). This is consistent with L1 templates being substantially answerable from the question wording, the option set, and the still-present discrete annotations of task phase and fault label. It also flags L1 as the level where overlap between LLM scores and a random/regression baseline is most expected.
    \item On Level~3 every model drops sharply, by 8 to 18 signed CC points; on Level~2 the two strongest signal-readers of the panel (gpt-5.1 and DeepSeek V3.1) also drop substantially ($-12.0$ and $-3.4$ respectively) while the weaker readers stay flat. The L3 drops are the most informative: L3 templates ask counterfactual ``what would have happened'' questions where the answer hinges on the actual signal trajectory rather than on a discrete annotation, and every panel member's L3 score collapses from clearly-above-chance to at-or-below-chance under substitution.
    \item On Level~4, gpt-5.1's $7.6\%$ original score collapses to $0.0\%$ under noise substitution. Inspection of the per-item rubric notes shows that gpt-5.1's L4 mass on the original data came almost entirely from correctly emitting ``no anomaly is present'' on truly nominal episodes; once the time series is replaced by Gaussian noise, the same nominal episodes look anomalous to the model and it now hallucinates faults on them, losing its only mode of scoring. The other three models score $\leq 0.005$ in either condition and have no comparable mode to lose.
\end{enumerate}

\begin{table}[htbp]
    \centering
    \caption{Noise-substitution check: signed chance-corrected mean scores (\%) on the original FactoryBench Q\&A pairs (Orig) vs.\ the noise-substituted variant (Sub) constructed by replacing every numerical time-series value with cumulative Gaussian increments seeded at the original first value ($\sigma=0.5$). L4 uses the uncorrected free-form rubric. Questions, options, and gold answers are identical between the two runs. $\Delta = \text{Sub} - \text{Orig}$; large negative values are evidence that the model was using the data from the time series rather than the question alone.}
    \label{tab:noise_substitution}
    \scriptsize
    \begin{tabular}{@{}l ccc ccc ccc ccc@{}}
        \toprule
                         & \multicolumn{3}{c}{L1} & \multicolumn{3}{c}{L2} & \multicolumn{3}{c}{L3} & \multicolumn{3}{c}{L4}                                                                    \\
        \cmidrule(lr){2-4}\cmidrule(lr){5-7}\cmidrule(lr){8-10}\cmidrule(lr){11-13}
        Method           & Orig                   & Sub                    & $\Delta$               & Orig                   & Sub  & $\Delta$ & Orig & Sub  & $\Delta$ & Orig & Sub & $\Delta$ \\
        \midrule
        Claude Haiku 4.5 & -12.7                  & -8.7                   & +4.0                   & +2.1                   & +4.5 & +2.4     & +8.6 & -5.6 & -14.2    & 0.5  & 0.0 & -0.5     \\
        DeepSeek V3.1    & -15.7                  & -16.5                  & -0.8                   & -1.7                   & -5.1 & -3.4     & +5.3 & -2.8 & -8.1     & 3.5  & 0.0 & -3.5     \\
        gpt-5.1          & +2.8                   & +3.5                   & +0.7                   & +7.2                   & -4.8 & -12.0    &+12.6 & -2.6 & -15.2    & 7.6  & 0.0 & -7.6     \\
        Mistral-Large-3  & -1.2                   & +2.5                   & +3.7                   & -0.7                   & -1.6 & -0.9     &+17.9 & -0.2 & -18.1    & 0.0  & 0.0 &  0.0     \\
        \bottomrule
    \end{tabular}
\end{table}

\subsection{Human expert baseline}
\label{app:human_baseline}

The strongest evidence that an item is solvable from the released telemetry is a human solving it. We therefore had a robotics expert answer a $60$-item sample spanning Levels~1--3 under open-tool, no-LLM conditions, and scored the answers with the benchmark's own grader.

\paragraph{Protocol.}
\emph{Open tools.} The annotator was free to use any analysis instrumentation, and built a small Python toolkit for the task: series parsing, windowed condition search and threshold crossings, linear extrapolation, per-channel signal statistics, forward-kinematics consistency checks, and nominal-versus-actual absolute-difference comparison. The reference we want is what a competent engineer with normal instrumentation can extract from the telemetry, not what a person can do reading serialized numbers unaided.
\emph{No model in the answer path.} The annotator did not query a language model to produce, check, or revise any answer. Ordinary coding assistance was used while building the toolkit; this does not couple the baseline to the systems under evaluation, because the toolkit is deterministic code whose outputs are reproducible from the released telemetry, and every released item carries the runnable derivation that produces its answer together with that code's captured output. The comparison in Table~\ref{tab:human_baseline} is therefore independent of the panel in the only sense that matters: no panel model, and no language model, contributed to any expert answer.
\emph{Identical items, identical scoring.} The annotator worked from the same items and prompts the model panel saw, and the answers were scored by the benchmark's own grader: the same acceptance bounds for scalar and vector answers, the same exact-match rule for single-select and ranking, and the same per-position partial credit on multi-select. No separate human rubric was introduced, and the human receives no credit the panel would not receive.
\emph{Duration.} The entire process took a total of two work days of question answering and tool design by the expert.

\paragraph{Composition.}
$13$ items at Level~1, $37$ at Level~2, and $10$ at Level~3, drawn across $14$ distinct templates and covering all five answer formats. We plan on extending the baseline to address this current imbalance.

\paragraph{Results.} Table~\ref{tab:human_baseline} reports the expert against the model panel on the identical $60$-item sample.

\begin{table}[htbp]
    \centering
    \caption{Human expert baseline against the model panel, raw scores on the identical $60$-item sample. The expert reaches $0.917$ overall ($49/60$ items exactly correct).}
    \label{tab:human_baseline}
    \begin{tabular}{lcccccc}
        \toprule
         & Expert & Claude & Mistral & GPT-5.1 & Qwen3 & DeepSeek \\
         &        & Sonnet 4.6 & Large 3 &     & 235B  & V3.2 \\
        \midrule
        Level 1 & \textbf{0.962} & 0.712 & 0.500 & 0.635 & 0.519 & 0.481 \\
        Level 2 & \textbf{0.980} & 0.662 & 0.444 & 0.392 & 0.385 & 0.351 \\
        Level 3 & 0.625 & 0.325 & 0.550 & 0.500 & 0.575 & \textbf{0.675} \\
        \bottomrule
    \end{tabular}
\end{table}

The expert clears $0.96$ at both Level~1 and Level~2 while the best model sits at $0.68$ and the rest between $0.39$ and $0.46$. This settles the solvability question directly: the items are answerable from the released signals, so the low model scores reflect model limitations rather than broken or ambiguous items. It also shows that human experts are capable of a much deeper comprehension of machine state, whether healthy or anomalous, even though that comprehension is arrived at through a time-consuming process of measurement and computation (usually causing costly downtime) rather than intuitive signal comprehension.

The margin is not uniform across templates. It is narrowest where the answer can be read off a single channel with a short calculation, and widest on the templates that require committing to a specific root cause or a specific machine from the signal.

\paragraph{Level 3 is materially harder for the expert too.}
At Level~3 the expert reaches $0.625$ with only $3$ of $10$ items exactly correct, and is not the top scorer on that level. This is generally explained by how undercovered and difficult the problem of counterfactual prediction is in time series analysis, leading to a lack of specific tools: the toolkit that carries the expert to near-ceiling on Levels~1--2 has no counterpart for reasoning about an intervention that was never executed. We read this as a property of the level rather than of the items, and it tempers the interpretation of the Level~3 column in Figure~\ref{fig:results}: the human ceiling there is much closer to the panel than at Levels~1--2.

\paragraph{Extension: a verified $102$-item baseline released per item.}
The $60$-item run above has since been superseded by a substantially larger and
more carefully controlled effort, released in full with this paper. An expert,
working from the raw telemetry with a purpose-built analysis toolkit, produced
answers for \textbf{$102$ items}: $25$ at Level~1, $42$ at Level~2, $23$ at
Level~3, and $12$ at Level~4. For \emph{every} item we release the question, its
option set and context, the expert's answer, a prose derivation, the runnable
code that derives the answer from the telemetry, and that code's captured
output. This discharges in full the commitment to release, per item, the
expert's answer and the method used to reach it, and it puts a human reference
point on all four levels, including the decision-making level that carries this
paper's central claim.

\paragraph{Admission criteria.}
Items were admitted only after clearing two independent bars, and the second is
the one that matters for a benchmark. \emph{Real provenance:} the item's
question text, digit-normalised, had to match an entry in the official
generator's output byte-for-byte rather than merely resemble one, and items that
had been hand-authored at any point were excluded however well-formed they
looked. \emph{No known unfixed shortcut:} an item either never had a
construct-validity defect, or the defect was fixed and then re-verified
\emph{blind} by an independent solver working from the corrected content with no
sight of the stored answer. Per-item provenance is retained (dataset, episode
identifier, subseries offsets and phase), so any item can be re-derived from a
fresh pull of the release. The defects found and fixed along the way are filed as
per-level template reviews, which are released alongside the baseline; several
document shortcuts a model could otherwise have exploited, and are as much a
part of the benchmark's construct validity as the answers are.

\paragraph{What this baseline does and does not establish.}
It establishes that the items are solvable from the released signals, and it
makes each judgement auditable: a reader who doubts an answer can read the
derivation and run the code. It is \emph{not} a measurement of human accuracy
under the benchmark's grader. The expert worked with instrumentation and without
a time limit, and the released answers are the verified ground truth for those
items rather than an attempt scored against a separate key, so no human score is
reported here and none should be inferred by comparing these items to the panel.
Blind re-verification, which is the closest thing to an independent human
attempt, was run on a subset rather than the whole set: of the $67$ Level-1 and
Level-2 items carrying a grading record, $20$ have a blind verdict, of which $19$
reproduced the stored answer and one did not and is annotated. Extending blind
verification to the remaining items, and scoring a genuine human attempt with the
benchmark's own grader, are the two things that would turn this into a human
accuracy number.

\paragraph{Relation to the two earlier passes.}
The counts of the three efforts are not pooled, and only this one should be cited
as the released baseline. The $60$-item run of
Table~\ref{tab:human_baseline} remains the source of the scored human-versus-panel
comparison reported above, because it alone was put through the benchmark's
grader; its per-item artifact was lost before release and cannot be recovered. An
intermediate $21$-item pass, one item per active template, is superseded by the
$102$-item release in both coverage and rigour. The per-level target of $25$ items
is met at Levels~1 and~2 and not yet at Levels~3 ($23$) and~4 ($12$).

\subsection{Distribution-shift analysis}
\label{app:arm_disturbance}

\begin{figure}[htbp]
    \centering
    \includegraphics[width=0.75\textwidth]{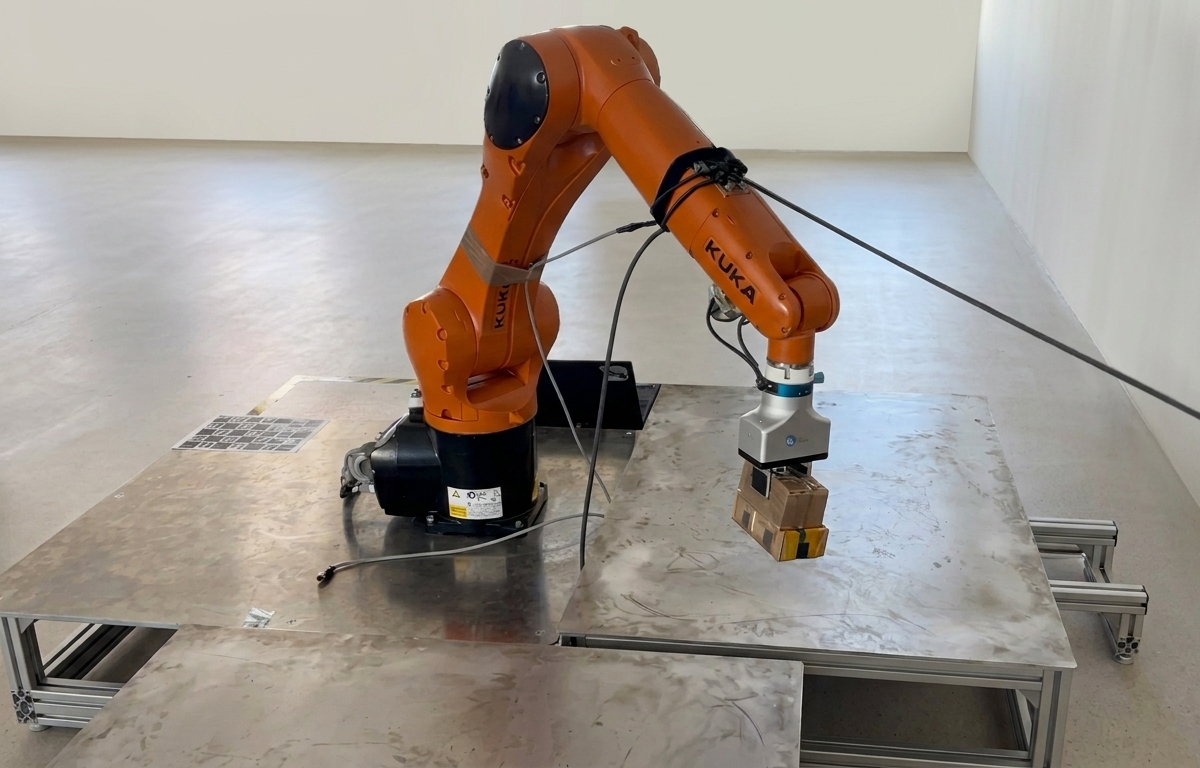}
    \caption{The arm-disturbance fault as physically applied to the KUKA in FactoryWave.}
    \label{fig:arm_disturbance_kuka}
\end{figure}

\begin{figure}[htbp]
    \centering
    \includegraphics[width=\textwidth]{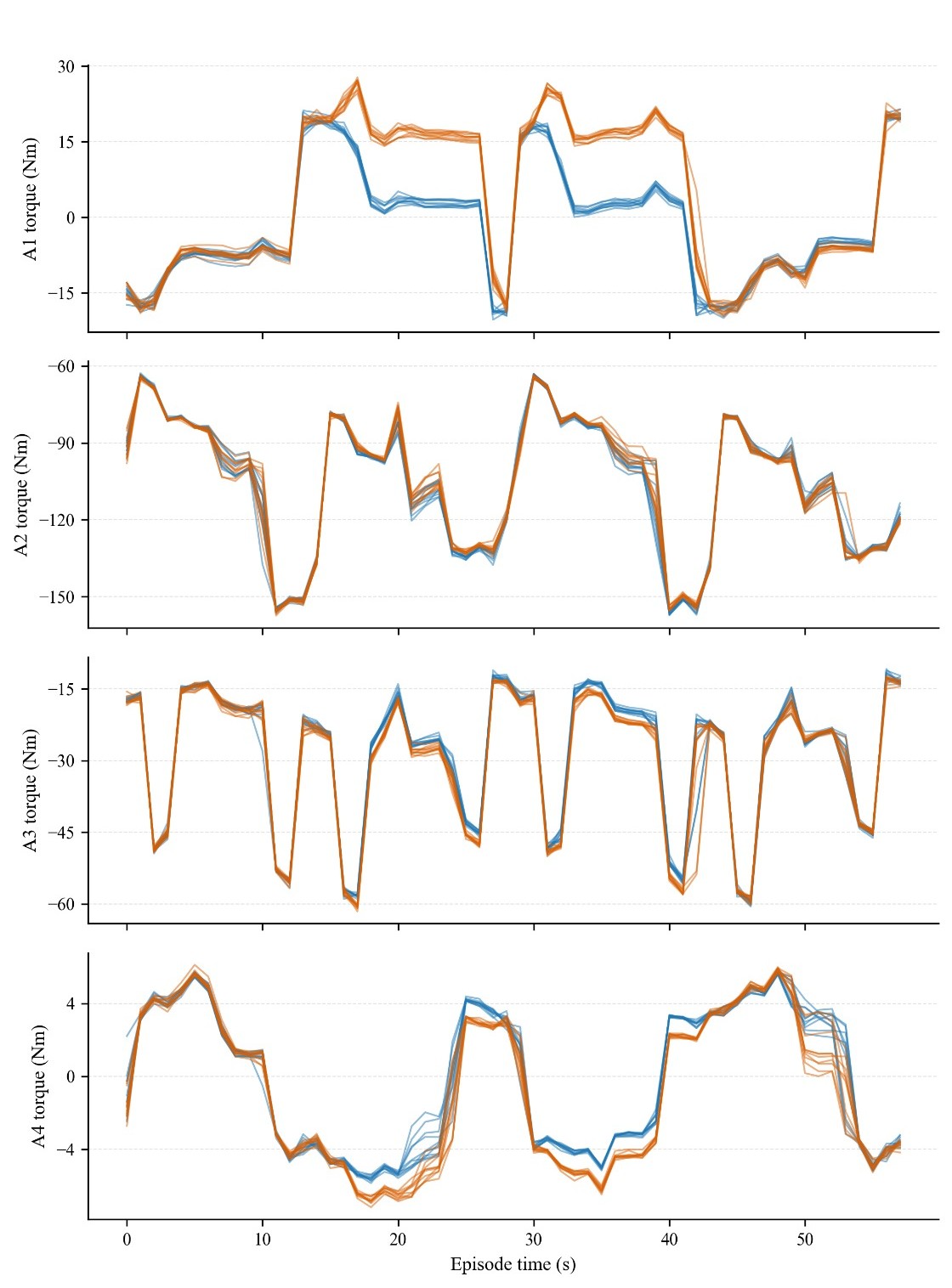}
    \caption{Sensor data distribution across multiple healthy KUKA episodes (blue) versus multiple faulty episodes with an arm-disturbance fault (orange) in FactoryWave. The visible shift between the two groups confirms that templates with small noise-substitution $\Delta$ are still solvable from the released signal; the magnitude of the shift varies across joints because the disturbance angle loads them unequally.}
    \label{fig:arm_disturbance}
\end{figure}

For the templates that produced a small noise-substitution $\Delta$ in the previous subsection, we confirm that they are nevertheless solvable by further empirical analysis, for example by inspecting how the sensor data distribution shifts when a fault is present. Figure~\ref{fig:arm_disturbance} illustrates this by comparing multiple healthy KUKA episodes (blue) against multiple faulty episodes with an arm-disturbance fault (orange) drawn from FactoryWave (Figure~\ref{fig:arm_disturbance_kuka} shows the physical setup of the fault for visual intuition), and reveals a clear distributional shift between the two groups on the channels the affected templates read from. The shift is more pronounced on some joints than on others, as the angle at which the disturbance is applied loads them unequally.

\section{Details of the FactoryWave dataset}
\label{app:data_recording}

FactoryWave was collected on two one-armed platforms: the \textbf{Universal Robots UR3} (cobot; OnRobot Screwdriver and two-finger gripper variants) and the \textbf{KUKA KR10} (industrial arm). Each episode corresponds to one complete task cycle; episodes are recorded under three conditions (normal, fault, and counterfactual), described below. Because state interpretation rather than policy quality is the object of study, motion is generated by a deterministic scripted controller. Every sample carries extensive labeling metadata, and the full native-rate sensor stream is logged without online feature selection. Table~\ref{tab:factorywave_distribution} gives the breakdown of the 8{,}983 episodes across robots, tasks, and conditions.

\begin{table}[htbp]
    \centering
    \caption{Distribution of FactoryWave episodes by robot, task, and condition. Counterfactual entries count CF groups (one baseline plus one retained fault run per group), as each one represents data of one studied scenario.}
    \label{tab:factorywave_distribution}
    \small
    \begin{tabular}{@{}llrrrr@{}}
        \toprule
        Robot          & Task           & Normal        & Fault         & CF groups    & Total         \\
        \midrule
        UR3            & Pick-and-Place & 1053          & 3238          & 109          & 4400          \\
        UR3            & Screwing       & 230           & 850           & 40           & 1120          \\
        UR3            & Peg-in-Hole    & 322           & 1200          & 100          & 1622          \\
        KUKA KR10      & Pick-and-Place & 578           & 1221          & 42           & 1841          \\
        \midrule
        \textbf{Total} &                & \textbf{2183} & \textbf{6509} & \textbf{291} & \textbf{8983} \\
        \bottomrule
    \end{tabular}
\end{table}

\subsection{Tasks and phase structure}

Three industrial tasks are represented in the dataset: Pick-and-Place, Screwing, and Peg-in-Hole. All three are executed on both robots, using matched scripted controllers and task layouts so that cross-embodiment comparisons hold the task fixed. We structure each task as a deterministic sequence of phases (Tables~\ref{tab:phases_pp}, \ref{tab:phases_screw}, \ref{tab:phases_peg}); the phase index is written as a discrete label at every timestep, providing an interpretable segmentation that downstream models can condition on or that can be used to localize questions to a specific stage of the cycle. Within this fixed phase skeleton, we randomize as much of the episode as the task permits so that a question template that targets a given phase does not accidentally learn a fixed pose, timing, or trajectory. Pick and place locations are sampled uniformly from predefined workspace regions; phase-specific joint velocities, accelerations, and end-effector (EEF) rotations are drawn from bounded per-phase ranges; and approach angles to contact events are perturbed within the tolerances of the controller. This forces the Q\&A pairs generated from FactoryWave to reflect phase-level reasoning rather than memorisation of a single scripted trajectory, which in turn makes them generalizable to a wide variety of trajectories and variation found in real recordings.

\begin{table}[htbp]
    \centering
    \caption{Task phases for Pick-and-Place. EEF refers to End-Effector}
    \label{tab:phases_pp}
    \small
    \begin{tabular}{clp{8cm}}
        \toprule
        Phase index & Phase name           & Description                                       \\
        \midrule
        0           & \texttt{above\_pick} & EEF moves to a waypoint directly above the object \\
        1           & \texttt{descend}     & EEF lowers toward the object                      \\
        2           & \texttt{settle}      & Brief hold to let physics settle before grasping  \\
        3           & \texttt{close}       & Gripper closes around the object                  \\
        4           & \texttt{lift}        & EEF lifts the grasped object off the surface      \\
        5           & \texttt{move\_xy}    & Lateral transfer toward the bin                   \\
        6           & \texttt{lower}       & EEF lowers into the bin                           \\
        7           & \texttt{open}        & Gripper opens, object released                    \\
        8           & \texttt{retract}     & EEF moves away from the bin                       \\
        9           & \texttt{return}      & EEF returns to home configuration                 \\
        \bottomrule
    \end{tabular}
\end{table}

\begin{table}[htbp]
    \centering
    \caption{Task phases for Screwing.}
    \label{tab:phases_screw}
    \small
    \begin{tabular}{clp{8cm}}
        \toprule
        Phase index & Phase name         & Description                                            \\
        \midrule
        0           & \texttt{approach}  & EEF moves to a pre-contact waypoint above the fastener \\
        1           & \texttt{descend}   & EEF lowers to engage the fastener                      \\
        2           & \texttt{screw}     & Continuous downward force + wrist rotation to thread   \\
        3           & \texttt{disengage} & Gripper opens / tool lifts off                         \\
        4           & \texttt{retract}   & EEF returns to a safe height                           \\
        5           & \texttt{descend}   & EEF lowers to engage the fastener (loosening pass)     \\
        6           & \texttt{loosen}    & Loosen the screw                                       \\
        7           & \texttt{engage}    & Gripper closes on the retrieved screw                  \\
        8           & \texttt{retract}   & EEF returns to home position                           \\
        \bottomrule
    \end{tabular}
\end{table}

\begin{table}[htbp]
    \centering
    \caption{Task phases for Peg-in-Hole.}
    \label{tab:phases_peg}
    \small
    \begin{tabular}{clp{8cm}}
        \toprule
        Phase index & Phase name             & Description                                     \\
        \midrule
        0           & \texttt{above\_hole}   & EEF moves above the hole and aligns             \\
        1           & \texttt{insert}        & EEF pushes peg downward into the hole           \\
        2           & \texttt{disengage}     & Gripper opens                                   \\
        3           & \texttt{above\_hole}   & EEF rises back to the waypoint above the hole   \\
        4           & \texttt{retract}       & EEF returns to home position                    \\
        5           & \texttt{above\_object} & EEF moves above the object and aligns with it   \\
        6           & \texttt{engage}        & Gripper closes                                  \\
        7           & \texttt{above\_object} & EEF rises back to the waypoint above the object \\
        8           & \texttt{retract}       & EEF returns to home position                    \\
        \bottomrule
    \end{tabular}
\end{table}

\subsection{Episode metadata}

Alongside the sensor stream, every episode carries a fixed set of metadata fields describing the robot, task, condition, payload, and end-effector configuration (Table~\ref{tab:metadata}). These fields are constant within an episode and are used both as filters during question generation (e.g.\ to restrict a template to a specific payload regime) and as provenance in the released Q\&A pairs. This includes fault-specific fields (such as injection timestamp or physical offset).

\begin{table}[htbp]
    \centering
    \caption{Default per-episode metadata fields.}
    \label{tab:metadata}
    \small
    \begin{tabular}{lp{9.5cm}}
        \toprule
        Field                                 & Description                                                          \\
        \midrule
        \texttt{episode\_id}                  & Unique identifier for the episode                                    \\
        \texttt{robot\_model}                 & Robot model (UR3, KUKA KR10)                                         \\
        \texttt{task}                         & Task (pick-and-place, screwing, peg-in-hole)                         \\
        \texttt{condition}                    & Episode condition (normal, fault, counterfactual)                    \\
        \texttt{fault\_id}                    & Fault index for fault episodes; null otherwise                       \\
        \texttt{weight\_of\_box}              & Mass of the transported object in kg (pick-and-place only)           \\
        \texttt{shape\_of\_box}               & Shape/dimensions of the transported object (pick-and-place only)     \\
        \texttt{position\_of\_box}            & Initial XYZ position of the object on the table, where measurable    \\
        \texttt{gripper\_model}               & Gripper model and type (vacuum, two-finger, screwdriver)             \\
        \texttt{payload\_mass\_configured}    & Payload mass configured at the controller (kg)                       \\
        \texttt{payload\_cog\_configured}     & Payload CoG offset configured at the controller (x, y, z in mm)      \\
        \texttt{tcp\_offset\_configured}      & TCP position offset configured at the controller (x, y, z in mm)     \\
        \texttt{tcp\_orientation\_configured} & TCP orientation configured at the controller (rx, ry, rz in radians) \\
        \bottomrule
    \end{tabular}
\end{table}

\subsection{Fault taxonomy}

FactoryWave includes 27 distinct fault types (Table~\ref{tab:faults}). Each fault is paired with a plain-language description of the underlying physical or control-level mechanism and the injection procedure used to induce it during data collection. The last three columns indicate which tasks include the fault in the collected dataset; faults that apply to all three tasks (e.g.\ collisions with a shared workspace object, misconfigurations of the shared controller) are recorded under each applicable task with task-specific signal imprints.

    {\renewcommand{\arraystretch}{1.35}
        \begin{longtable}{@{}p{3.1cm}p{4.2cm}p{4.6cm}ccc@{}}
            \caption{Fault catalogue. Columns \textit{PP}, \textit{Scr}, \textit{PiH} indicate whether the fault is included in the Pick-and-Place, Screwing, and Peg-in-Hole recordings.}
            \label{tab:faults}                                                                                                                                                                                                                                                                                                                                                                                                                                                   \\
            \toprule
            Fault                             & Explanation                                                                                                                                                                                                     & Injection                                                                                                                                                               & PP         & Scr        & PiH        \\
            \midrule
            \endfirsthead
            \multicolumn{6}{l}{\small\textit{Table \ref{tab:faults} (cont.)}}                                                                                                                                                                                                                                                                                                                                                                                                    \\
            \toprule
            Fault                             & Explanation                                                                                                                                                                                                     & Injection                                                                                                                                                               & PP         & Scr        & PiH        \\
            \midrule
            \endhead
            Damaged screw thread              & Screw thread physically damaged, preventing proper engagement during tightening.                                                                                                                                & Pre-damaged the screw thread with sandpaper.                                                                                                                            & \texttimes & \checkmark & \texttimes \\
            Missing screw                     & Tightening is attempted with no screw present.                                                                                                                                                                  & Removed the screw from screwdriver before the cycle.                                                                                                                    & \texttimes & \checkmark & \texttimes \\
            Damaged plate thread              & The threaded hole in the plate is damaged, preventing engagement.                                                                                                                                               & Pre-damaged the plate hole with a metal screwdriver.                                                                                                                    & \texttimes & \checkmark & \texttimes \\
            Loosening phase                   & A counterclockwise loosening rotation replaces tightening; a normal phase with a distinct signal. It is counted here as an anomaly to match the fault schema of the AURSAD dataset \cite{leporowski2022aursad}. & Reversed the rotation direction in the controller program to execute a counterclockwise loosening pass in place of tightening.                                          & \texttimes & \checkmark & \texttimes \\
            Gripper activation failure        & The vacuum gripper fails to activate and never picks up the box.                                                                                                                                                & Disabled the gripper activation command in the robot program so the pick cycle completes without a grasp.                                                               & \checkmark & \texttimes & \texttimes \\
            Gripper release during motion     & The gripper releases the payload mid-trajectory, causing an abrupt payload loss.                                                                                                                                & Triggered a programmatic gripper-release command at a scripted timestep mid-trajectory.                                                                                 & \checkmark & \texttimes & \texttimes \\
            Additional axis payload           & A dead weight attached to one link increases inertia and gravity loading on all joints.                                                                                                                         & Bolted calibrated weights of varying mass to one robot link.                                                                                                            & \checkmark & \checkmark & \texttimes \\
            Collision with foam object        & Contact with a soft foam block produces a brief TCP force spike without a protective stop.                                                                                                                      & Placed a foam cube directly in the programmed TCP trajectory.                                                                                                           & \checkmark & \checkmark & \checkmark \\
            Unexpected payload weight         & The transported box has a weight that deviates from the nominal payload configuration.                                                                                                                          & Swapped the nominal box for a known heavier or lighter one. This is done in a counterfactual context, ie. switching weights and asking about alternate weight scenario. & \checkmark & \texttimes & \texttimes \\
            Invalid gripping position         & Timing error: the gripper closes after the arm has begun lifting, gripping the object off-center.                                                                                                               & Inserted a brief delay in the controller program so gripper closure lags the lift motion.                                                                               & \checkmark & \texttimes & \texttimes \\
            Unstable mounting platform        & Base instability introduces low-frequency vibrations into the arm.                                                                                                                                              & Placed 8 layers of towels under the base plate.                                                                                                                         & \checkmark & \texttimes & \texttimes \\
            Joint position limit violation    & A joint moves outside its configured soft or hard position limits, triggering a safety stop.                                                                                                                    & Random waypoint slightly beyond the configured soft joint limit.                                                                                                        & \checkmark & \texttimes & \texttimes \\
            TCP frame misconfiguration        & TCP frame or mounting orientation misconfigured at the controller; gravity torques deviate.                                                                                                                     & Entered an incorrect TCP offset or mounting angle in the controller installation settings.                                                                              & \checkmark & \checkmark & \checkmark \\
            Payload weight misconfiguration   & Payload mass misconfigured while a tool or workpiece is physically attached.                                                                                                                                    & Left the configured payload weight at multiple wrong mass configurations while a task-dependent gripper remained mounted.                                               & \checkmark & \checkmark & \texttimes \\
            External arm disturbance          & A continuous external force pulls or pushes the TCP during motion.                                                                                                                                              & Anchored an elastic rope between the bench frame and the tool flange.                                                                                                   & \checkmark & \checkmark & \checkmark \\
            Mild payload CoG misconfiguration & Payload CoG specified at an incorrect offset from the tool flange; gravity compensation is wrong.                                                                                                               & Entered a CoG offset in the payload configuration that differs from the physical tool CoG.                                                                              & \checkmark & \checkmark & \checkmark \\
            Collision with hanging cable      & A loose cable drags along a robot link or the tool flange during motion.                                                                                                                                        & Suspended a loose cable across the trajectory at arm height.                                                                                                            & \checkmark & \checkmark & \checkmark \\
            Collision with cardboard object   & A cardboard carton in the trajectory provides moderate resistance; may trigger a protective stop.                                                                                                               & Placed a free-standing or braced cardboard box in the programmed trajectory.                                                                                            & \checkmark & \checkmark & \checkmark \\
            Collision with rigid object       & A rigid plastic block in the TCP path triggers an immediate protective stop.                                                                                                                                    & Clamped a rigid plastic block in the programmed TCP path. Unlike the other collisions, this fault consistently causes safety stops.                                     & \checkmark & \checkmark & \checkmark \\
            Peg insertion misalignment        & Peg approaches the hole at an angular or lateral offset and jams against the rim.                                                                                                                               & Added a recorded lateral offset at the insertion approach waypoint.                                                                                                     & \texttimes & \texttimes & \checkmark \\
            Hole obstruction                  & Foreign object or debris in the hole prevents full insertion.                                                                                                                                                   & Placed a small object (metal screw) inside the hole before insertion.                                                                                                   & \texttimes & \texttimes & \checkmark \\
            Incorrect insertion depth         & Insertion terminates at an incorrect Z depth due to a misconfigured waypoint.                                                                                                                                   & Shifted the insertion endpoint waypoint Z by recorded offset from nominal.                                                                                              & \texttimes & \texttimes & \checkmark \\
            Peg surface contamination         & Contamination or roughness on the peg raises insertion friction and produces stick-slip.                                                                                                                        & Applied dry chalk powder to the peg surface.                                                                                                                            & \texttimes & \texttimes & \checkmark \\
            Fixture displacement              & Insertion fixture shifted laterally, breaking the match between programmed approach and actual hole.                                                                                                            & Physically shifted the hole fixture by recorded offset without updating the robot program.                                                                              & \texttimes & \texttimes & \checkmark \\
            Self-collision                    & Arm configuration causes a link to contact the robot body or mounting fixture, triggering a protective stop.                                                                                                    & Random waypoint that forces a self-colliding configuration or offset the mounting fixture into the trajectory.                                                          & \checkmark & \texttimes & \texttimes \\
            Missing box                       & Box absent from the pick position; the gripper closes on empty air.                                                                                                                                             & Removed the box from the pick position before the episode started.                                                                                                      & \checkmark & \texttimes & \texttimes \\
            Missing peg                       & Arm descends into the hole empty-handed; insertion produces no contact force.                                                                                                                                   & Removed the peg from the gripper before the episode started.                                                                                                            & \texttimes & \texttimes & \checkmark \\
            \bottomrule
        \end{longtable}
    }

\subsection{Counterfactual episodes}

Counterfactual pairs are constructed for the subset of faults in Table~\ref{tab:faults} whose injection moment can be controlled at a specific timestep rather than being present throughout the episode. For each such fault, the initial conditions of the physical setup (object identity and pose, payload configuration, controller settings) are held as constant as the platform allows, one baseline execution is recorded without any injection, and several fault executions are then recorded under the same initial conditions with the fault applied at a shared sampled timestep. Because the physical system is not perfectly reproducible, the pre-injection sensor trajectories of the baseline and the fault candidates diverge slightly; we retain, as the counterfactual partner of the baseline, the fault candidate whose pre-injection window is closest to the baseline under the signature kernel MMD, and discard the others. This yields an approximate counterfactual pair in which the early history is as matched as the real platform permits, while the post-injection divergence isolates the effect of the intervention. The approximation error introduced by this procedure, relative to the exact counterfactuals available in simulation, motivates the sim2real analysis in Section~\ref{app:simulation_setup}.

\subsection{Signal schema}

The UR3 is recorded at 125\,Hz, yielding 136 per-sample signals grouped by semantic role (Table~\ref{tab:ur3_signals}): controller \emph{setpoints}, actual \emph{effort / feedback} readings from joints and the TCP, \emph{context / health} signals covering joint temperatures, voltages, and internal controller state, a small block of controller \emph{status / mode} indicators, and digital/analog \emph{I/O} lines. A handful of episode-level provenance fields (episode identifier, robot type, task, fault label, payload mass, recording date) are appended per episode but are constant within an episode.

\begin{table}[htbp]
    \centering
    \caption{UR3 signal groups recorded at 125\,Hz (excluding episode-level provenance columns).}
    \label{tab:ur3_signals}
    \small
    \begin{tabular}{@{}lrp{7.5cm}@{}}
        \toprule
        Group             & \# signals   & Contents                                                                                                                                                      \\
        \midrule
        Setpoint (intent) & 42           & Target joint position, velocity, acceleration, current, torque; target TCP pose and speed.                                                                    \\
        Effort / feedback & 48           & Actual joint position, velocity, current, current-as-torque, joint control output; actual TCP pose, speed, and force/torque.                                  \\
        Context / health  & 33           & Joint temperatures, joint voltages, joint modes, tool accelerometer, raw F/T wrench, main/robot voltage and current, momentum, speed scaling, execution time. \\
        Status / mode     & 7            & Timestamp, robot mode, robot status, safety mode, safety status bits, runtime state, configured payload mass.                                                 \\
        I/O               & 6            & Digital inputs/outputs, two analog inputs, two analog outputs.                                                                                                \\
        \midrule
        \textbf{Total}    & \textbf{136} &                                                                                                                                                               \\
        \bottomrule
    \end{tabular}
\end{table}

The KUKA KR10 is controlled through the KRC4 controller. We stream 53 per-sample signals at 83\,Hz, grouped by semantic role in Table~\ref{tab:kuka_signals} using the same partition as the UR3 schema. A handful of signals that the UR3 exposes natively are unavailable on KSS 8.3, specifically joint velocities, commanded TCP pose, joint voltage, and joint-level control outputs; TCP force/torque is also absent because no force sensor is fitted. To recover tool-side feedback, the collection setup adds a 9-DoF inertial measurement unit on the gripper, logging its triaxial linear acceleration (\texttt{acc\_x/y/z}), angular rate (\texttt{gyro\_x/y/z}), and orientation (\texttt{angle\_x/y/z}) at the controller's 83\,Hz cadence.

\begin{table}[htbp]
    \centering
    \caption{KUKA KR10 signal groups recorded at 83\,Hz via RSI/KRL (excluding episode-level provenance columns).}
    \label{tab:kuka_signals}
    \small
    \begin{tabular}{@{}lrp{7.5cm}@{}}
        \toprule
        Group             & \# signals  & Contents                                                                                                                                                                           \\
        \midrule
        Setpoint (intent) & 6           & Commanded joint positions (degrees).                                                                                                                                               \\
        Effort / feedback & 24          & Actual joint positions (degrees), actual TCP pose (mm / degrees), per-axis motor current (\% of max), and per-axis motor torque (Nm).                                              \\
        Context / health  & 11          & Per-axis motor temperature (Kelvin), Cartesian acceleration on X, Y, Z and absolute magnitude (m/s$^2$), and speed override (\%).                                                  \\
        Tool IMU          & 9           & Gripper-mounted 9-DoF IMU: triaxial linear acceleration \texttt{acc\_x/y/z} (m/s$^2$), angular rate \texttt{gyro\_x/y/z} (rad/s), and orientation \texttt{angle\_x/y/z} (degrees). \\
        Status / mode     & 1           & Controller process state (FREE / ACTIVE / STOP / END).                                                                                                                             \\
        I/O               & 2           & Digital inputs bitmask and digital outputs bitmask.                                                                                                                                \\
        \midrule
        \textbf{Total}    & \textbf{53} &                                                                                                                                                                                    \\
        \bottomrule
    \end{tabular}
\end{table}

\section{Time-series foundation model baselines}
\label{app:chronos-baseline}

A natural concern about a benchmark whose motivation references \emph{time-series models} is whether the LLM panel is the right comparison set. To address it we evaluate two pretrained time-series foundation models (TSFMs) of comparable parameter count, drawn from architecturally distinct families with broadly disjoint pretraining corpora: \textsc{Chronos-Bolt}~\cite{ansari2024chronos} (\texttt{amazon/chronos-bolt-base}, $\sim$200\,M parameters, encoder--decoder with a 9-quantile probabilistic head) and \textsc{TimesFM-2.5}~\cite{das2024decoder} (\texttt{google/timesfm-2.5-200m-pytorch}, $\sim$200\,M parameters, decoder-style with separate point and quantile heads). Both are evaluated zero-shot, with no fine-tuning, prompting, or context-window engineering on either side, under one shared parsing, scoring, and chance-correction pipeline. Evaluating two rather than one is deliberate: a single TSFM result invites the objection that the numbers are model-specific, whereas two independently pretrained TSFMs converging on the same per-template, per-signal, and per-horizon profile is a statement about the benchmark rather than about a checkpoint.

\paragraph{Template applicability.} A pure forecaster has neither a language interface nor a notion of options, ranking, or counterfactual conditioning. Of the 21 active question templates in FactoryBench, only three reduce to a well-posed forecasting problem: L1.7 (scalar forecast), L2.4 (scalar forecast under a known anomaly), and L2.5 (six-joint tensor forecast under a known anomaly). The remaining 18 templates (MCQ classification, ranking, counterfactual conditioning, free-form root-cause and protocol writing) are structurally outside the scope of any TSFM that lacks a language head. \textbf{This 14\,\% template coverage is itself a finding}: any TSFM, regardless of size or pretraining quality, can address only the forecasting slice of FactoryBench, and the other 86\,\% of the benchmark measures capabilities that no current TSFM provides.

\paragraph{Shared methodology.} Items are filtered to the three applicable templates, giving $n = 334 + 74 + 61 = 469$. The text-encoded series carried in the LLM prompt is reverse-parsed back into numeric arrays through the acronym mapping stored in \texttt{context.time\_series\_format}; target signals and horizons come from \texttt{acceptance\_bounds} (for L2.4, the \texttt{T+\{n\}ms} pattern in the question text, converted to integer steps via the median sample interval), and for L2.5 the base signal is expanded into its six per-joint columns. Scoring uses the same piecewise-margin rule as the main paper (Appendix~\ref{app:answer-formats}): score 1 if $|\hat v - v^\star| \le m$, $0.5$ if $|\hat v - v^\star| \le 2m$, else 0, with the signed chance correction $(s-E)/(1-E)$ of Appendix~\ref{app:answer-formats} at $E=1/4$, so an item scored $0$ contributes $-1/3$ rather than being floored. Confidence intervals are 95\,\% percentile intervals from 1000 bootstrap resamples at a fixed seed. Parsing, scoring, chance correction, and the bootstrap are held fixed across the two models; the only thing that varies is the forecaster.

\paragraph{The L1.7 margin caveat.} L1.7 items in the released dataset do \emph{not} carry a \texttt{margin} field, so the scorer falls through to exact match ($|\hat v - v^\star| < 10^{-4}$), which is unattainable for a real-valued forecast. We therefore score both TSFMs under \emph{both} conventions: the intended calibration $m = R/12$ that puts L1.7 at the same chance level as the rest of the benchmark, and the strict exact-match fallback. Both appear in Table~\ref{tab:chronos-results}. The missing margin is a generation-pipeline oversight rather than a deliberate scoring choice; either shipping the margin in the released items or re-running the scorer at $R/12$ closes the discrepancy.

\paragraph{Per-model inference deltas.} For Chronos-Bolt we form a univariate \texttt{float32} context per target channel and take the median quantile at the last predicted step; for tensor answers we run six independent per-joint forecasts and concatenate. TimesFM-2.5 returns a \texttt{(point, quantile)} tuple and we use the published point forecast, its canonical scalar output; it additionally requires a one-time \texttt{compile()} with bounds $(512, 64)$, which subsumes every FactoryBench item ($T \le 77$, $h \le 10$). Neither model has a native multivariate mode, so cross-joint correlations are unavailable to both by construction.

\begin{table}[htbp]
    \centering
    \caption{Zero-shot performance of \textsc{Chronos-Bolt} and \textsc{TimesFM-2.5} on the three forecasting templates of FactoryBench ($n=469$). Confidence intervals are 95\,\% percentile intervals from 1000 bootstrap resamples. Chance-corrected (CC) scores apply the signed $\tilde{s}=(s-1/4)/(1-1/4)$, the same rule as the main results. The two L1.7 rows per model differ only in the margin convention: $m=R/12$ is the intended calibration that puts L1.7 on the same chance level as the rest of the benchmark, while the exact-match fallback is what the scorer applies in the absence of a stored margin field.}
    \label{tab:chronos-results}
    \small
    \begin{tabular}{@{}llr cc cc@{}}
        \toprule
        Model        & Template                                        & $n$ & Raw mean & Raw 95\% CI       & CC mean & CC 95\% CI        \\
        \midrule
        Chronos-Bolt & L1.7 (state, scalar; $m=R/12$ margin)           & 334 & 0.636    & [0.588,\ 0.681]   & 0.515   & [0.451,\ 0.575]   \\
                     & L2.4 (intervention, scalar)                     &  74 & 0.601    & [0.493,\ 0.710]   & 0.469   & [0.324,\ 0.613]   \\
                     & L2.5 (intervention, tensor[6])                  &  61 & 0.623    & [0.537,\ 0.708]   & 0.497   & [0.383,\ 0.610]   \\
                     & L1.7 (state, scalar; exact-match fallback)      & 334 & 0.027    & [0.012,\ 0.045]   & 0.000   & ---               \\
        \midrule
        TimesFM-2.5  & L1.7 (state, scalar; $m=R/12$ margin)           & 334 & 0.672    & [0.626,\ 0.714]   & 0.563   & [0.501,\ 0.619]   \\
                     & L2.4 (intervention, scalar)                     &  74 & 0.601    & [0.500,\ 0.696]   & 0.469   & [0.333,\ 0.595]   \\
                     & L2.5 (intervention, tensor[6])                  &  61 & 0.626    & [0.540,\ 0.706]   & 0.501   & [0.386,\ 0.608]   \\
                     & L1.7 (state, scalar; exact-match fallback)      & 334 & 0.036    & [0.018,\ 0.057]   & 0.000   & ---               \\
        \bottomrule
    \end{tabular}
\end{table}

\paragraph{Results.} Under the calibrated $R/12$ margin, Chronos-Bolt reaches chance-corrected accuracy of $51.5\%$, $46.9\%$, and $49.7\%$ on L1.7, L2.4, and L2.5; TimesFM-2.5 reaches $56.3\%$, $46.9\%$, and $50.1\%$ on the same templates. For reference, the strongest LLM in the panel is Claude Sonnet~4.6 at $+11.8\%$ on the L1 level aggregate and $+14.6\%$ on the L2 aggregate (Section~\ref{sec:zero-shot}). \textbf{That comparison is level-versus-template and must not be read as a head-to-head.} It differs in two ways at once: the LLM aggregates average over every L1 and L2 template, of which only L1.7, L2.4 and L2.5 are shared with the TSFMs; and the panel is scored on the balanced FactoryBench-Lite subset (Appendix~\ref{app:lite}) whereas this slice is drawn from the undivided released pool. A per-template re-score of the panel on the identical items would give the strict head-to-head, but the per-reply files that operation needs are not part of the released artifact bundle. What the numbers do support is the weaker and still consequential claim that on the forecasting slice a 200\,M-parameter pretrained forecaster is not obviously outclassed by frontier LLMs.

\paragraph{Head-to-head between the two TSFMs.} The within-table comparison of Chronos-Bolt and TimesFM-2.5 \emph{is} apples-to-apples in every respect except the model, since both see the same items under the same scorer. They agree to within bootstrap CI overlap on every template: TimesFM-2.5 holds a small lead on L1.7 ($0.563$ vs $0.515$ chance-corrected, CIs $[0.501, 0.619]$ and $[0.451, 0.575]$, overlapping on $[0.501, 0.575]$) and is statistically indistinguishable on L2.4 and L2.5. Both collapse to $\le 4\%$ raw accuracy on L1.7 under the exact-match fallback, which confirms that the missing L1.7 margin is a scoring-pipeline artefact and not a model-capability result.

\paragraph{Score distribution, signal coverage, and horizon.} Per-item scores are strongly bimodal for both models, landing at $1.0$ or $0.0$ with a smaller mass at the half-credit band, as expected from a margin-thresholded scorer rather than a graded loss; the small fractional masses on L2.5 come from averaging six per-joint scores. Aggregated by target-signal family, both models are roughly uniformly capable across positions ($n=270$; $0.620$ and $0.633$ raw mean), speeds ($n=140$; $0.630$ and $0.701$), torques and efforts ($n=32$; $0.719$ and $0.719$), and contact forces ($n=27$; $0.611$ and $0.556$), all well clear of chance, so neither baseline selectively succeeds on smooth signals and fails on noisy ones. Both degrade monotonically with forecast horizon, from roughly $0.77$--$0.78$ raw at one step to $0.54$--$0.55$ at ten, with TimesFM-2.5 slightly ahead in the four-to-eight step band and tied or marginally behind at the extremes; at the longest horizons the bootstrap intervals overlap chance for both. The two models produce the \emph{same shape} of per-signal and per-horizon profile despite disjoint pretraining, which we read not as the models being equivalent in detail but as evidence that FactoryBench's forecasting items carry a model-class-intrinsic difficulty profile rather than an artefact of one architecture.

\paragraph{Inference cost.} On a single NVIDIA A10G, Chronos-Bolt processes the 469-item slice in $24.1$\,s (mean $0.05$\,s/item, including model loading) and TimesFM-2.5 in $88.4$\,s (mean $0.19$\,s/item), the difference coming from TimesFM's larger fixed-length compiled graph. Both are two to three orders of magnitude cheaper than running an LLM panel over the same items, which makes either usable as a standing reference number in future versions of the benchmark.

\paragraph{Fairness of comparison.} The TSFM-versus-LLM comparison is apples-to-apples on scoring rule, chance correction, and zero-shot setting, but \emph{not} on (i) input representation: LLMs see acronymized text-encoded values alongside the question, robot description, and option set, while both TSFMs see raw univariate float arrays only; nor on (ii) question awareness: on L2.4 and L2.5 the LLM is told which anomaly is present, information neither TSFM can use. A multivariate or text-conditioned TSFM (e.g.\ Moirai-MoE, Time-MoE) might exploit these signals; we do not, to keep the baselines architecturally minimal. We therefore report TSFM numbers as template-specific rather than as level aggregates.

\paragraph{Limitations.} (i)~Two TSFMs is more than one but still a small sample of the design space; Moment, Lag-Llama, and Moirai are the obvious follow-ups on the same plumbing. (ii)~FactoryBench contexts run 32--77 timesteps while both models were pretrained on contexts up to 512, so these baselines likely under-use each model's effective receptive field. (iii)~Until the L1.7 margin discrepancy is closed, the L1.7 rows should be read in pairs, one per scoring convention. (iv)~For TimesFM-2.5 we use the published point forecast; reading the median quantile instead is numerically very close in spot checks but we have not characterised the difference at population level.

\paragraph{Interpretation.} Two messages emerge. \textbf{First, on the templates where forecasting is the right tool, a 200\,M-parameter pretrained forecaster is competitive with frontier LLMs}, consistent with the main paper's observation that LLMs struggle with precise numerical state extraction at L1, and the finding is not a property of one checkpoint: two independently pretrained TSFMs land on the same numbers. This opens the door to tool-assisted LLM agents that delegate forecasting subtasks to a specialised TSFM and reserve the LLM for the linguistic, classification, and decision-making templates where it has the structural advantage, a design we build and measure in Appendix~\ref{app:agent-baseline}. \textbf{Second, the remaining 18 of 21 FactoryBench templates (classification, ranking, counterfactual reasoning, free-form troubleshooting and optimization) are structurally outside any current TSFM's scope.} Together these reinforce the paper's central claim: the gap between current models and operational machine understanding is not closed by solely scaling LLMs or deploying purpose-built forecasters.

\paragraph{Reproducibility.} The complete pipeline is in \texttt{experiments/v1/}. \texttt{download\_data.py} fetches the JSONL files, \texttt{parse\_qa.py} reverse-parses the encoded series and identifies target channels and horizons, \texttt{run\_chronos2.py} and \texttt{run\_timesfm.py} run the two models and score each item, and \texttt{analyze.py} (via \texttt{--results} and \texttt{--prefix}) computes the bootstrap intervals and the signal- and horizon-level breakdowns for either. \texttt{report\_full.qmd} and \texttt{report\_timesfm.qmd} render the full reports. End to end on a single A10G this is under one minute for Chronos-Bolt and under two for TimesFM-2.5, model loading included.

\section{Tool-augmented LLM agent baseline}
\label{app:agent-baseline}

The zero-shot panel isolates the LLM's ability to read industrial signals from text alone; the Chronos-Bolt baseline isolates a specialised forecaster on the templates it can address. Neither captures the design most likely to be deployed in practice: an LLM driver that delegates numerical subtasks to purpose-built tools and reserves its own capacity for symbolic reasoning. We instantiate that design as a ReAct-style~\cite{yao2023react} agent built on GPT-5.1 with four tools and evaluate it on the same benchmark.

The composition is deliberately unambitious. Every component is off the shelf, the tools are the obvious ones for the task, the driver is capped at eight calls per item, and no part of the system is tuned on FactoryBench. We built it this way on purpose: a strong bespoke agent would leave open whether its gains came from tool use or from engineering effort invested in the agent itself, whereas gains from a pipeline this plain are attributable to composition alone, and are a lower bound on what the approach can deliver.

\paragraph{Pipeline.} The agent takes a FactoryBench item as input, decides which tools (if any) to invoke, executes them, observes their outputs, and emits a final answer in the item's expected format. It has access to four tools:
(i)~a \emph{signal-statistics} tool returning per-channel summaries (min, max, mean, standard deviation, $p_5$, $p_{95}$, derivative) computed from the item's time-series context;
(ii)~a \emph{forecaster}, the same \textsc{Chronos-Bolt} (200\,M) model used in Appendix~\ref{app:chronos-baseline}, called on a single channel with a horizon the agent chooses;
(iii)~a \emph{numerical sandbox}, a restricted Python interpreter with standard scientific-computing libraries and the item's time series pre-loaded;
(iv)~a \emph{manual retriever}, a dense-vector index over 9 vendor PDFs (UR3e, KUKA KR6/KR10 AGILUS, voraus-AD; 1{,}762 chunks embedded in 3072 dimensions), returning the three closest chunks for a natural-language query.
The driver LLM (GPT-5.1) invokes tools through a structured tool-calling interface, is capped at 8 tool calls per item, and receives a system prompt that maps question shapes to preferred tools: numerical prediction to the forecaster, segment localisation to the numerical sandbox, and root-cause or remediation questions to the manual retriever.

\paragraph{Evaluation protocol.} We evaluate on a stratified subset of FactoryBench-Lite: $25$ items per question template (seed 42), giving $525$ items in total and a per-template floor that keeps each template's estimate usable rather than only the level aggregate. The agent and zero-shot GPT-5.1 answer the \emph{same} items, so the comparison is paired throughout, and we report bootstrap $95\%$ confidence intervals over the per-item differences rather than over each arm separately. As a check on representativeness, zero-shot on this subset reproduces its full-set level scores to within $3$ points on L1--L3.

\paragraph{Results.} Table~\ref{tab:agent-results} reports the paired comparison. The headline is that tool access is close to a wash: $-2.9$ pp overall, with a confidence interval spanning zero ($[-7.4, +1.6]$). Underneath that average the tools do two opposite things. Where the task is computational, they help: L1 gains $+9.8$ pp, carried by segment localisation (L1.1, $+21.3$ pp) and multi-select state reading (L1.3, $+20.0$ pp), both of which the numerical sandbox can attack directly and neither of which invokes the forecaster. Where the agent delegates its extrapolation, they hurt: L3 loses $10.5$ pp with the interval excluding zero, and the loss is concentrated in the two templates that call the forecaster on every item, L3.4 ($-26.7$ pp) and L3.5 ($-16.0$ pp). Inspecting the traces shows the mechanism is literal rather than statistical: on L3.4 the agent's final answer is character-for-character the forecaster's \texttt{predicted\_value\_at\_horizon}, so a call to \textsc{Chronos-Bolt} replaces the model's own estimate outright. Appendix~\ref{app:chronos-baseline} measures that forecaster near chance on these templates, which is why the substitution costs accuracy. The effect is not uniform across every forecasting template, L2.4 and L2.5 come out marginally ahead ($+2.7$ pp each), so the reliable statement is narrower than \emph{tools hurt}: delegation is harmful exactly where the delegate is weaker than the caller.

\begin{figure}[htbp]
    \centering
    \includegraphics[width=\linewidth]{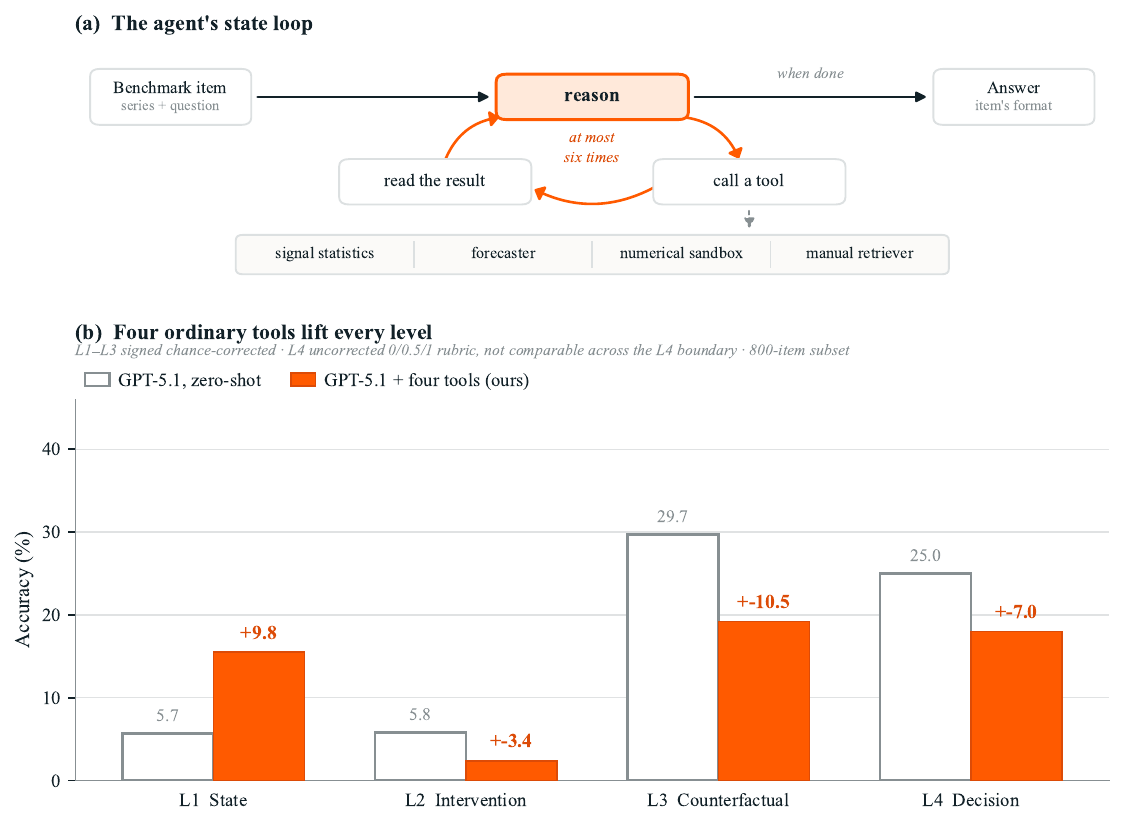}
    \caption{\textbf{A three-step loop, and what it costs.} (a)~The agent as a state machine. The driver cycles through three states, reason, call a tool, read the result, and re-enters the cycle at most eight times before leaving it to answer. Tools are reachable only from the calling state. Nothing is tuned on FactoryBench. (b)~Accuracy per level for GPT-5.1 zero-shot and for the same model given those tools, on a paired 525-item stratified subset; the sandbox gains on L1 while delegated forecasting loses on L3, and the net effect across levels is a wash. L1--L3 are signed chance-corrected; L4 uses the uncorrected free-form judge rubric and is not comparable across the L4 boundary.}
    \label{fig:agent-pipeline}
\end{figure}

Figure~\ref{fig:agent-pipeline} shows the state loop and the per-level lift.

\paragraph{Tools help where computation helps, and hurt where the agent defers.} We expected a clear lift and did not get one: four off-the-shelf tools leave GPT-5.1 statistically level overall. That average hides the useful result, which is that the two halves of the pipeline behave in opposite directions. The numerical sandbox pays for itself on the templates where the bottleneck is arithmetic over a long window, and it is worth around $20$ points on both L1 templates that use it. The forecaster does the reverse: the driver routes every extrapolation subtask to it and then adopts its point estimate verbatim, discarding its own reasoning in favour of a $200$\,M model that this appendix elsewhere measures near chance on the same templates. The practical lesson is not that tool use is unhelpful on FactoryBench but that it is only as good as the delegation policy: a tool is worth calling when it is better than the caller at that subtask, and this agent has no mechanism for deciding whether that holds. It never consults the forecaster's own uncertainty, which the tool returns alongside the point estimate, and it never checks a tool answer against its prior before accepting it. Both are cheap to add, which makes this a concrete target rather than a capability ceiling. We report the null result because the pipeline we built is the one a practitioner would reach for first.

\begin{table}[htbp]
    \centering
    \caption{Tool-augmented GPT-5.1 agent vs.\ GPT-5.1 zero-shot on FactoryBench, evaluated on a paired stratified subset of 25 items per template (525 items, seed 42). L1--L3 report signed chance-corrected accuracy ($\tilde{s} = (s-E)/(1-E)$); L4 reports raw judge-ensemble accuracy under the L4 rubric (uncorrected, per Appendix~\ref{app:answer-formats}).}
    \label{tab:agent-results}
    \begin{tabular}{@{}l r r r r@{}}
        \toprule
        \textbf{Level} & \textbf{GPT-5.1 zero-shot} & \textbf{GPT-5.1 agent} & $\boldsymbol{\Delta}$ & \textbf{95\% CI} \\
        \midrule
        L1 (State, signed CC)            & $+5.7\%$  & $\mathbf{+15.5\%}$ & $+9.8$  & $[-1.0, +20.5]$  \\
        L2 (Intervention, signed CC)     & $\mathbf{+5.8\%}$  & $+2.4\%$  & $-3.4$  & $[-10.0, +3.1]$  \\
        L3 (Counterfactual, signed CC)   & $\mathbf{+29.7\%}$ & $+19.2\%$ & $-10.5$ & $[-20.4, -0.7]$  \\
        L4 (Decision, raw judge)         & $\mathbf{25.0\%}$  & $18.0\%$  & $-7.0$  & $[-19.0, +4.0]$  \\
        \midrule
        All levels                       & $\mathbf{+13.3\%}$ & $+10.4\%$ & $-2.9$  & $[-7.4, +1.6]$   \\
        \bottomrule
    \end{tabular}
\end{table}

\paragraph{Tool usage.} Across the $525$-item evaluation the agent issued $471$ tool calls, dominated by the forecaster ($370$), with the numerical sandbox second ($91$) and signal statistics and the manual retriever almost unused ($9$ and $1$). The distribution is itself informative: the driver spends most of its budget on the one tool that costs it accuracy, and effectively ignores the retriever even on L4, where vendor protocol text is the evidence the rubric rewards. A better policy would invert those proportions.

\section{Example question-answer pairs and question templates}
\label{app:qa_examples}

Table~\ref{tab:qa_examples} presents seventeen representative Q\&A pairs spanning every causal level and every answer format the generator produces (single-select MCQ, multi-select MCQ, tensor, ranking, free-form). For readability we omit the underlying sensor context (tens to hundreds of time-series rows), truncate verbose boilerplate such as ``Answer only with $\ldots$'', and abbreviate long option strings.

\begin{small}
    \begin{longtable}{@{}>{\raggedright\arraybackslash}p{0.05\textwidth}>{\raggedright\arraybackslash}p{0.15\textwidth}>{\raggedright\arraybackslash}p{0.74\textwidth}@{}}
        \caption{Representative Q\&A pairs from FactoryBench spanning all four causal levels and every answer format.}\label{tab:qa_examples}                                     \\
        \toprule
        \textbf{Level} & \textbf{Answer format} & \textbf{Prompt, options, and reference answer}                                                                                  \\
        \midrule
        \endfirsthead
        \multicolumn{3}{@{}l}{\footnotesize\emph{Table~\ref{tab:qa_examples} continued from previous page}}                                                                       \\
        \toprule
        \textbf{Level} & \textbf{Answer format} & \textbf{Prompt, options, and reference answer}                                                                                  \\
        \midrule
        \endhead
        \midrule
        \multicolumn{3}{r@{}}{\footnotesize\emph{continued on next page}}                                                                                                         \\
        \endfoot
        \bottomrule
        \endlastfoot

        \textbf{L1}    & Tensor                 &
        \textbf{Q:} The robot is performing a screwing task. We want to isolate the screwing phase in the robot's time series. Assuming a fixed window length of 10 timesteps, at which timestamp should the window begin? \newline
        \textbf{A:} \texttt{14}~~(accepted within $\pm3$)                                                                                                                         \\
        \midrule

        \textbf{L1}    & Single-select MCQ      &
        \textbf{Q:} What changed between the two instances of robotic time series data? \newline
        \textbf{Options:} \newline
        \hspace*{1.2em}\begin{tabular}{@{}l@{\,\,}p{0.63\textwidth}@{}}
                           \textbf{a.} & different anomalous states (exactly one anomalous, or both but distinct) \\
                           \textbf{b.} & different tasks                                                          \\
                           \textbf{c.} & different robots                                                         \\
                           \textbf{d.} & same task, different phases                                              \\
                       \end{tabular} \newline
        \textbf{A:} \texttt{a}                                                                                                                                                    \\
        \midrule

        \textbf{L1}    & Single-select MCQ      &
        \textbf{Q:} What robot does this sensor data originate from? \newline
        \textbf{Options:} \newline
        \hspace*{1.2em}\begin{tabular}{@{}l@{\,\,}p{0.63\textwidth}@{}}
                           \textbf{a.} & Agile Robots Yu~5 Industrial \\
                           \textbf{b.} & KUKA KR~10 R1100-2           \\
                           \textbf{c.} & Universal Robots UR3         \\
                       \end{tabular} \newline
        \textbf{A:} \texttt{c}                                                                                                                                                    \\
        \midrule

        \textbf{L1}    & Tensor                 &
        \textbf{Q:} Given the sensor stream below, what is the expected value of the position of joint~2 at T+606\,ms? \newline
        \textbf{A:} \texttt{102.7777}                                                                                                                                             \\
        \midrule

        \textbf{L2}    & Tensor                 &
        \textbf{Q:} The sensor stream below is from a robot exhibiting a collision with a cardboard object. What is the expected value of the velocity of joint~3 at T+68\,ms? \newline
        \textbf{A:} \texttt{-0.272044}~~(accepted within $\pm2.42$)                                                                                                               \\
        \midrule

        \textbf{L2}    & Tensor                 &
        \textbf{Q:} The sensor stream below is from a robot exhibiting a collision with a cardboard object. What are the expected values of joint positions at T+506\,ms? \newline
        \textbf{A:} \texttt{[17.884, -115.897, 121.049, -95.595, -90.258, 293.860]}                                                                                               \\
        \midrule

        \textbf{L2}    & Tensor                 &
        \textbf{Q:} Knowing that the robot suffers from a hole obstruction, we want to isolate the ``rise back above the object'' phase. Assuming a fixed window length of 16 timesteps, at which timestamp should the window begin? \newline
        \textbf{A:} \texttt{202}~~(accepted within $\pm3$)                                                                                                                        \\
        \midrule

        \textbf{L2}    & Single-select MCQ      &
        \textbf{Q:} Given the sensor data, determine what anomaly is present. \newline
        \textbf{Options:} \newline
        \hspace*{1.2em}\begin{tabular}{@{}l@{\,\,}p{0.63\textwidth}@{}}
                           \textbf{a.} & sustained command/measurement deviation                                                  \\
                           \textbf{b.} & no anomaly                                                                               \\
                           \textbf{c.} & measured joint speeds drop abruptly with a TCP force spike (typical collision signature) \\
                           \textbf{d.} & isolated unrealistic force-sensor spike inconsistent with motion                         \\
                       \end{tabular} \newline
        \textbf{A:} \texttt{c}                                                                                                                                                    \\
        \midrule

        \textbf{L3}    & Ranking                &
        \textbf{Q:} Given the baseline stream and the counterfactual scenario where a collision foam object occurs at timestep 7366\,ms, rank the four labeled signal segments in the order they would appear as the event manifests. \newline
        \textbf{A:} \texttt{abdc}                                                                                                                                                 \\
        \midrule

        \textbf{L3}    & Multi-select MCQ       &
        \textbf{Q:} Given the counterfactual scenario where a collision hanging cable occurs at timestep 8363\,ms, select all statements that would apply. \newline
        \textbf{Options:} \newline
        \hspace*{1.2em}\begin{tabular}{@{}l@{\,\,}p{0.63\textwidth}@{}}
                           \textbf{a.} & motor current rises $\geq 33\%$ while speed stays $\leq 10\%$ of baseline (stall-like) \\
                           \textbf{b.} & contact-force spike $\geq 36\%$ above baseline                                         \\
                           \textbf{c.} & tracking-error rise $\geq 34\%$ above pre-event mean                                   \\
                           \textbf{d.} & at least one joint speed drops sharply ($\geq 31\%$)                                   \\
                       \end{tabular} \newline
        \textbf{A:} \texttt{FFFT}                                                                                                                                                 \\
        \midrule

        \textbf{L3}    & Tensor                 &
        \textbf{Q:} In the counterfactual scenario where a collision foam object occurs at timestep 7659\,ms, what would be the expected value of the position of joint~2 at T+100\,ms? \newline
        \textbf{A:} \texttt{72.144939}~~(accepted within $\pm2.77$)                                                                                                               \\
        \midrule

        \textbf{L4}    & Free form              &
        \textbf{Q:} Given the sensor stream below, does the machine show signs of anomalous behavior? If yes, identify the most likely root cause and describe the steps you would take to fix it. \newline
        \textbf{A:} No anomalous behavior detected; the machine is operating normally; no remediation is required.                                                                \\
        \midrule

        \textbf{L4}    & Free form              &
        \textbf{Q:} Given the sensor stream below, does the machine show signs of anomalous behavior? If yes, identify the most likely root cause and describe the steps you would take to fix it. \newline
        \textbf{A:} \newline
        \hspace*{1.2em}\begin{tabular}{@{}l@{\,\,}p{0.63\textwidth}@{}}
                           \textbf{a.} & Inspect the workspace and remove any soft obstructions from the programmed trajectory.                        \\
                           \textbf{b.} & If a protective stop occurred, acknowledge it in the robot controller and verify arm posture before resuming. \\
                           \textbf{c.} & Reduce speed if repeated false-positive collision detections occur.                                           \\
                       \end{tabular}                            \\
        \midrule

        \textbf{L4}    & Free form              &
        \textbf{Q:} Given the sensor stream below, does the machine show signs of anomalous behavior? If yes, identify the most likely root cause and describe the steps you would take to fix it. \newline
        \textbf{A:} \newline
        \hspace*{1.2em}\begin{tabular}{@{}l@{\,\,}p{0.63\textwidth}@{}}
                           \textbf{a.} & Stop the robot and update the payload configuration to include the additional axis weight.    \\
                           \textbf{b.} & Set the correct CoG offset for the modified arm in the installation settings.                 \\
                           \textbf{c.} & Reduce motion accelerations and speeds to account for increased effective inertia.            \\
                           \textbf{d.} & Run a slow-speed test cycle to verify dynamic loads stay within rated limits before resuming. \\
                       \end{tabular}                                            \\
        \midrule

        \textbf{L4}    & Free form              &
        \textbf{Q:} An engineer wants to increase the effectiveness and accuracy of this machine. Based on the sensor stream below, what operational changes or parameter adjustments could help achieve this? \newline
        \textbf{A:} The payload mass is set to 0.0\,kg but the actual payload weighs 1.5\,kg. Update the payload mass in the installation settings to 1.5\,kg.                    \\
        \midrule

        \textbf{L4}    & Free form              &
        \textbf{Q:} An engineer wants to increase the effectiveness and accuracy of this machine. Based on the sensor stream below, what operational changes or parameter adjustments could help achieve this? \newline
        \textbf{A:} The TCP offset is misconfigured at [0, 0.3, 55] instead of the correct [0, 0, 55]. Update the TCP position offset in the installation settings to [0, 0, 55]. \\
        \midrule

        \textbf{L4}    & Free form              &
        \textbf{Q:} An engineer wants to increase the effectiveness and accuracy of this machine. Based on the sensor stream below, what operational changes or parameter adjustments could help achieve this? \newline
        \textbf{A:} The payload center of gravity is set to [25, 0, 20] but the correct value is [0, 0, 20]. Update the CoG offset in the installation settings to [0, 0, 20].    \\
    \end{longtable}
\end{small}

\section{Simulation setup and sim-to-real gap analysis}
\label{app:simulation_setup}

We run sim2real rollouts in Isaac Sim using the built-in UR3 asset, with phase-consistent waypoint replay of the recorded \texttt{target\_joint\_*} signals. The simulated task is UR3 pick-and-place, following the 10-phase structure of Table~\ref{tab:phases_pp}. Payload settings are matched episode-wise to the FactoryWave chronology.

Each real episode is exported, converted to a per-episode simulation config, and replayed in Isaac. Simulated outputs preserve the FactoryWave field schema (\texttt{joint\_*}, \texttt{tcp\_*}, \texttt{task\_phase}, etc.) and are paired with their real counterpart by episode ID. Gap metrics are computed phase-wise on aligned trajectories: joint RMSE (deg), TCP position RMSE (mm), TCP rotation error (mrad), and Wasserstein-1 on \texttt{joint\_current\_*} (A).

Across 1{,}155 paired episodes, pooled per-episode metrics are: joint RMSE mean/median $=3.65/2.83$\,deg, TCP position RMSE $=13.16/13.26$\,mm, and W1 effort mean $=0.83/0.81$\,A. TCP rotation error remains broad (median $2605$\,mrad), indicating orientation mismatch is the least stable component. Translational kinematics and effort-proxy alignment are substantially more consistent than rotational alignment.

\section{Representation probing: does the model encode what it cannot report?}
\label{app:probing}

FactoryBench, as presented in the main text, is a purely \emph{behavioural} evaluation: a model is prompted with telemetry and graded on its final answer. A wrong answer therefore conflates two qualitatively different failure modes: (a)~the relevant machine concept is \emph{absent} from the model's internal representation, or (b)~the concept is \emph{present} in the representation but the model cannot surface it through its output. Distinguishing them matters for interpreting the Level-4 collapse of Section~\ref{subsec:l4}, and for localising where progress must come from (better decoding versus better pretraining). We separate the two with a targeted linear-probing study~\cite{alain2016probes} on an open-weight panel member, in the spirit of the geometric analyses of motion transformers in \emph{Words in Motion}~\cite{tas2025words}. All numbers below reproduce from the released probing code.

\paragraph{Setup.} We probe \textbf{Qwen3-4B} (frozen, no fine-tuning). Each item is passed through the model once on the \emph{exact} prompt the zero-shot panel saw, and we cache the residual-stream activation at every layer ($0$=embedding through $36$) at two read-out positions: the last prompt token, and the mean over the time-series token span. All activations are computed with causal attention, verified by a startup self-test that checks that layer-$k$ predictions are invariant to removing later tokens.

\paragraph{Probes and controls.} For each (concept, layer, read-out) we fit an $L_2$ logistic-regression probe on the top-$256$ principal components of the standardised activations, with the standardiser and PCA fit \emph{inside} each cross-validation fold so no validation point shapes its own projection. The regularisation strength is chosen by \emph{episode-grouped} 3-fold cross-validation, and the reported layer is the one that maximises this CV score (never the test accuracy, to avoid a layer-selection peek). A one-hidden-layer MLP is a non-linear reference. Four controls guard the read-out: (i)~\textbf{selectivity}~\cite{hewitt2019control}, the probe accuracy minus that of an identical probe trained on \emph{shuffled} labels, which stays near zero unless the probe reads a real concept rather than memorising; (ii)~a \textbf{raw-signal baseline}, the same probe fit on per-channel signal statistics (mean/std/min/max/last, with the timestamp stripped so no positional information leaks), which measures what is decodable from the numbers with \emph{no model}; (iii)~a \textbf{random-init control}, the same probe pipeline on a same-architecture network with \emph{untrained} weights, which measures how much of the decodability is generic random-feature structure of the input rather than something the model learned in pretraining; and (iv)~\textbf{episode-disjoint} train/test splits with bootstrap $95\%$ confidence intervals that resample whole \emph{episodes} (not items, which are correlated within an episode). Headline probe accuracies are the mean over three random seeds (split, PCA, and $C$-selection). We probe two machine concepts. Neither concept is a benchmark label in its own right: both are derived from the fault identifier recorded with each episode. \emph{Fault type} collapses that identifier to the six most frequent injected mechanisms, chosen so the classes carry comparable numbers of items and a six-way probe is not dominated by one mechanism; the long tail of rarer faults is excluded rather than merged into a catch-all, which would make the sixth class a mixture and the accuracy uninterpretable. \emph{Fault presence} is the binary reduction of the same identifier, healthy against any injected fault.

\paragraph{Behavioural read-out \texorpdfstring{$a$}{a}.} To place each probe against the model's own competence we ask the same frozen model the concept directly, on the same held-out probe items: fault presence as a yes/no question, and fault type as an A--F multiple choice over exactly the probe's six classes, with the option-to-letter assignment \emph{randomised per item} so a positional answer bias cannot masquerade as knowledge (scored against the recorded fault identifier). We let the model reason (chain-of-thought, up to $1024$ tokens for the six-way question), take its committed answer when it gives one, and for items where it reasons without committing we read its choice \emph{parse-free} from the logits over the option tokens after a ``therefore the answer is'' cue, so every item is scored and no inference is discarded. Probe accuracy $p$ and behavioural accuracy $a$ are computed on the identical test items, giving a matched read: $p$ asks whether the concept is decodable, $a$ whether the model uses it.

\begin{figure}[htbp]
    \centering
    \includegraphics[width=\linewidth]{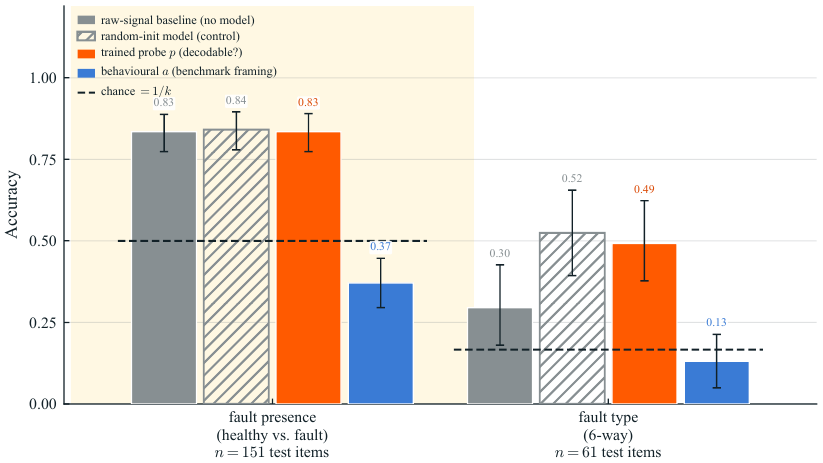}
    \caption{\textbf{Decodable is not the same as learned, and neither is the same as reported.} For fault presence and fault type we compare the trained linear probe's accuracy $p$ (is the concept decodable from Qwen3-4B's frozen residual stream?) with the model's own behavioural accuracy $a$ on the same question, alongside a raw-signal baseline (no model), a random-init control (same architecture, untrained weights), and chance. Fault presence is decodable at $p{=}0.83$, but the model never approaches it behaviourally: $a{=}0.37$ under the benchmark's anomaly wording, where it replies ``anomalous: yes'' to every episode, far below $0.83$. Raw signal statistics ($0.83$) and even an untrained network ($0.84$) decode it just as well, so this availability is surface, not learned. Fault type is only weakly decodable ($p{=}0.49$, CV-selected) and, crucially, \emph{not above the random-init control} ($0.52$), which in fact edges it out; the model answers at $a{=}0.13$. Chance is $1/k$ (uniform random guessing): $0.5$ for the binary presence task, $1/6$ for the six-way type task. Each probe is read from the CV-selected best layer for the last-token read-out, layer~$2$ for fault presence and layer~$1$ for fault type; the layer is chosen by episode-grouped cross-validation on the training folds and never on the test split, so the reported accuracy is not a layer-selection peek. Error bars are episode-level bootstrap $95\%$ CIs and are shown for \emph{every} condition, including the raw-signal and random-init baselines, so no bar is read as a point estimate against an interval-bearing one; accuracies are on the held-out test split. The random-init control is run at the same sample size, seed count and split as the trained probes ($600$ items per concept, three seeds, $151$/$61$ test items), so the intervals are directly comparable. They are wide relative to the gaps: on fault presence the trained probe, the raw-signal baseline and the random-init control overlap almost entirely, and on fault type the control's estimate ($0.52$) sits slightly \emph{above} the trained probe's ($0.49$), with each inside the other's interval. This is why we read the fault-type result as no learned gain rather than a small one.}
    \label{fig:probe_vs_readout}
\end{figure}

\paragraph{Fault presence is decoded but not reliably reported.} Fault presence is linearly decodable at $p{=}0.83$ (three-seed mean $0.86{\pm}0.02$; mean-pool $0.86$; selectivity $0.34$, so this is not probe memorisation). The model's behavioural read-out, by contrast, is unreliable. On the benchmark's own anomaly wording it answers ``yes'' to \emph{every} episode and scores $a{=}0.37$, below the $0.5$ chance of random guessing, because only $38\%$ of episodes are faulty, so an always-``anomaly'' policy scores at that base rate (Figure~\ref{fig:probe_vs_readout}, left). The model therefore applies a miscalibrated policy biased toward predicting a fault, and does not approach the $0.83$ that is linearly available. The gap is moreover entirely at the read-out: the raw-signal baseline ($0.83$) and the random-init network ($0.84$) both decode fault presence just as well, so it is an ``easy'' property already carried by signal statistics and even by untrained features, to which the pretrained model adds no representational value.

\begin{figure}[htbp]
    \centering
    \includegraphics[width=\linewidth]{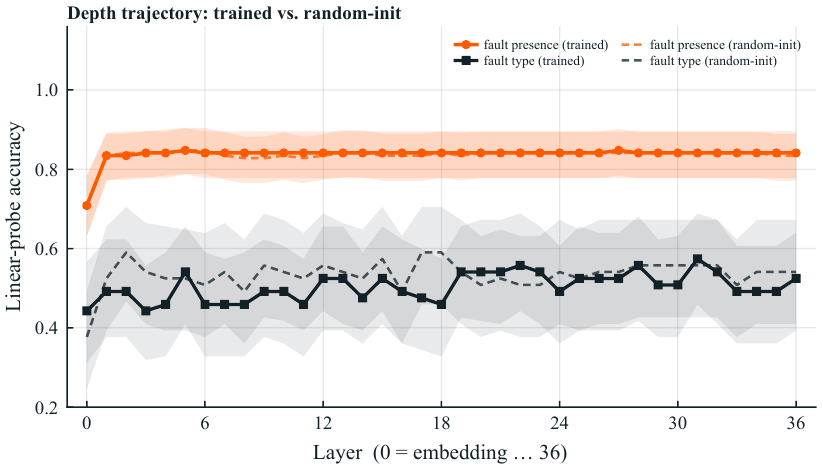}
    \caption{\textbf{An untrained network decodes fault type as well as the trained one.} Per-layer linear-probe accuracy (last-token read-out), trained (solid) vs.\ random-init control (dashed). Fault presence is decodable already at the embedding ($0.71$) and saturates immediately, but the random-init curve matches it. Fault type's trained curve tracks its random-init curve at \emph{every} depth, showing no learned gain, and the CV-selected best is shallow. Behavioural read-out accuracies for both concepts are reported in Figure~\ref{fig:probe_vs_readout}. Episode-level bootstrap $95\%$ CIs are shaded for both the trained probes and the random-init controls, which are distinguished by their dashed centre lines; for fault type the two bands overlap at every depth. Trained probes and control are run at the same sample size, seed count and split, so the bands are directly comparable.}
    \label{fig:probe_layerwise}
\end{figure}

\paragraph{Fault type is neither learned nor reported.} Fault \emph{type} is only weakly decodable: $p{=}0.45{\pm}0.07$ (three-seed mean; the CV-selected layer is shallow, e.g.\ layer~$1$), above the raw-signal baseline ($0.30$) but \emph{not} above the random-init control ($0.52$ at its CV-selected layer, $0.45{\pm}0.06$ across three seeds against the trained probe's $0.45{\pm}0.07$). All three clear the $1/6{\approx}0.17$ chance of the six-way task, so the signal does carry some fault-type structure, but the trained probe extracts no more of it than an untrained network. A same-architecture network with untrained weights decodes fault type as well as the pretrained one, and its per-layer curve tracks the trained probe at every depth (Figure~\ref{fig:probe_layerwise}); the trained model's marginal contribution over an untrained one is within noise. This decodability therefore reflects generic random-feature structure of the input rather than a representation the model learned in pretraining. Behaviourally the model reaches only $a{=}0.13$, below the $1/6$ chance of the six-way task. Given a chain-of-thought budget it does commit an answer on $84\%$ of items, and its errors now spread across five of the six classes: the earlier apparent ``collapse onto two classes'' was an artefact of a fixed option order, which we remove by randomising the A--F assignment per item. For fault type, then, we find neither a learned representation nor a usable read-out.

\paragraph{Takeaway.} The two concepts sit at different points on a representation--read-out spectrum. Fault \emph{presence} is a \emph{read-out} failure: it is trivially decodable ($p{=}0.83$, matched by raw signal statistics and by an untrained network), yet the model answers below chance ($0.37$), applying an anomaly-predicting policy it cannot calibrate. Fault \emph{type} is weaker still: the pretrained model shows no decodable representation beyond a random-init baseline ($0.45{\pm}0.07$ vs.\ $0.45{\pm}0.06$ across three seeds) \emph{and} cannot report it ($a{=}0.13$): a representational gap, not merely a decoding one. Part of the Level-4 collapse of Section~\ref{subsec:l4} is therefore a genuine read-out/calibration failure where the concept is present, and part reflects the absence of a learned representation where it is not. Near-term progress will need both: better decoding and calibration for the surface concepts, and better representation learning, from pretraining or fine-tuning on a frozen backbone, for the computed ones.

\paragraph{Scope.} The study probes a single open-weight model (the only panel member exposing activations and within computational reach) and two concepts with dense ground truth. Two cautions on interpretation. First, a linear probe measures a \emph{lower bound} on decodability under a specific read-out: a near-chance probe bounds what a linear read-out recovers, not the information content of the signal, so our fault-type result is ``the model does not linearly represent this beyond random features,'' not ``the information is absent'' (the random-init control already recovers $0.44$, so the signal is not empty). Second, the probe is supervised on frozen features whereas the model answers zero-shot, an asymmetry intrinsic to probing, where $p$ measures latent \emph{availability} and $a$ measures zero-shot \emph{usability}. Full per-layer numbers for both read-outs, all four controls, the multi-seed spreads, and the exact reproduction command are released with the benchmark code.

\section{Causal schema (SCE): full details}
\label{app:sce}

This appendix gives the long-form definition of the Setpoint--Context--Effort\,+\,Feedback (SCE) schema used to organize every time-series channel in FactoryBench, together with the residual-based fault definition that the schema enables. The motivation is summarized in Section~\ref{sec:sce}; here we cover the per-group channel breakdown, the residual formalism, and the per-robot calibration that makes faults computable rather than imputed.

The schema groups every recorded channel into one of three causal roles. \textbf{Setpoint} captures the controller's commanded target: per-axis position, velocity, and acceleration setpoints emitted by the trajectory generator. \textbf{Context} captures physical and configured conditions affecting behavior, split into \emph{static} context (episode-level metadata such as payload mass, payload center of gravity, TCP offset, gripper model, and box / peg geometry) and \emph{dynamic} context (per-sample channels such as joint temperatures, supply voltages, controller mode, and safety mode). \textbf{Effort\,+\,Feedback} captures the machine's measured response: motor currents and torques (effort), measured joint positions and TCP pose (feedback), tool accelerometer, and acoustic emission where available. The exact per-robot channel mapping appears in the signal-group tables in this appendix (UR3 and KUKA KR10).

This split enables a single operational definition of a fault that applies across machines and tasks. Under healthy operation, the measured Effort\,+\,Feedback response $\mathbf{y}_t$ is by definition a function of the Setpoint command $\mathbf{S}$ and the Context $\mathbf{C}_t$:
$$\mathbf{y}_t = f(\mathbf{S}, \mathbf{C}_t),$$
where $f$ represents the nominal plant dynamics (inverse-dynamics-plus-controller transfer) calibrated per robot. Faults manifest as structured residuals from this expected baseline: $\mathbf{y}_t - f(\mathbf{S}, \mathbf{C}_t)$ forms a clear, fault-specific spatiotemporal pattern. Because every episode supplies the paired $(\mathbf{S}, \mathbf{C}_t, \mathbf{y}_t)$ streams, this residual is computable directly from the data, giving ground truth that does not have to be hand-imputed. Figure~\ref{fig:sce_schema} summarises the causal relationships between the three groups.

\begin{figure}[htbp]
    \centering
    \begin{tikzpicture}[
            every node/.style={font=\sffamily},
            causalNode/.style={
                    fill=white, rounded corners=3pt, align=center,
                    minimum width=2.0cm, minimum height=1.1cm, inner sep=3pt,
                    line width=1pt
                }
        ]
        \fill[warm_amber, rounded corners=4pt] (-0.5, 0) rectangle (8.0, 4.0);
        \draw[flicker, line width=1.2pt, rounded corners=4pt] (-0.5, 0) rectangle (8.0, 4.0);

        \node[causalNode, draw=forgis_orange] (nodeS) at (0.8, 2.5) {%
            \textbf{\textcolor{forgis_orange}{$S$}}\\[-0.4ex]
            \scriptsize\textcolor{gunmetal}{Setpoint}\\[-0.4ex]
            \scriptsize\textcolor{black}{(target commands)}
        };
        \node[causalNode, draw=blue] (nodeC) at (3.75, 2.5) {%
            \textbf{\textcolor{blue}{$C$}}\\[-0.4ex]
            \scriptsize\textcolor{gunmetal}{Context}\\[-0.4ex]
            \scriptsize\textcolor{black}{(environment)}
        };
        \node[causalNode, draw=green] (nodeE) at (6.7, 2.5) {%
            \textbf{\textcolor{green}{$E$}}\\[-0.4ex]
            \scriptsize\textcolor{gunmetal}{Effort+Feedback}\\[-0.4ex]
            \scriptsize\textcolor{black}{(response)}
        };
        \draw[forgis_orange, line width=1.2pt, -{Stealth}] (nodeS.north) to[bend left=18] ([xshift=-0.1cm]nodeE.north);
        \draw[blue, line width=1.2pt, -{Stealth}] (nodeC.east) -- (nodeE.west);
        \node[font=\sffamily\small, text=gunmetal] at (3.75, 1.3) {Healthy: $E = f(S, C)$ \quad \textbar \quad Fault: $E \neq f(S, C)$};
        \node[font=\sffamily\scriptsize, text=black] at (3.75, 0.65) {Fault $\Leftrightarrow$ empirical deviation in \textit{Effort} given \textit{Setpoint} and \textit{Context}};
    \end{tikzpicture}
    \caption{Physical causal schema. Setpoint commands $S$ and contextual variables $C$ jointly determine the measured Effort\,+\,Feedback response $E$ under healthy operation. Faults manifest as residuals $E - f(S, C)$ from this nominal mapping.}
    \label{fig:sce_schema}
\end{figure}

\section{License and availability}
\label{app:license}

FactoryBench and FactoryWave are freely available for academic and other non-commercial use under a two-track licence: the \textbf{generator source code, evaluation scripts, and tooling} (this repository) are released under \ifanon\textbf{a non-commercial source-code licence}\else\textbf{the Forgis Source Code License (Non-Commercial)}\fi, and the \textbf{data and benchmark artefacts} (FactoryWave episodes, question templates, paraphrase banks, LLM-as-judge prompts, and the full Q\&A dataset) are released under \textbf{Creative Commons Attribution-NonCommercial 4.0 (CC~BY-NC~4.0)}, with attribution to be given to the FactoryBench authors. Commercial use of either track requires a separate licence from the authors. Both artefact tracks are distributed via the public Hugging Face repository \texttt{FactoryBench/FactoryBench}. To guard against test-set contamination in future model releases, we hold back a private $15\%$ subset of the Q\&A pairs at the \emph{episode} level (so all questions grounded in any held-out episode stay private across all four levels) and keep its raw episode data, gold answers, and rubric annotations off the public release; this private slice is used internally for contamination checks and may also be used to score future evaluees on request. The remaining $85\%$ is fully public and is released as a single undivided pool, alongside the balanced $3{,}000$-item FactoryBench-Lite subset of Appendix~\ref{app:lite}.

We provide a versioned release track (e.g., \texttt{v1.0}, \texttt{v1.1}) so that reported numbers always refer to a fixed benchmark state; corrections to labels or templates are published as minor-version updates with a public changelog, and the exact version used in any evaluation is stamped into every result file. \textbf{All counts, tables, and results reported in this paper correspond to release} \texttt{\textbf{FactoryBench/FactoryBench:v1.0.0}}\textbf{, the paper-submission snapshot} (69{,}691 released Q\&A items grounded in 8{,}983 FactoryWave episodes). The live \texttt{main} branch of the dataset repository may contain additional episodes and Q\&A items added after the submission snapshot; those additions are versioned in the changelog and do not retroactively affect the numbers here. Bug reports and label-correction requests are tracked on the public repository at \ifanon\url{https://anonymous.4open.science/r/FactoryBench/README.md}\else\url{https://github.com/Forgis-Labs/Factorybench}\fi.

\ifneuripschecklist\input{checklist}\fi

\end{document}